\documentclass[a4paper,final,10pt]{elsarticle}

\usepackage{amsmath,amssymb,amsfonts}
\usepackage{mathrsfs}
\usepackage{float}
\usepackage{booktabs}
\usepackage{pdflscape}
\usepackage{makeidx} 
\usepackage{algorithm}
\usepackage{algorithmic}
\usepackage{theorem}
\usepackage{graphics}                 
\usepackage{accents}
\usepackage[margin=10pt,font=small,labelfont=bf]{caption}
\usepackage{tikz}
\usepackage{verbatim}
\usetikzlibrary{arrows,automata}
\usepackage{exscale,relsize}
\usepackage{mleftright}
\mleftright 

\newcommand{\dist}{{\textstyle \mathsmaller{\varDelta}}}
\newcommand{\surdist}{\dist^{\mathrm{sur}}}

\def\D{\mathcal{D}}
\graphicspath{{}}

\usepackage[english]{babel}
\usepackage{todonotes}

\begin{document}

\sloppy 

\begin{frontmatter}


\title{A Bag-of-Paths Framework for Network Data Analysis \\
\small{(ArXiv preprint manuscript submitted for publication)}
}

 \author[UCL]{Kevin Fran\c{c}oisse}
 \author[UCL]{Ilkka Kivim\"aki}
 \author[Yahoo]{Amin Mantrach}
 \author[Paris]{\\Fabrice Rossi} 
 \author[UCL,ULB]{Marco Saerens} 
 
 \address[UCL]{Universit\'{e} catholique de Louvain, Belgium}
 \address[ULB]{Universit\'{e} Libre de Bruxelles, Belgium}
 \address[Yahoo]{Yahoo! Research, Sunnyvale, California, USA}
 \address[Paris]{Universit\'{e} Paris 1 Panth\'{e}on-Sorbonne, France}



\begin{abstract}
This work develops a generic framework, called the bag-of-paths (BoP),
for link and network data analysis. The central idea is to assign
a probability distribution on the set of all paths in a network.
More precisely, a Gibbs-Boltzmann distribution is defined over a bag of paths
in a network, that is, on a representation that considers all paths
independently. We show that, under this distribution,
the probability of drawing a path connecting two nodes can easily
be computed in closed form by simple matrix inversion. This probability
captures a notion of relatedness between nodes of the graph: two nodes
are considered as highly related when they are connected by many,
preferably low-cost, paths.
As an application, two families of distances between nodes are derived from the BoP probabilities.
Interestingly, the second distance family interpolates between the shortest path distance and the resistance distance. In addition, it extends the Bellman-Ford formula for computing the shortest path distance in order to integrate sub-optimal paths by simply replacing the minimum operator by the soft minimum operator.
Experimental results on semi-supervised classification show that both of the
new distance families are competitive with other state-of-the-art approaches.
In addition to the distance measures studied in this paper, the bag-of-paths framework enables straightforward computation of many other relevant network measures.
\end{abstract}



\end{frontmatter}

\section{Introduction}

\subsection{General introduction}

Network and link analysis is a highly studied field, subject of much recent work in various areas of science: applied mathematics, computer science, social science, physics, chemistry, pattern recognition,
applied statistics, data mining \& machine learning, to name a few \cite{Barabasi-2015,chung06,Estrada-2012,Kolaczyk-2009,Lewis09,Newman-2010,Thelwall04,Wasserman-1994}.
Within this context, one key issue is the proper quantification of the structural relatedness between nodes of a network by taking both direct and indirect connections into account.
This problem is faced in all disciplines involving networks in various types of problems such as link prediction, community detection, node classification, and network visualization to name a few popular ones.

The main contribution of this paper is in presenting in detail the \textbf{bag-of-paths} (BoP) framework and defining relatedness as well as distance measures between nodes from this framework. The BoP builds on and extends previous work dedicated to the exploratory analysis of network data \cite{Kivimaki-2012,Kivimaki-2014,Mantrach-2009,Yen-08K}. 
The introduced distances are constructed to capture the global structure of the graph by using paths on the graph as a building block. In addition to relatedness/distance measures, various other quantities of interest can be derived within the probabilistic BoP framework in a principled way, such as betweenness measures quantifying to which extent a node is in between two sets of nodes \cite{Lebichot-2014}, extensions of the modularity criterion for, e.g., community detection \cite{Devooght-2014}, measures capturing the criticality of the nodes or robustness of the network, graph cuts based on BoP probabilities, and so on.

\subsection{The bag-of-paths framework}
More precisely, we assume given a weighted directed, strongly connected, graph or network $G$ where a cost is associated to each edge. Within this context, we consider a bag containing all the possible (either absorbing or non-absorbing) paths\footnote{Also called walks in the litterature.} between pairs of nodes in $G$. In a first step, following \cite{Akamatsu-1996,Mantrach-2009,Saerens-2008,Yen-08K}, a probability distribution on this countable set of paths can be defined by minimizing the total expected cost between all pairs of nodes while fixing the total relative entropy spread in the graph. This results in a Gibbs-Boltzmann distribution, depending on a temperature parameter $T$, on the set of paths such that long (high-cost) paths have a low probability of being sampled from the bag, while short (low-cost) paths have a high probability of being sampled.

In this probabilistic framework, the \textbf{BoP probabilities}, $\text{P}(s=i,e=j)$, that a sampled path has node $i$ as its starting node and node $j$ as its ending node can easily be computed in closed form by a simple $n \times n$ matrix inversion, where $n$ is the number of nodes in the graph. These BoP probabilities play a crucial role in our framework for that they capture the relatedness between two nodes $i$ and $j$ -- the BoP probability will be high when the two nodes are connected by many, short, paths.
In summary, the BoP framework has several interesting properties:
\begin{itemize}
\item It has a clear, intuitive, interpretation.
\item The temperature parameter $T$ allows to monitor randomness by controlling the balance between exploitation and exploration.
\item The introduction of independent costs results in a large degree of customization of the model, according to the problem requirements: some paths could be penalized because they visit undesirable nodes having adverse features.
\item The framework is rich. Many useful quantities of interest can be defined according to the BoP probabilistic framework: distance measures, betweenness measures, etc. This is discussed in the conclusion.
\item The quantities of interest are easy to compute.
\end{itemize}

It, however, also suffers from a drawback: the different quantities are computed by solving a system of linear equations, or by matrix inversion. More precisely, the distance between a particular node and all the other nodes can be computed by solving a system of $n$ linear equations, while all pairwise distances can
be computed at once by inverting an $n \times n$ square matrix. This results in $\mathcal{O}(n^{3})$ computational complexity. Even more importantly, the matrix of distances necessitates $\mathcal{O}(n^{2})$ storage, altough this can be alleviated by using, e.g., incomplete matrix factorization techniques.

This means that the different quantities can only be computed reasonably on small to medium size graphs (containing a few tens of thousand nodes). However, in specific applications like classification or extraction of top eigenvectors, we can avoid computing explicitly the matrix inversion (see PageRank and the power method \cite{Langville-2006}, or large scale semi-supervised classification on graphs \cite{Mantrach-2011}). In addition, it is also possible to restrict the set of paths to ``efficient paths", that is, paths that do not backtrack
(always getting further from the starting node), and compute efficiently the distances from the starting node by a recurrence formula, as proposed in transportation theory \cite{Dial71}.

\subsection{Deriving node distances from the BoP framework}

The paper first introduces the BoP framework in detail. After that, the two families of distances between nodes are defined, and are coined the \textbf{surprisal distance} and the \textbf{potential distance}. Both distance measures satisfy the triangle inequality, and thus satisfy the axioms of a metric. Moreover, the potential distance has the interesting property of generalizing the shortest path and the commute cost distances by computing an intermediate distance, depending on the temperature parameter $T$. When $T$ is close to zero, the distance reduces to the standard shortest path distance (emphasizing exploitation) while for $T \rightarrow \infty$, it reduces to the commute cost distance (focusing on exploration). The commute cost distance is closely related to the resistance distance \cite{FoussKDE-2005,Klein-1993}, as the two functions are proportional to each other (as well as to the commute time distance) \cite{Chandra-1989,Kivimaki-2012}.

This is of primary interest as it has been shown that both the shortest path distance and the resistance distance suffer from some significant flaws. While relevant in many applications, the shortest path distance cannot always be considered as a good candidate distance in network data. Indeed, this measure only depends on the shortest paths and thus does not integrate the ``degree of connectivity" between the two nodes. In many applications, for a constant shortest path distance, nodes connected by many indirect paths should be considered as ``closer" than nodes connected by only a few paths. This is especially relevant when considering relatedness of nodes based on communication, movement, etc, in a network which do not always happen optimally, nor completely randomly.

While the shortest path distance fails to take the whole structure of the graph into account, it has also been shown that the resistance distance converges to a useless value, only depending on the degrees of the two nodes, when the size of the graph increases (the random walker is getting ``lost in space" because the Markov chain mixes too fast, see \cite{vonLuxburg-2010}). Moreover, the resistance distance, which is proportional to the commute cost distance, assumes a completely random movement or communication in the network, which is also unrealistic.

In short, shortest paths do not integrate the amount of connectivity between the two nodes whereas random walks quickly loose the notion of proximity to the initial node when the graph becomes larger \cite{vonLuxburg-2010}.

There is therefore a need for introducing distances interpolating between the shortest path distance and the resistance distance, thus hopefully avoiding the drawbacks appearing at the ends of the spectrum. These quantities capture the notion of relative \emph{accessibility} between nodes, a combination of both proximity in the network and amount of connectivity.

Furthermore, and interestingly, a simple local recurrence expression, extending the Bellman-Ford formula for computing the potential distances from one node of interest to all the other nodes is also derived. It relies on the use of the so-called soft minimum operator \cite{Cook-2011} instead of the usual minimum. Finally, our experiments show that these distance families provide competitive results in semi-supervised learning.

\subsection{Contributions and organization of the paper}

Thus, in summary, this work has several contributions:
\begin{itemize}
\item It introduces a well-founded bag-of-paths framework capturing the global structure of the graph by using network paths as a building block.
\item It is shown that the bag-of-hitting-paths probabilities can easily be computed in closed form. This fundamental quantity defines an intuitive relatedness measure between nodes.
\item It defines two families of distances capturing the structural dissimilarity between the nodes in terms of relative accessibility. The distances between all pairs of nodes can be computed conveniently by inverting a $n \times n$ matrix.
\item It is shown that one of these distance measures has some interesting properties; for instance it is graph-geodetic and it interpolates between the shortest path distance and the resistance distance (up to a scaling factor).
\item The framework is extended to the case where non-uniform priors are defined on the nodes.
\item We prove that this distance generalizes the Bellman-Ford formula computing shortest path distances, by simply replacing the $\mathrm{min}$ operator by the $\mathrm{softmin}$ operator.
\item The distances obtain promising empirical results in semi-supervised classification tasks when compared to other, kernel-based, methods.
\end{itemize}

Section \ref{sec:RelatedWork} develops related work and introduces the necessary background and notation. Section \ref{sec:BOP} introduces the BoP framework, defines BoP probabilities and shows how it can be computed in closed form. Section \ref{Sec_bag_hitting_paths01} extends the framework to hitting, or absorbing, paths. In Section \ref{sec:DistanceMeasures}, the two families of distances as well as their properties are derived. Section \ref{Eq_non_uniform_priors01} generalizes the framework to non-uniform priors on the nodes. An experimental study of the BoP framework with application to semi-supervised classification is presented in Section \ref{sec:Exp}. Concluding remarks and extensions are discussed in Section \ref{Sec_conclusion01}.

\section{Related work, background, and notation}
\label{sec:RelatedWork}

\subsection{Related work}

This work is related to similarity measures on graphs for which some background is presented in this section. The presented BoP framework also has applications in semi-supervised classification, on which our experimental section will focus on in Section~\ref{sec:Exp}. A short survey related to this problem can be found in subsection \ref{sec:SemiSupervised}.

Similarity measures on a graph determine to what extent two nodes in a graph
resemble each other, either based on the information contained in the node
attributes or based on the graph structure. In this work, only measures based
on the graph structure will be investigated. Structural similarity measures
can be categorized into two groups: local and global \cite{Lu-2011}. On the
one hand, local similarity measures between nodes consider the direct links from a node to the other nodes as features and use these features in various way to provide similarities. Examples include the cosine coefficient \cite{Dunham-2003} and the standard correlation \cite{Wasserman-1994}. On the other hand, global similarity measures consider the whole graph structure to compute similarities. Our short review of similarity measures is largely inspired by the surveys appearing in \cite{FoussKernelNN-2011,Mantrach-2009,Yen-2008,Yen-08K}.

Certainly the most popular and useful distance between nodes of a graph is the shortest path distance. However, as discussed in the introduction, it is not always relevant for quantifying the similarity of nodes in a network.

Alternatively, similarity measures can be based on random walk models on the graph, seen as a Markov chain. 
As an example, the commute time (CT) kernel has been introduced in \cite{FoussKDE-2005,Saerens04PCA} as the Moore-Penrose pseudoinverse, $\mathbf{L}^{+}$, of the Laplacian matrix. The CT kernel was inspired by the work of Klein \& Randic \cite{Klein-1993} and Chandra et al.~\cite{Chandra-1989}. More precisely, Klein \& Randic \cite{Klein-1993} suggested to use the effective resistance between two nodes as a meaningful distance measure, called the resistance distance. 
Chandra et al.~\cite{Chandra-1989} then showed that the resistance distance equals the commute time distance, up to a constant factor. The CT distance is defined as the average number of steps that a random walker, starting in a given node, will take before entering another node for the first time (this is called the average first-passage time \cite{Norris-1997}) and going back to the initial node.

It was then shown \cite{Saerens04PCA} that the elements of $\mathbf{L}^{+}$ are inner products of the node vectors in the Euclidean space where these node vectors are exactly separated by the square root of the CT distance. The square root of the CT distance is therein called the Euclidean CT distance. The relationships between the Laplacian matrix and the \emph{commute cost} distance (the expected \emph{cost} (and not steps as for the CT) of reaching a destination node from a starting node and going back to the starting node) were studied in \cite{FoussKDE-2005}. Finally, an electrical interpretation of the elements of $\mathbf{L}^{+}$ can be found in \cite{Yen-2008}.  However, we saw in the introduction that these random-walk based distances suffer from some drawbacks (e.g., the so-called ``lost in space" problem, \cite{vonLuxburg-2010})

Sarkar et al.\ \cite{Sarkar2007} suggested a fast method for computing truncated commute time neighbors.  
At the same time, several authors defined an embedding that preserves the commute time distance with applications in various fields such as clustering \cite{Luh-2005}, collaborative filtering \cite{FoussKDE-2005,Brand-05}, dimensionality reduction of manifolds \cite{Ham2004} and image segmentation \cite{Qiu2005}.

Instead of taking the pseudoinverse of the Laplacian matrix, a simple regularization leads to a kernel called the regularized commute time kernel \cite{Ito-2005,Chebotarev-1997,Chebotarev-1998a}. Ito et al.\ \cite{Ito-2005}, further propose the modified regularized Laplacian kernel by introducing another parameter controlling the importance of nodes. This modified regularized Laplacian kernel is also closely related to a graph regularization framework introduced by Zhou \& Scholkopf in \cite{Zhou-04}, extended to directed graphs in \cite{Zhou-05}. 

The exponential diffusion kernel, introduced by Kondor \& Lafferty \cite{Kondor-2002} and the Neumann diffusion kernel, introduced in \cite{Scholkopf-2002} are similar and based on power series of the adjacency matrix. A meaningful alternative to the exponential diffusion kernel, called the Laplacian exponential diffusion kernel (see \cite{Kondor-2002,Smola-03}) is a diffusion model that substitutes the adjacency matrix with 
 the Laplacian matrix.

Random walk with restart kernels, inspired by the PageRank algorithm and adapted to provide relative similarities between nodes, appeared relatively recently in \cite{Gori-2006WebKDD,Pan-2004,Tong-2007}.
Nadler et al. \cite{Nadler-2005,Nadler-2006} and Pons et al. \cite{Pons-2005,Pons-2006} suggested a distance measure between nodes of a graph based on a diffusion process, called the diffusion distance.  The Markov diffusion kernel has been derived from this distance measure in \cite{FoussKernelNN-2011} and \cite{Yen-2011}. The
natural embedding induced by the diffusion distance was called
diffusion map by Nadler et al.\ \cite{Nadler-2005,Nadler-2006} and is related to correspondence analysis
\cite{Yen-2011}.

More recently, Mantrach et al.\@ \cite{Mantrach-2009}, inspired by \cite{Akamatsu-1996,Bell-1995} and subsequently by \cite{Saerens-2008}, introduced a link-based covariance measure between nodes of a weighted directed graph, called the sum-over-paths (SoP) covariance. They consider, in a similar manner as in this paper, a Gibbs-Boltzmann distribution on the set of paths such that high-cost paths occur with low probability whereas low-cost paths occur with a high probability. Two nodes are then considered as highly similar if they often co-occur together on the same -- preferably short -- path. A related co-betweenness measure between nodes has been defined in \cite{Kolaczyk-2009c}.

Moreover, as both the shortest path distance and the resistance distance show some issues, there were several attempts to define families of distances interpolating between the shortest path and more ``global" distances, such as the resistance distance.
In this context, inspired by \cite{Akamatsu-1996,Bell-1995,Saerens-2008}, a parametrized family of dissimilarity measures, called the randomized shortest path (RSP) dissimilarity, reducing to the shortest path distance at one end of the parameter range, and to the resistance distance (up to a constant scaling factor) at the other end, was proposed in \cite{Yen-08K} and extended in \cite{Kivimaki-2012}. Similar ideas appeared at the same time in \cite{Chebotarev-2011,Chebotarev-2012}, based on considering the co-occurences of nodes in forests of a graph, and in \cite{Herbster-2009, vonLuxburg-2011}, based on a generalization of the effective resistance in electric circuits. These two last families are metrics while the RSP dissimilarity does not satisfy the triangle inequality. The potential and the surprisal distances introduced in this work fall under the same catalogue of distance families. See also \cite{Kivimaki-2012,Guex-2015,Guex-2016} for other, closely related, formulations of families of distances based on free energy and network flows.


\subsection{Background and notation}

We now introduce the necessary notation for the bag-of-paths (BoP) framework, providing both a relatedness index and a distance measure between nodes of a network. First, note that, in the sequel, column vectors are written in bold lowercase while matrices are in bold uppercase.

Consider a weighted directed graph or network, $G = (\mathcal{V}, \mathcal{E})$, assumed strongly connected, with a set $\mathcal{V}$ of $n$ nodes (or vertices) and a set $\mathcal{E}$ of edges (or arcs, links). An edge between node $i$ and node $j$ is denoted by $i \rightarrow j$ or $(i,j)$.
%
Furthermore, it is assumed that we are given an adjacency matrix $\mathbf{A}$
with elements $a_{ij} \ge 0$ quantifying in some way the affinity between node $i$ and
node $j$. When $a_{ij} > 0$, node $i$ and node $j$ are said to be adjacent, that is, connected by an edge. Conversely, $a_{ij} = 0$ means that $i$ and $j$ are not connected. We further assume that there are no self-loops, that is, the $a_{ii}=0$. From this adjacency matrix, a standard random walk on the graph is defined in
the usual way. The transition probabilities associated to each node are simply proportional to the affinities and then normalized:
\begin{equation}
p_{ij}^{\mathrm{ref}} = \frac{a_{ij}}{{\sum_{j'=1}^{n}} a_{ij'}}
\label{Eq_Transition_probabilities01}
\end{equation}
Note that these transition probabilities will be used as reference probabilities later; hence the superscript ``ref''.
The matrix $\mathbf{P}^{\mathrm{ref}}$, containing elements $p_{ij}^{\mathrm{ref}}$, is stochastic and called the transition matrix of the natural or reference random walk on the graph.

In addition, we assume that a transition cost, $c_{ij} \ge 0$, is
associated to each link $i \rightarrow j$ of the graph $G$. If there is no
edge between $i$ and $j$, the cost is assumed to take an infinite value,
$c_{ij} = \infty$. For consistency, $c_{ij} = \infty$ if and only if $a_{ij} = 0$.
The cost matrix $\mathbf{C}$ is the matrix containing
the immediate costs $c_{ij}$ as elements. We will assume that at least one element
of $\mathbf{C}$ is strictly positive. A path $\wp$ is a finite sequence of jumps
to adjacent nodes on $G$
(including loops), initiated from a starting node $s=i$, and stopping in an
ending node $e=j$. The \emph{total cost} of a path $\wp$ is simply the sum of
the local costs $c_{ij}$ along $\wp$, while the \emph{length} of a path is the number of steps, or jumps, needed for following that path.

The costs are set independently of the adjacency matrix; they quantify the cost of a transition, depending on the problem at hand. They can, e.g., be defined according to some properties, or features, of the nodes or the edges in order to bias the probability distribution of choosing a path. In the case of a social network, we may, for instance, want to bias the paths in favor of domain experts. In that case, the cost of jumping to a node could be set proportional to the degree of expertise of the corresponding person. Therefore, walks visiting a large proportion of persons with a low degree of expertise would be penalized versus walks visiting persons with a high degree. Another example aims to favor hub-avoiding paths penalizing paths visiting hubs. Then, the cost can be simply set to the degree of the node. If there is no reason to bias the paths with respect to some features, costs are simply set equal to $1$ (paths are penalized by their length) or equal to $c_{ij} = 1/a_{ij}$ (the elements of the adjacency matrix can then be considered as conductances and the costs as resistances).


\section{The basic bag-of-paths framework}
\label{sec:BOP}

Roughly speaking, the BoP model will be based on the probability that a path drawn from a ``bag of paths'' has nodes $i$ and $j$ as its starting and ending nodes, respectively. 
According to this model, the probability of drawing a path starting in node $i$ and ending in node $j$ from the bag-of-paths can easily be computed in closed form. This probability distribution then serves as a building block for several extensions.

The bag-of-paths framework is introduced by first considering bounded paths and then paths of arbitrary length. For simplicity, we discuss non-hitting (or non-absorbing) paths first and then develop the more interesting bag-of-hitting-paths framework in the next section.

\subsection{Sampling bounded paths according to a Gibbs-Boltzmann distribution} 
\label{Subsec_Boltzmann_distribution01}

The present section describes how the probability distribution on
the set of paths is assigned.
In order to make the presentation rigorous, we will first have to consider paths of \emph{bounded length} $t$. Later, we will extend the results for paths with arbitrary length.
Let us first choose two
nodes, a starting node $i$ and an ending node $j$ and define
the set of paths (including cycles) of length $t$ from $i$ to $j$ as
$\mathcal{P}_{ij}(t) = \{\wp_{ij}(t)\}$. Thus, $\mathcal{P}_{ij}(t)$
contains all the paths $\wp_{ij}(t)$ allowing to reach node $j$ from node $i$ in \emph{exactly} $t$ \emph{steps}.

Let us further denote as $\tilde{c}(\wp_{ij}(t))$ the total cost associated to path
$\wp_{ij}(t)$. Here, we assume that $\wp_{ij}(t)$ is a valid path from
node $i$ to node $j$, that is, 
it consists of a sequence of nodes $(k_0=i) \rightarrow k_1 \rightarrow k_2  \rightarrow \dots  \rightarrow (k_{t}=j)$ where
$c_{k_{\tau-1}k_{\tau}} < \infty$ for all $\tau \in [1,t]$.
As already mentioned, we assume that the total cost associated to a path is additive, i.e.
$\tilde{c}(\wp_{ij}(t))={\textstyle \sum\nolimits _{\tau=1}^{t}}c_{k_{\tau-1}k_{\tau}}$.
Then, let us define the set of all $t$-length paths through the
graph between all pairs of nodes as $\mathcal{P}(t)=\cup_{i,j=1}^{n}\mathcal{P}_{ij}(t)$.

Finally, the set of all bounded paths \emph{up to length} $t$ is denoted by $\mathcal{P}(\le t)=\cup_{\tau=0}^{t}\mathcal{P}(\tau)$.
Note that, by convention, for $i = j$ and $t=0$, zero-length paths are allowed with zero associated cost.
Other types of paths will be introduced later; a summary of the mathematical notation appears in Table \ref{Tab_notation01}.

Now, we consider a probability distribution on this finite set $\mathcal{P}(\le t)$, representing
the probability of drawing a path $\wp\in\mathcal{P}(\le t)$ from a bag containing all paths up to length $t$. We search for the
distribution of paths P$(\wp)$ minimizing the expected total cost-to-go,
$\mathbb{E}[ \tilde{c}(\wp)]$,
among all the distributions having a fixed relative entropy $J_0$ with
respect to a reference distribution, here the natural random walk on the graph (see Equation (\ref{Eq_Transition_probabilities01})). This choice naturally defines a probability distribution on the set of paths of maximal length $t$ such that high-cost paths occur with a low probability while short paths occur with a high probability.
In other words, we are seeking for path probabilities, $\textnormal{P}(\wp)$, $\wp\in\mathcal{P}(\le t)$,
minimizing the expected total cost subject to a constant relative
entropy constraint\footnote{In theory, non-negativity constraints should be added, but this is not necessary as the resulting probabilities are automatically non-negative.}:

\begin{equation}
\label{eq:BoPmin}
\vline\,\begin{array}{llll}
\underaccent{\{ \textnormal{P}(\wp) \} : \wp \in \mathcal{P}(\le t) }{\textnormal{minimize}} & {\displaystyle \sum_{\wp\in\mathcal{P}(\le t)}}\text{P}(\wp)\tilde{c}(\wp)\\[0.5cm]
\textnormal{subject to} & \sum_{\wp\in\mathcal{P}(\le t)}\textnormal{P}(\wp)\log(\textnormal{P}(\wp)/\tilde{\text{P}}^{\mathrm{ref}}(\wp))=J_{0} \\ & \sum_{\wp\in\mathcal{P}(\le t)}\textnormal{P}(\wp)=1
\end{array}
\end{equation}
where $J_{0} > 0$ is provided a priori by the user, according to the desired
degree of randomness and $\tilde{\text{P}}^{\mathrm{ref}}(\wp)$ represents the probability of following
the path $\wp$ when walking according to the reference transition probabilities $p_{ij}^{\mathrm{ref}}$ of the natural random walk on $G$ (see Equation
(\ref{Eq_Transition_probabilities01})).

More precisely, we define $\tilde{\pi}^{\mathrm{ref}}(\wp) = \prod_{\tau=1}^{t} p_{k_{\tau-1} k_{\tau}}^{\mathrm{ref}}$, 
that is, the product of the transition probabilities along path $\wp$ -- the likelihood of the path when the starting and ending nodes are known.
Now, if we assume a uniform (non-uniform priors are considered in Section \ref{Sec_bag_hitting_paths01}), independent, a priori probability, $1/n$, for choosing both the starting and the ending node, then we set 
$\tilde{\text{P}}^{\mathrm{ref}}(\wp) = \tilde{\pi}^{\mathrm{ref}}(\wp)/\sum_{\wp' \in \mathcal{P}(\le t)} \tilde{\pi}^{\mathrm{ref}}(\wp')$, which ensures that the reference probability is properly normalized\footnote{We will see later that the path likelihoods $\tilde{\pi}^{\mathrm{ref}}(\wp)$ are already properly normalized in the case of hitting, or absorbing, paths: $\sum_{\wp \in \mathcal{P}^{\mathrm{h}}} \tilde{\pi}^{\mathrm{ref}}(\wp) = 1$. See \ref{app_hitting_paths_likelihood}.}.

The problem~(\ref{eq:BoPmin}) can be solved by introducing the following Lagrange
function
\begin{equation}
\mathscr{L} = \sum_{\wp\in\mathcal{P}(\le t)}\text{P}(\wp)\tilde{c}(\wp) + \lambda\left[\sum_{\wp\in\mathcal{P}(\le t)}\text{P}(\wp)\log\frac{\text{P}(\wp)}{\tilde{\text{P}}^{\mathrm{ref}}(\wp)} - J_{0}\right] + \mu\left[\sum_{\wp\in\mathcal{P}(\le t)}\text{P}(\wp)-1\right]
\label{Eq_Lagrange_function01}
\end{equation}
and optimizing over the set of path probabilities $\{ \text{P}(\wp) \}_{\wp\in\mathcal{P}(\le t)}$. As could be expected, setting its partial derivative
with respect to $\text{P}(\wp)$ to zero and solving the equation yields a \textbf{Gibbs-Boltzmann probability distribution} on the set of paths up to length $t$ \cite{Mantrach-2009},
\begin{equation}
\text{P}(\wp)
= \frac{\tilde{\text{P}}^{\mathrm{ref}}(\wp)\exp\left[-\theta \tilde{c}(\wp)\right]}{{\displaystyle \sum_{\wp' \in\mathcal{P}(\le t)}}\tilde{\text{P}}^{\mathrm{ref}}(\wp')\exp\left[-\theta \tilde{c}(\wp')\right]}
\label{Eq_Probability_distribution01}
\end{equation}
where the Lagrange parameter $\lambda$ plays the role of a temperature
$T$ and $\theta=1/\lambda$ is the inverse temperature. 

Thus, as desired, short paths $\wp$ (having a low cost $\tilde{c}(\wp)$) are favored
in that they have a large
probability of being followed. From Equation (\ref{Eq_Probability_distribution01}),
we clearly observe that when $\theta \rightarrow 0$, the path probabilities reduce
to the probabilities generated by the natural
random walk on the graph (characterized by the transition probabilities
$p_{ij}^{\mathrm{ref}}$ as defined in Equation (\ref{Eq_Transition_probabilities01})).
In this case, $J_0 \rightarrow 0$ as well.
But when $\theta$ is large, the probability distribution defined by
Equation (\ref{Eq_Probability_distribution01}) is biased towards low-cost paths
(the most likely paths are the shortest ones).
Note that, in the sequel, it will be assumed that the user provides the
value of the parameter $\theta$ instead of $J_0$, with $\theta > 0$.
Also notice that the model could be derived thanks to a maximum entropy
principle instead \cite{Jaynes-1957,Kapur-1992}.


\subsection{The bag-of-paths probabilities}
\label{sec:BoPprobabilities}

Our BoP framework will be based on the computation of another important quantity derived from Equation (\ref{Eq_Probability_distribution01}): the probability of drawing a path starting in some node $s=i$ and ending in some other node $e=j$ from the bag of paths. For paths up to length $t$ this is provided by
\begin{align}
\text{P}^{(\le t)}(s=i,e=j) &= \frac{{\displaystyle \sum_{\wp\in\mathcal{P}_{ij}(\le t)}}\tilde{\text{P}}^{\mathrm{ref}}(\wp)\exp\left[-\theta \tilde{c}(\wp) \right]}{{\displaystyle \sum_{\wp' \in\mathcal{P}(\le t)}} \tilde{\text{P}}^{\mathrm{ref}}(\wp')\exp\left[-\theta \tilde{c}(\wp') \right]} \nonumber \\
&= \frac{{\displaystyle \sum_{\wp\in\mathcal{P}_{ij}(\le t)}}\tilde{\pi}^{\mathrm{ref}}(\wp)\exp\left[-\theta \tilde{c}(\wp)\right]}{{\displaystyle \sum_{\wp' \in\mathcal{P}(\le t)}}\tilde{\pi}^{\mathrm{ref}}(\wp')\exp\left[-\theta \tilde{c}(\wp')\right]}
\label{Eq_bag_of_bounded_paths_probability01}
\end{align}
where $\mathcal{P}_{ij}(\le t)$ is the set of paths connecting node $i$ and node $j$ up to length $t$.
From (\ref{Eq_Probability_distribution01}), this quantity simply computes the probability mass of drawing a path connecting $i$ to $j$.
The paths in $\mathcal{P}_{ij}(\le t)$ can contain loops and could visit nodes $i$ and $j$ several times during the trajectory\footnote{Note that another interesting class of paths, the hitting, or absorbing, paths -- allowing only one single visit to the ending node $j$ -- will be considered in the next section \ref{Sec_bag_hitting_paths01}.}.

\begin{table}[t]
\begin{center}
\small
\begin{tabular}{|l|l|}
\hline 
$\wp$ & a particular path \\\hline
$\text{P}(\wp)$ & the probability of drawing path $\wp$ \\\hline
$\mathcal{P}_{ij}(t)$     & set of paths connecting $i$ to $j$ in exactly $t$ steps \\\hline 
$\mathcal{P}_{ij}(\le t)$ & set of paths connecting $i$ to $j$ in at most $t$ steps \\\hline 
$\mathcal{P}(\le t) = \cup_{i,j = 1}^{n} \mathcal{P}_{ij}(\le t)$ & set of all paths of at most $t$ steps \\\hline
$\mathcal{P}_{ij}$        & set of paths of arbitrary length connecting $i$ to $j$ \\\hline 
$\mathcal{P} = \cup_{i,j = 1}^{n} \mathcal{P}_{ij}$ & set of all paths of arbitrary length \\\hline
$\mathbf{P}^{\mathrm{ref}}$ & transition probability matrix with elements $p_{ij}^{\mathrm{ref}}$ \\\hline
$\mathbf{C}$ & cost matrix with elements $c_{ij}$ \\\hline
$\tilde{\text{P}}^{\mathrm{ref}}(\wp)$ & likelihood of following path $\wp$ according to $p_{ij}^{\mathrm{ref}}$ \\\hline
$\tilde{c}(\wp)$ & total cumulated cost when following path $\wp$ \\\hline
\end{tabular}
\caption{Summary of notations for the enumeration of paths in graph $G$.}
\label{Tab_notation01}
\end{center}
\end{table}

\subsubsection{Computation of the bag-of-paths probabilities for bounded paths}
The analytical expression allowing to compute the quantity defined by Equation (\ref{Eq_bag_of_bounded_paths_probability01}) will be derived in this subsection. Then, in the following subsection, its definition will be extended to the set of paths of arbitrary length (unbounded paths) by taking the limit $t \rightarrow \infty$.

We start from the cost matrix, $\mathbf{C}$,
from which we build a new matrix, $\mathbf{W}$, as
\begin{equation}
\mathbf{W}=\mathbf{P}^{\mathrm{ref}}\circ\exp\left[-\theta\mathbf{C}\right] = \exp\left[-\theta\mathbf{C}+\log\mathbf{P}^{\mathrm{ref}}\right]
\label{Eq_W_matrix01}
\end{equation}
where $\mathbf{P}^{\mathrm{ref}}$ is the transition probability
matrix\footnote{Do not confuse matrix $\mathbf{P}^{\mathrm{ref}}$ in bold with $\tilde{\text{P}}^{\mathrm{ref}}(\wp)$ representing the reference probability of path $\wp$. A summary of the notation appears in Table \ref{Tab_notation01}.} of the natural random walk on the graph containing the elements $p_{ij}^{\mathrm{ref}}$, and the logarithm/exponential
functions are taken elementwise. Moreover, $\circ$ is the elementwise
(Hadamard) matrix product. Note that the matrix $\mathbf{W}$ is not symmetric in general.

Then, let us first compute the numerator of Equation (\ref{Eq_bag_of_bounded_paths_probability01}). Because all the quantities
in the exponential of Equation (\ref{Eq_bag_of_bounded_paths_probability01}) are summed along a path, $\log\tilde{\pi}^{\mathrm{ref}}(\wp)={\textstyle \sum\nolimits _{\tau=1}^{t}}\log p_{k_{\tau-1}k_{\tau}}^{\mathrm{ref}}$
and $\tilde{c}(\wp)={\textstyle \sum\nolimits _{\tau=1}^{t}}c_{k_{\tau-1}k_{\tau}}$
where each link $k_{\tau-1}\rightarrow k_{\tau}$ lies on path $\wp$,
we immediately observe that element $i,j$ of the matrix $\mathbf{W}^{\tau}$
($\mathbf{W}$ to the power $\tau$) is $[\mathbf{W}^{\tau}]_{ij}={\textstyle \sum\nolimits _{\wp \in \mathcal{P}_{ij}(\tau)}}\exp[-\theta \tilde{c}(\wp)+\log\tilde{\pi}^{\mathrm{ref}}(\wp)]$
where $\mathcal{P}_{ij}(\tau)$ is the set of paths connecting the starting node
$i$ to the ending node $j$ in \emph{exactly} $\tau$ steps.

Consequently, the sum in the numerator of Equation (\ref{Eq_bag_of_bounded_paths_probability01}) is
\begin{align}
&{\displaystyle \sum_{\wp\in\mathcal{P}_{ij}(\le t)}}\tilde{\pi}^{\mathrm{ref}}(\wp)\exp\left[-\theta \tilde{c}(\wp) \right] 
= {\displaystyle \sum\limits _{\tau=0}^{t}} {\displaystyle \sum_{\wp\in\mathcal{P}_{ij}(\tau)}}\tilde{\pi}^{\mathrm{ref}}(\wp)\exp\left[-\theta \tilde{c}(\wp) \right] \nonumber \\
&\qquad \qquad \qquad = {\displaystyle \sum\limits _{\tau=0}^{t}} \left[\mathbf{W}^{\tau}\right]_{ij} = \left[{\displaystyle \sum\limits _{\tau=0}^{t}}\mathbf{W}^{\tau}\right]_{ij}
= \mathbf{e}_i^{\text{T}}\left({\displaystyle \sum\limits _{\tau=0}^{t}}\mathbf{W}^{\tau}\right)\mathbf{e}_j
 \label{Eq_numerator_bag_paths01}
\end{align}
where $\mathbf{e}_i$ is a column vector full of 0's, except in position $i$ where it contains a 1. By convention, at time step 0, the random walker appears in node $i$ with probability one and a zero cost: $\mathbf{W}^{0} = \mathbf{I}$.
This means that \emph{zero-length paths} (without any transition step) are allowed in $\mathcal{P}_{ij}(\le t)$. If, on the contrary, we want to dismiss zero-length paths, we could redefine $\mathcal{P}_{ij}(\le t)$ as the set as paths of length at least one (the summation starts at $t=1$ instead of $t=0$) and proceed in the same manner.


This previous Equation (\ref{Eq_numerator_bag_paths01}) allows to derive the analytical form of the probability of drawing a bounded path (up to length $t$) starting in node $i$ and ending in $j$. Indeed, replacing Equation (\ref{Eq_numerator_bag_paths01}) in Equation (\ref{Eq_bag_of_bounded_paths_probability01}), and recalling that $\mathcal{P}(\le t) = \cup_{i,j = 1}^{n} \mathcal{P}_{ij}(\le t)$, we obtain
\begin{align}
\text{P}^{(\le t)}(s=i,e=j)
&= \frac{\mathbf{e}_i^{\text{T}}\left({\displaystyle \sum\limits _{\tau=0}^{t}}\mathbf{W}^{\tau}\right)\mathbf{e}_j}{\displaystyle \sum_{i,j=1}^{n}\mathbf{e}_i^{\text{T}}\left({\displaystyle \sum\limits _{\tau=0}^{t}}\mathbf{W}^{\tau}\right)\mathbf{e}_j}
= \frac{\mathbf{e}_i^{\text{T}}\left({\displaystyle \sum\limits _{\tau=0}^{t}}\mathbf{W}^{\tau}\right)\mathbf{e}_j}{\mathbf{e}^{\text{T}}\left({\displaystyle \sum\limits _{\tau=0}^{t}}\mathbf{W}^{\tau}\right)\mathbf{e}}
\label{Eq_bag_of_bounded_paths_probability02}
\end{align}
where $\mathbf{e} = [1,1,\dots,1]^{\text{T}}$ is a vector of 1's. Of course, there is no a priori reason to choose a particular path length; we will therefore consider paths of arbitrary length in the next section.

\subsubsection{Proceeding with paths of arbitrary length}

Let us now consider the problem of computing the probability of drawing a path starting in $i$ and ending in $j$ from a bag containing paths of \emph{arbitrary} length, and therefore usually containing an infinite number of 
paths. Following the definition in the bounded case (Equation (\ref{Eq_bag_of_bounded_paths_probability01})), this quantity will be denoted as and defined by 
\begin{equation}
\text{P}(s=i,e=j)
= \lim_{t \rightarrow \infty} \text{P}^{(\le t)}(s=i,e=j)
= \frac{{\displaystyle \sum_{\wp\in\mathcal{P}_{ij}}}\tilde{\pi}^{\mathrm{ref}}(\wp)\exp\left[-\theta \tilde{c}(\wp) \right]}{{\displaystyle \sum_{\wp' \in\mathcal{P}}} \tilde{\pi}^{\mathrm{ref}}(\wp')\exp\left[-\theta \tilde{c}(\wp') \right]}
\label{Eq_bag_of_paths_probability01}
\end{equation}
where $\mathcal{P}_{ij}$ is the set of paths (of all lengths) connecting $i$ to $j$ in the graph and the denominator is called the \textbf{partition function} of the bag-of-paths system,
\begin{equation}
\mathcal{Z} = \sum_{\wp\in\mathcal{P}} \tilde{\pi}^{\mathrm{ref}}(\wp)\exp\left[-\theta \tilde{c}(\wp) \right]
\label{Eq_bag_of_paths_partition_function01}
\end{equation}
The quantity $\text{P}(s=i,e=j)$ in Equation (\ref{Eq_bag_of_paths_probability01}) will be called the \textbf{bag-of-paths probability} of drawing a path of arbitrary length starting from node $i$ and ending in node $j$. As already stated, this key quantity captures a notion of relatedness, or similarity, between nodes of $G$. From Equation (\ref{Eq_bag_of_paths_probability01}), we observe that two nodes are considered as highly related (high probability of sampling them) when they are connected by many, preferably low-cost, paths, that is, when they are highly accessible. The quantity therefore integrates the concept of (indirect) connectivity, in addition to proximity (low-cost paths).

Now, from Equation (\ref{Eq_bag_of_bounded_paths_probability02}), we need to compute
\begin{equation}
\text{P}(s=i,e=j)
= \lim_{t \rightarrow \infty} \text{P}^{(\le t)}(s=i,e=j)
= \lim_{t \rightarrow \infty} \frac{\mathbf{e}_i^{\text{T}}\left({\displaystyle \sum\limits _{\tau=0}^{t}}\mathbf{W}^{\tau}\right)\mathbf{e}_j}{\mathbf{e}^{\text{T}}\left({\displaystyle \sum\limits _{\tau=0}^{t}}\mathbf{W}^{\tau}\right)\mathbf{e}}
\label{Eq_definition_limit_bag_of_paths01}
\end{equation}

We thus need to compute the well-known power series of $\mathbf{W}$
\begin{equation}
\lim_{t \rightarrow \infty} {\displaystyle \sum\limits _{\tau=0}^{t}}\mathbf{W}^{\tau}
= {\displaystyle \sum\limits _{t=0}^{\infty}}\mathbf{W}^{t}
= (\mathbf{I}-\mathbf{W}\mathbf{)}^{-1} \label{Eq_W_series01}
\end{equation}
which converges if the spectral radius of $\mathbf{W}$ is less than
$1$, $\rho(\mathbf{W}) < 1$. Because the matrix $\mathbf{W}$ only contains non-negative elements and $G$ is strongly connected, a sufficient condition for $\rho(\mathbf{W}) < 1$ is that it is substochastic \cite{Meyer-2000}, which is always achieved for $\theta > 0$ as $c_{ij} \ge 0 $ for all $i, j$ and we assume that at least one element of $\mathbf{C}$ is strictly positive. We therefore assume a $\theta > 0$.

Now, if we pose
\begin{equation}
\mathbf{Z}=(\mathbf{I}-\mathbf{W}\mathbf{)}^{-1}
\label{Eq_fundamental_matrix01}
\end{equation}
with $\mathbf{W}$ given by Equation (\ref{Eq_W_matrix01}), we can pursue the computation of the numerator of Equation (\ref{Eq_definition_limit_bag_of_paths01}),
\begin{align}
\mathbf{e}_i^{\text{T}}\left({\displaystyle \sum\limits _{t=0}^{\infty}}\mathbf{W}^{t}\right)\mathbf{e}_j
 =  \mathbf{e}_i^{\text{T}} (\mathbf{I}-\mathbf{W})^{-1} \mathbf{e}_j 
 =  \mathbf{e}_i^{\text{T}} \mathbf{Z} \mathbf{e}_j = [\mathbf{Z}]_{ij}
 =  z_{ij}
\label{Eq_W_computation01}
\end{align}
where $z_{ij}$ is element $i,j$ of $\mathbf{Z}$. By analogy with Markov chain theory,
$\mathbf{Z}$ is called the \textbf{fundamental matrix} \cite{Kemeny-1960}.
Elementwise, following Equations (\ref{Eq_numerator_bag_paths01}-\ref{Eq_W_computation01}), we have that
\begin{equation}
z_{ij} = {\displaystyle \sum_{\wp\in\mathcal{P}_{ij}}}\tilde{\pi}^{\mathrm{ref}}(\wp)\exp\left[-\theta \tilde{c}(\wp) \right] = \left[(\mathbf{I}-\mathbf{W}\mathbf{)}^{-1}\right]_{ij}
\label{Eq_fundamental_matrix_elementwise01}
\end{equation}
which is actually related to the potential of a Markov chain \cite{Cinlar-1975,Norris-1997}.
From the previous equation, $z_{ij}$ can be interpreted as
\begin{align}
z_{ij} &= {\displaystyle \sum\limits _{t=0}^{\infty}} [\mathbf{W}^{t}]_{ij} 
= \delta_{ij} + p_{ij}^{\mathrm{ref}} \, e^{-\theta c_{ij}}
+ \sum_{k_{1} = 1}^{n} p_{ik_{1}}^{\mathrm{ref}} p_{k_{1}j}^{\mathrm{ref}} \, e^{-\theta (c_{ik_{1}}+c_{k_{1}j})} \nonumber \\
&\quad + \sum_{k_{1} = 1}^{n} \sum_{k_{2} = 1}^{n} p_{ik_{1}}^{\mathrm{ref}} p_{k_{1}k_{2}}^{\mathrm{ref}} p_{k_{2}j}^{\mathrm{ref}} \, e^{-\theta c_{ik_{1}}} e^{-\theta c_{k_{1}k_{2}}} e^{-\theta c_{k_{2}j}} + \cdots
\end{align}

\noindent
For the denominator of Equation (\ref{Eq_bag_of_paths_probability01}) and (\ref{Eq_definition_limit_bag_of_paths01}), we immediately find
\begin{equation}
\mathcal{Z} = \mathbf{e}^{\text{T}} \mathbf{Z} \mathbf{e} = z_{\bullet \bullet}
\end{equation}
where $z_{\bullet \bullet} = \sum_{i,j=1}^{n} z_{ij}$ is the value of the partition function $\mathcal{Z}$.
Therefore, from Equation (\ref{Eq_definition_limit_bag_of_paths01}), the probability of drawing a path starting in $i$ and ending in $j$ in our bag-of-paths model is simply
\begin{equation}
\text{P}(s=i,e=j) = \frac{z_{ij}}{\mathcal{Z}}, \text{ with } \mathbf{Z}=(\mathbf{I}-\mathbf{W}\mathbf{)}^{-1} \text{ and } \mathcal{Z} = z_{\bullet \bullet}
\label{Eq_bag_of_paths_probabilities01}
\end{equation}
or, in matrix form,
\begin{equation}
\boldsymbol{\Pi} = \frac{\mathbf{Z}}{z_{\bullet \bullet}}, \text{ with } \mathbf{Z}=(\mathbf{I}-\mathbf{W}\mathbf{)}^{-1}
\label{Eq_bag_of_paths_probabilities02}
\end{equation}
where $\boldsymbol{\Pi}$, called the \textbf{bag-of-paths probability matrix}, contains the probabilities for each starting-ending pair of nodes. Note that this matrix is not symmetric in general; therefore, in the case of an undirected graph, we might instead compute the probability of drawing a path $i \leadsto j$ or $j \leadsto i$. The result is a symmetric matrix,
\begin{equation}
\boldsymbol{\Pi}_{\mathrm{sym}} = \boldsymbol{\Pi} + \boldsymbol{\Pi}^{\text{T}}
\label{Eq_symmetric_BoP_probability01}
\end{equation}
and only the upper (or lower) triangular part of the matrix is relevant.

\subsubsection{An intuitive interpretation of the ${z_{ij}}$}
An intuitive interpretation of the elements $z_{ij}$ of the $\mathbf{Z}$ matrix can be provided as follows \cite{Saerens-2008,Mantrach-2009}.
Consider a special random walk defined by the transition probability matrix $\mathbf{W}$ whose elements are $[\mathbf{W}]_{ij} = p^{\mathrm{ref}}_{ij} \exp[-c_{ij}]$. As $\mathbf{W}$ has some row sums less than one (the rows $i$ of $\textbf{C}$ containing at least one strictly positive cost $c_{ij}$), the random walker has a nonzero probability of disappearing in each of these nodes which is equal to $(1 - \sum_{j=1}^{n}w_{ij})$ at each time step. Indeed, from Equation (\ref{Eq_W_matrix01}), it can be observed that the probability of surviving during a transition $i \rightarrow j$ is proportional to $\exp[-\theta c_{ij}]$, which makes sense: there is a smaller probability to survive edges with a high cost. In this case, the elements of the $\mathbf{Z}$ matrix, $z_{ij} = [\mathbf{Z}]_{ij}$, can be interpreted as the expected number of times that an ``evaporating'', or ``killed'' random walk, starting from node $i$, visits node $j$ (see for instance \cite{Snell-1984,Kemeny-1960}) before being killed.

\section{Working with hitting/absorbing paths: the bag of hitting paths}
\label{Sec_bag_hitting_paths01}

The bag-of-hitting-paths model described in this section is a restriction of the previously introduced bag-of-paths model in which the ending node of each path only appears once -- at the end of the path. In other words, no intermediate node on the path is allowed to be the ending node $j$, thus prohibiting looping on this node $j$. Technically this constraint will be enforced by making the ending node \emph{absorbing}\footnote{And \emph{killing}, see later.}, as in the case of an absorbing Markov chain \cite{Snell-1984,Isaacson-1976,Kemeny-1960,Norris-1997}. We will see later in this section that this model has some nice properties.

\subsection{Definition of the bag-of-hitting-paths probabilities}

Let $\mathcal{P}^{\mathrm{h}}_{ij}$ be the set of \emph{hitting} paths starting from $i$ and stopping once node $j$ has been reached for the first time ($j$ is made absorbing). Let $\mathcal{P}^{\mathrm{h}} = \cup_{ij} \mathcal{P}^{\mathrm{h}}_{ij}$ be the complete set of such hitting paths. Following the same reasoning as in the previous subsection, from Equation (\ref{Eq_bag_of_paths_probability01}), when putting a Gibbs-Boltzmann distribution on $\mathcal{P}^{\mathrm{h}}$, the probability of drawing a hitting path starting in $i$ and ending in $j$ is

\begin{equation}
\text{P}_{\mathrm{h}}(s=i,e=j)
= \frac{{\displaystyle \sum_{\wp\in\mathcal{P}^{\mathrm{h}}_{ij}}} \tilde{\pi}^{\mathrm{ref}}(\wp)\exp\left[-\theta \tilde{c}(\wp)\right]}{{\displaystyle \sum_{\wp'\in\mathcal{P}^{\mathrm{h}}}} \tilde{\pi}^{\mathrm{ref}}(\wp')\exp\left[-\theta \tilde{c}(\wp')\right]}
= \frac{{\displaystyle \sum_{\wp\in\mathcal{P}^{\mathrm{h}}_{ij}}} \tilde{\pi}^{\mathrm{ref}}(\wp)\exp\left[-\theta \tilde{c}(\wp)\right]}{\mathcal{Z}_{\mathrm{h}}}
\label{Eq_bag_of_paths_probability_hitting01}
\end{equation}
and the denominator of this expression is also called the \textbf{partition function}, $\mathcal{Z}_{\mathrm{h}} = \sum_{\wp\in\mathcal{P}^{\mathrm{h}}} \tilde{\pi}^{\mathrm{ref}}(\wp)\exp\left[-\theta \tilde{c}(\wp)\right]$, for the hitting paths system this time. The quantity $\text{P}_{\mathrm{h}}(s=i,e=j)$ will be called the \textbf{bag-of-hitting-paths probability} of drawing a hitting path starting in $i$ and ending in $j$.
Note that in the case of unbounded hitting paths, the reference path probabilities can be simply defined as $\tilde{\text{P}}^{\mathrm{ref}} = \frac{1}{n^{2}} \tilde{\pi}^{\mathrm{ref}}$ if we assume a uniform reference probability for drawing the starting and ending nodes. With this definition, it is shown in \ref{app_hitting_paths_likelihood} that the probability is properly normalized, i.e., $\sum_{\wp\in\mathcal{P}^{\mathrm{h}}} \tilde{\text{P}}^{\mathrm{ref}}(\wp) = 1$.

Obviously, for hitting paths, if we adopt the convention that zero-length paths are allowed, paths of length greater than 0 starting in node $i$ and ending in the same node $i$ are prohibited -- in that case, the zero-length path is the only allowed path starting and ending in $i$ and we set its $\tilde{\pi}^{\mathrm{ref}}$ equal to 1.

Now, following the same reasoning as in previous section, the numerator of Equation (\ref{Eq_bag_of_paths_probability_hitting01}) is
\begin{align}
{\displaystyle \sum_{\wp\in\mathcal{P}^{\mathrm{h}}_{ij}}}\tilde{\pi}^{\mathrm{ref}}(\wp)\exp\left[-\theta \tilde{c}(\wp) \right] 
 &= \mathbf{e}_i^{\text{T}}\left({\displaystyle \sum\limits _{t=0}^{\infty}} (\mathbf{W}^{(-j)})^{t}\right)\mathbf{e}_j 
 = \mathbf{e}_i^{\text{T}} (\mathbf{I} - \mathbf{W}^{(-j)})^{-1} \mathbf{e}_j \nonumber \\
 &= \mathbf{e}_i^{\text{T}} \mathbf{Z}^{(-j)} \mathbf{e}_j = z_{ij}^{(-j)}
 \label{Eq_numerator_bag_paths_hitting01}
\end{align}
where $\mathbf{W}^{(-j)}$ is now matrix $\mathbf{W}$ of Equation (\ref{Eq_W_matrix01}) where the $j$th row has been set to $\mathbf{0}^{\text{T}}$ (node $i$ is absorbing and killing meaning that the $j$th row of the transition matrix, $\mathbf{P}^{\mathrm{ref}}$, is equal to zero) and $\mathbf{Z}^{(-j)} = (\mathbf{I} - \mathbf{W}^{(-j)})^{-1}$. This means that when the random walker reaches node $j$, he immediately stops his walk there. This matrix is given by $\mathbf{W}^{(-j)} = \mathbf{W} - \mathbf{e}_j (\mathbf{w}^{r}_j)^{\text{T}}$ with $\mathbf{w}^{r}_j = \mathbf{col}_j(\mathbf{W}^{\text{T}}) = \mathbf{W}^{\text{T}} \mathbf{e}_j$ being a column vector containing the $j$th row of $\mathbf{W}$.

\subsection{Computation of the bag-of-hitting-paths probabilities}

In \ref{app_zEntries}, it is shown from a bag-of-paths framework point of view that the elements of $\mathbf{Z}^{(-j)}$ can be computed simply and efficiently by
\begin{equation}
z_{ij}^{(-j)} = [\mathbf{Z}^{(-j)}]_{ij} = \frac{z_{ij}}{z_{jj}}
\label{Eq_hitting_paths_fundamental_matrix01}
\end{equation}
which is a noteworthy result by itself.
Note that this result has been re-derived in a more conventional, but also more tedious, way through the Sherman-Morrison formula by \cite{Kivimaki-2012} in the context of computing randomized shortest paths dissimilarities in closed form.

Using this result, Equation (\ref{Eq_numerator_bag_paths_hitting01}) can be developed as
\begin{equation}
{\displaystyle \sum_{\wp\in\mathcal{P}^{\mathrm{h}}_{ij}}} \tilde{\pi}^{\mathrm{ref}}(\wp)\exp\left[-\theta \tilde{c}(\wp) \right] 
 = z_{ij}^{(-j)} = \frac{z_{ij}}{z_{jj}} \triangleq z_{ij}^{\mathrm{h}}
 \label{Eq_backward_variable_hitting_paths01}
\end{equation}
%
where we define the matrix containing the elements $z_{ij}^{(-j)} = z_{ij}/z_{jj}$ as $\mathbf{Z}_{\mathrm{h}}$ -- the \textbf{fundamental matrix} for hitting paths. The elements of the matrix $\mathbf{Z}_{\mathrm{h}}$ are denoted by $z_{ij}^{\mathrm{h}}$. From Equation (\ref{Eq_backward_variable_hitting_paths01}), this matrix can be computed as $\mathbf{Z}_{\mathrm{h}} = \mathbf{Z}\mathbf{D}_{\mathrm{h}}^{-1}$ with $\mathbf{D}_{\mathrm{h}} = \mathbf{Diag}(\mathbf{Z})$.
Note that the diagonal elements of $\mathbf{Z}_{\mathrm{h}}$ are equal to 1, $z_{ii}^{\mathrm{h}} = 1$.
Moreover, when $\theta \rightarrow \infty$, $z_{jj} \rightarrow 1$ and $z_{ij}^{\mathrm{h}} \rightarrow z_{ij}$ (at the limit, only shortest paths, without loops, are considered).

We immediately deduce the bag-of-hitting-paths probability including zero-length paths (Equation (\ref{Eq_bag_of_paths_probability_hitting01})),
\begin{align}
\text{P}_{\mathrm{h}}(s=i,e=j)
 &= \frac{{\displaystyle \sum_{\wp\in\mathcal{P}^{\mathrm{h}}_{ij}}} \tilde{\pi}^{\mathrm{ref}}(\wp)\exp\left[-\theta \tilde{c}(\wp) \right]}{{\displaystyle \sum_{i',j'=1}^{n} \sum_{\wp' \in \mathcal{P}^{\mathrm{h}}_{i'j'}}} \tilde{\pi}^{\mathrm{ref}}(\wp')\exp\left[-\theta \tilde{c}(\wp') \right]} \nonumber \\
 &= \frac{z_{ij}/z_{jj}}{\displaystyle \sum_{i',j'=1}^{n} (z_{i'j'}/z_{j'j'})}
  = \frac{z_{ij}^{\mathrm{h}}}{\mathcal{Z}_{\mathrm{h}}}
\label{Eq_bag_of_paths_probability_hitting02}
\end{align}
where the denominator of Equation (\ref{Eq_bag_of_paths_probability_hitting02}) is the partition function of the hitting paths model,
\begin{equation}
\mathcal{Z}_{\mathrm{h}} = {\displaystyle \sum_{i,j=1}^{n} \sum_{\wp\in\mathcal{P}^{\mathrm{h}}_{ij}}} \tilde{\pi}^{\mathrm{ref}}(\wp)\exp\left[-\theta \tilde{c}(\wp) \right] = \displaystyle \sum_{i,j=1}^{n} (z_{ij}/z_{jj})
\label{Eq_bag_of_paths_hitting_partition01}
\end{equation}

In matrix form, denoting by $\boldsymbol{\Pi}_{\mathrm{h}}$ the \textbf{matrix of bag-of-hitting-paths probabilities} $\text{P}_{\mathrm{h}}(s=i,e=j)$,
\begin{equation}
\boldsymbol{\Pi}_{\mathrm{h}} = \frac{\mathbf{Z}\mathbf{D}_{\mathrm{h}}^{-1}}{\mathbf{e}^{\text{T}} \mathbf{Z}\mathbf{D}_{\mathrm{h}}^{-1} \mathbf{e}}, \text{ with } \mathbf{Z}=(\mathbf{I}-\mathbf{W}\mathbf{)}^{-1} \text{ and } \mathbf{D}_{\mathrm{h}} = \mathbf{Diag}(\mathbf{Z})
\label{Eq_bag_of_hitting_paths_probabilities02}
\end{equation}

The algorithm for computing the matrix $\boldsymbol{\Pi}_{\mathrm{h}}$ is shown in Algorithm~\ref{Alg_bag_of_hitting_paths_probability01}. The symmetric version for hitting paths is obtained by applying Equation (\ref{Eq_symmetric_BoP_probability01}) after the computation of $\boldsymbol{\Pi}_{\mathrm{h}}$.
An interesting application would be to investigate graph cuts based on bag-of-hitting-paths probabilities instead of the standard adjacency matrix.

\begin{algorithm}[t!]
\caption{\small{Computing the bag-of-hitting-paths probability matrix of a graph.}}

\algsetup{indent=2em, linenodelimiter=.}

\begin{algorithmic}[1]
\small
\REQUIRE $\,$ \\
 -- A weighted, possibly directed, strongly connected, graph $G$ containing $n$ nodes. \\
 -- The $n\times n$ adjacency matrix $\mathbf{A}$ associated to $G$, containing affinities.\\
 -- The $n\times n$ cost matrix $\mathbf{C}$ associated to $G$.\\
 -- The inverse temperature parameter $\theta$.\\
 
\ENSURE $\,$ \\
 -- The $n \times n$ bag-of-hitting-paths probability matrix $\boldsymbol{\Pi}_{\mathrm{h}}$ with zero-length paths included containing the probability of drawing a path starting in node $i$ and ending in node $j$, when sampling paths according to a Gibbs-Boltzmann distribution.\\

\STATE $\mathbf{D} \leftarrow \mathbf{Diag}(\mathbf{A}\mathbf{e})$ \COMMENT{the row-normalization, or outdegree, matrix with $\mathbf{e}$ being a column vector full of 1's} \\
\STATE $\mathbf{P}^{\mathrm{ref}} \leftarrow \mathbf{D}^{-1} \mathbf{A}$ \COMMENT{the reference transition probabilities matrix} \\
\STATE $\mathbf{W} \leftarrow \mathbf{P}^{\mathrm{ref}}\circ\exp\left[-\theta\mathbf{C}\right]$ \COMMENT{elementwise exponential and multiplication $\circ$} \\
\STATE $\mathbf{Z} \leftarrow (\mathbf{I}-\mathbf{W}\mathbf{)}^{-1}$ \COMMENT{the fundamental matrix} \\
\STATE $\mathbf{D}_{\mathrm{h}} \leftarrow \mathbf{Diag}(\mathbf{Z})$ \COMMENT{the column-normalization matrix for hitting paths probabilities} \\
\STATE $\mathbf{Z}_{\mathrm{h}} \leftarrow \mathbf{Z} \mathbf{D}_{\mathrm{h}}^{-1}$ \COMMENT{column-normalize the fundamental matrix}
\STATE $\mathcal{Z}_{\mathrm{h}} \leftarrow \mathbf{e}^{\text{T}}\mathbf{Z}_{\mathrm{h}}\mathbf{e}$ \COMMENT{compute normalization factor -- the partition function} \\
\STATE $\boldsymbol{\Pi}_{\mathrm{h}} \leftarrow \dfrac{\mathbf{Z}_{\mathrm{h}}}{\mathcal{Z}_{\mathrm{h}}}$ \COMMENT{the bag-of-hitting-paths probability matrix with zero-paths included} \\
\RETURN $\boldsymbol{\Pi}_{\mathrm{h}}$ 

\end{algorithmic} \label{Alg_bag_of_hitting_paths_probability01} 
\end{algorithm}

\subsection{An intuitive interpretation of the ${z_{ij}^{\mathrm{h}}}$} 

In this section, we provide an intuitive description of the elements of the hitting paths fundamental matrix, $\mathbf{Z}_{\mathrm{h}}$. Let us consider a particular killed random walk with absorbing state $\alpha$ on the graph $G$ whose transition probabilities are given by the elements of $\mathbf{W}^{(-j)}$, that is, $w_{ij} = p^{\mathrm{ref}}_{ij} \exp[-\theta c_{ij}]$ when $i \ne \alpha$ and $w_{\alpha j} = 0$ otherwise. In other words, the node $\alpha$ is made \emph{absorbing} and \emph{killing} -- it corresponds to hitting paths with node $\alpha$ as hitting node. When the walker reaches this node, he stops his walk and disappears. Moreover, as $\exp[-\theta c_{ij}] \le 1$ for all $i,j$, the matrix of transition probabilities $w_{ij}$ is substochastic and the random walker has also a nonzero probability $(1 - \sum_{j=1}^{n} w_{ij})$ of disappearing at each step of its random walk and in each node $i$ for which $(1 - \sum_{j=1}^{n} w_{ij}) > 0$. This stochastic process has been called an ``evaporating random walk" in \cite{Saerens-2008} or an ``exponentially killed random walk" in \cite{Steele-2001}. 

Now, let us consider column $\alpha$ (corresponding to the hitting, or absorbing, node) of the fundamental matrix of non-hitting paths, $\mathbf{col}_{\alpha}(\mathbf{Z}) = \mathbf{Z}\mathbf{e}_{\alpha}$. Because the fundamental matrix is $\mathbf{Z} = (\mathbf{I} - \mathbf{W})^{-1}$ (Equation (\ref{Eq_fundamental_matrix01})), we easily obtain $(\mathbf{I} - \mathbf{W})(\mathbf{Z}\mathbf{e}_{\alpha}) = \mathbf{I}\mathbf{e}_{\alpha} = \mathbf{e}_{\alpha}$. Or, in elementwise form,
\begin{equation}
\begin{cases}
 z_{i\alpha} = \sum_{j=1}^{n} w_{ij} z_{j\alpha} &\text{for each } i \ne \alpha\\
 z_{\alpha \alpha} = \sum_{j=1}^{n} w_{\alpha j} z_{j\alpha} + 1 &\text{for absorbing node } \alpha
\end{cases}
\label{Eq_recurrence_relations_backward01}
\end{equation}

When considering hitting paths instead, $z_{\alpha \alpha}^{\mathrm{h}} = 1$ (see Equation (\ref{Eq_backward_variable_hitting_paths01})) because $w_{\alpha j} = 0$ for all $j$ (node $\alpha$ is made absorbing and killing) so that the second line of Equation (\ref{Eq_recurrence_relations_backward01}) -- the boundary condition -- becomes simply $z_{\alpha \alpha}^{\mathrm{h}} = 1$ for hitting paths. Moreover, we know that $z_{i\alpha}^{\mathrm{h}} = z_{i\alpha}/z_{\alpha \alpha}$ for any $i \ne \alpha$. Thus, dividing the first line of Equation (\ref{Eq_recurrence_relations_backward01}) by $z_{\alpha \alpha}$ provides
\begin{equation}
\begin{cases}
 z_{i\alpha}^{\mathrm{h}} = \sum_{j=1}^{n} w_{ij} \, z_{j\alpha}^{\mathrm{h}} &\text{for each } i \ne \alpha\\
 z_{\alpha \alpha}^{\mathrm{h}} = 1 &\text{for absorbing node } \alpha
\end{cases}
\label{Eq_recurrence_relations_hitting_backward01}
\end{equation}

Interestingly, this is exactly the set of recurrence equations computing the probability of hitting node $\alpha$ when starting from node $i$ (see, e.g., \cite{Kemeny-1960,Ross-2000,Taylor-1998}). Therefore, the $z_{i\alpha}^{\mathrm{h}}$ represent the \emph{probability of surviving} during the killed random walk from $i$ to $\alpha$ with transition probabilities $w_{ij}$ and node $\alpha$ made absorbing. Said differently, it corresponds to the probability of reaching absorbing node $j$ without being killed during the walk.


\section{Two novel families of distances based on hitting path probabilities}
\label{sec:DistanceMeasures}

In this section, two families of distance measures are derived from the hitting path probabilities including zero-length paths\footnote{The results do not hold for a bag of paths excluding zero-length paths.}.
The second one benefits from some nice properties that will be detailed.

\subsection{A first distance measure}

The first distance measure is directly derived from the bag-of-paths probabilities introduced in the previous section.

\subsubsection{Definition of the distance}

This section shows that the associated surprisal measure, 
\begin{equation*}
-\log \text{P}_{\mathrm{h}}(s=i,e=j),
\end{equation*} quantifying the ``surprise" generated by the outcome $(s=i) \wedge (e=j)$, when symmetrized, is a distance measure. This distance $\surdist_{ij}$ associated to the bag-of-hitting-paths is defined as follows
\begin{equation}
\surdist_{ij} \triangleq
  \begin{cases}
   -\dfrac{\log \text{P}_{\mathrm{h}}(s=i,e=j) + \log \text{P}_{\mathrm{h}}(s=j,e=i)}{2} & \text{if } i \ne j \\
   0       & \text{if } i=j
  \end{cases}
  \label{Eq_hitting_probability_distance01}
\end{equation}
where $\text{P}_{\mathrm{h}}(s=i,e=j)$ and $\text{P}_{\mathrm{h}}(s=j,e=i)$ are computed according to Equation (\ref{Eq_bag_of_paths_probability_hitting02}) or (\ref{Eq_bag_of_hitting_paths_probabilities02}) for the matix form.
Obviously, $\surdist_{ij} \ge 0$ and $\surdist_{ij}$ is symmetric. Moreover, $\surdist_{ij}$ is equal to zero if and only if $i=j$.

It is shown in \ref{app_surprisal_dist_proof} that this quantity is a distance measure since it satisfies the triangle inequality, in addition to the other mentioned properties. This distance will be called the bag-of-hitting-paths \textbf{surprisal distance}.

\subsubsection{Computation of the distance}

It can be computed by adding the following
matrix operations to Algorithm \ref{Alg_bag_of_hitting_paths_probability01}: 
\begin{itemize}
  \item $\boldsymbol{\Delta}_{\mathrm{sur}} \leftarrow -\frac{1}{2} \left[ \log(\boldsymbol{\Pi}_{\mathrm{h}}) + \log(\boldsymbol{\Pi}_{\mathrm{h}}^{\text{T}}) \right]$ \quad $\triangleright$ take elementwise logarithm for computing the potentials
  \item $\boldsymbol{\Delta}_{\mathrm{sur}} \leftarrow \boldsymbol{\Delta}_{\mathrm{sur}} - \mathbf{Diag}(\boldsymbol{\Delta}_{\mathrm{sur}})$ \quad $\triangleright$ put diagonal to zero
\end{itemize}

\noindent We now turn to the development of the second distance measure.


%
%
%
%
%

\subsection{A second distance measure}
\label{Subsec_second_distance01}

This subsection introduces a second measure enjoying some nice properties, based on the same ideas.

\subsubsection{Definition of the distance}

The second distance measure automatically follows from Inequality (\ref{Eq_bag_of_paths_probability_inequality04}) in \ref{app_surprisal_dist_proof} and is based on the quantity $\phi(i,j) = -\frac{1}{\theta} \log z_{ij}^{\mathrm{h}}$.
For convenience, let us recall this inequality,
\begin{equation}
\text{P}_{\mathrm{h}}(s=i,e=k) \ge \mathcal{Z}_{\mathrm{h}} \text{P}_{\mathrm{h}}(s=i,e=j) \, \text{P}_{\mathrm{h}}(s=j,e=k) \nonumber \\
\end{equation}

Then, from $\text{P}_{\mathrm{h}}(s=i,e=j) = z_{ij}^{\mathrm{h}}/\mathcal{Z}_{\mathrm{h}}$ (Equation (\ref{Eq_bag_of_paths_probability_hitting02})), we directly obtain $z_{ik}^{\mathrm{h}} \ge z_{ij}^{\mathrm{h}} \, z_{jk}^{\mathrm{h}}$. Taking $-\frac{1}{\theta} \log$ of both sides provides $-\frac{1}{\theta}\log z_{ik}^{\mathrm{h}} \le -\frac{1}{\theta}\log z_{ij}^{\mathrm{h}} -\frac{1}{\theta}\log z_{jk}^{\mathrm{h}}$, or,
\begin{equation}
\phi(i,k) \le \phi(i,j) + \phi(j,k)
\label{Eq_potential_function_inequality01}
\end{equation}
where we defined
\begin{equation}
\phi(i,j) \triangleq -\frac{1}{\theta} \log z_{ij}^{\mathrm{h}}
= -\frac{1}{\theta} \log \left(\dfrac{z_{ij}}{z_{jj}} \right)
\label{Eq_potential_function_definition01}
\end{equation}
and, from (\ref{Eq_potential_function_inequality01}), the $\phi(i,j)$ obviously verify the triangle inequality.


The quantity $\phi(i,j)$ will be called the \emph{potential} \cite{Cinlar-1975} of node $i$ with respect to node $j$. Indeed, it has been shown \cite{Garcia-Diez-2011b} that when computing the continuous-state continuous-time equivalent of the randomized shortest paths framework \cite{Saerens-2008}, $\phi(x,y)$ plays the role of a potential inducing a drift (external force) $\boldsymbol{\nabla} \phi$ in the corresponding diffusion equation. From the properties and the probabilistic interpretation of the $z_{ij}^{\mathrm{h}}$, both $\phi(i,j) \ge 0$ (as $0 \le z_{ij}^{\mathrm{h}} \le 1$) and $\phi(i,i)=0$ (as $z_{ij}^{\mathrm{h}} = 1$) hold.

This directed distance measure has three intuitive interpretations.
\begin{itemize}
  \item First, let us recall from Equation (\ref{Eq_backward_variable_hitting_paths01}) that $z_{ij}^{\mathrm{h}}$ is given by $z_{ij}^{\mathrm{h}} =$ ${\sum_{\wp\in\mathcal{P}^{\mathrm{h}}_{ij}}} \tilde{\pi}^{\mathrm{ref}}(\wp)\exp\left[-\theta \tilde{c}(\wp)\right]$ $= z_{ij}/z_{jj}$
where $z_{ij}$ is element $i,j$ of the fundamental matrix $\mathbf{Z}$ (see Equation (\ref{Eq_fundamental_matrix01})). From this last expression, $\phi(i,j)$ can be interpreted (up to a scaling factor) as the logarithm of the expectation of the reward $\exp[-\theta \tilde{c}(\wp)]$
with respect to the path likelihoods, when considering absorbing random walks starting from node $i$ and ending in node $j$.
  \item In addition, from Equation (\ref{Eq_recurrence_relations_hitting_backward01}), it also corresponds to minus the log-likelihood of surviving during the killed, absorbing, random walk from $i$ to $j$.
  \item Finally, it was shown in \cite{Kivimaki-2012}, investigating further developments of the randomized shortest paths (RSP) dissimilarity, that the potential distance also corresponds to the minimal free energy of the system of hitting paths from $i$ to $j$. Indeed, the RSP dissimilarity, defined as the expected total cost between $i$ and $j$, is not a distance measure as it does not satisfy the triangle inequality. However, subtracting the entropy from the expected total cost (that is, computing the free energy) leads to a distance measure that was shown to be equivalent to the potential distance. Therefore the potential distance was called the \textbf{free energy distance} in \cite{Kivimaki-2012}, which provides still another interpretation to the potential distance.
\end{itemize}

Inequality (\ref{Eq_potential_function_inequality01}) suggests to define the distance $\dist^{\phi}_{ij} = (\phi(i,j) + \phi(j,i))/2$. It has all the properties of a distance measure, including the triangle inequality, which is verified thanks to Inequality (\ref{Eq_potential_function_inequality01}).
Note that this distance measure can be expressed as a function of the surprisal distance (see Equation (\ref{Eq_hitting_probability_distance01})) as
$\dist^{\phi}_{ij} = (\surdist_{ij} - \log \mathcal{Z}_{\mathrm{h}})/\theta$ for $i \ne j$. This shows that the newly
introduced distance is equivalent to the previous one, up to the addition of a constant and a rescaling.

The definition of the bag-of-hitting-paths \textbf{potential distance} is therefore
\begin{equation}
\dist^{\phi}_{ij} \triangleq
  \begin{cases}
   \dfrac{\phi(i,j) + \phi(j,i)}{2} & \text{if } i \ne j \\
   0       & \text{if } i=j
  \end{cases}, \text{ where } \phi(i,j) = -\frac{1}{\theta} \log \left(\dfrac{z_{ij}}{z_{jj}} \right)
\label{Eq_bag_of_paths_potential_distance01}
\end{equation}
and $z_{ij}$ is element $i,j$ of the fundamental matrix $\mathbf{Z}$ (see Equation (\ref{Eq_fundamental_matrix01})).

\subsubsection{Computation of the distance}

From Equation (\ref{Eq_bag_of_hitting_paths_probabilities02}), it can be
easily seen that the matrix $\mathbf{Z}_{\mathrm{h}}$ containing the
$z_{ij}^{\mathrm{h}}$ can be computed thanks to Algorithm
\ref{Alg_bag_of_hitting_paths_probability01} without the normalization steps 7
and 8. The distance matrix with elements $\surdist_{ij}$ is denoted
as $\boldsymbol{\Delta}_{\mathrm{sur}}$ and can easily be obtained by adding the following
matrix operations to Algorithm \ref{Alg_bag_of_hitting_paths_probability01}:
\begin{itemize}
  \item $\boldsymbol{\Phi} \leftarrow -\log(\mathbf{Z}_{\mathrm{h}})/ \theta$ \quad $\triangleright$ take elementwise logarithm for computing the potentials
  \item $\boldsymbol{\Delta}_{\phi} \leftarrow (\boldsymbol{\Phi} + \boldsymbol{\Phi}^{\text{T}})/2$ \quad $\triangleright$ symmetrize the matrix
  \item $\boldsymbol{\Delta}_{\phi} \leftarrow \boldsymbol{\Delta}_{\phi} - \mathbf{Diag}(\boldsymbol{\Delta}_{\phi})$ \quad $\triangleright$ put diagonal to zero
\end{itemize}
Note that both the surprisal and the potential distances are well-defined as we assumed that $G$ is strongly connected.

\subsection{Some properties of the potential and surprisal distances}

The potential distance $\dist^{\phi}$ benefits from some interesting properties proved in the appendix:
\begin{itemize}
  \item \textbf{The potential distance is graph-geodetic}, meaning that $\dist^{\phi}_{ik} = \dist^{\phi}_{ij} + \dist^{\phi}_{jk}$ if and only if every path from $i$ to $k$ passes through $j$ \cite{Chebotarev-2011} (see \ref{app_geodetic} for the proof). 

  \item \textbf{For an undirected graph $G$, the distance $\dist^{\phi}_{ij}$ approaches the shortest path distance} when $\theta$ becomes large, $\theta \rightarrow \infty$. In that case, the Equation (\ref{Eq_bag_of_paths_potential_distance01}) reduces to the Bellman-Ford formula (see, e.g., \cite{Bertsekas-2000,Christofides_1975,Cormen-2009}) for computing the shortest path distance, $\dist^{\mathrm{SP}}_{ik} = \min_{j \in \mathcal{S}ucc(i)} \{ c_{ij} + \dist^{\mathrm{SP}}_{jk} \}$ and $\dist^{\mathrm{SP}}_{kk} = 0$ (see \ref{app_shortestPathDist} for the proof). 
  The convergence is, however,
 slow\footnote{It was observed, e.g., that the convergence of the RSP dissimilarity is much faster when $\theta$ increases.} and numerical underflows could appear before complete convergence to the shortest path distances (convergence is linear in $\theta$ -- see the appendix for details). Therefore, if solutions close to the shortest paths distance are needed (with very large $\theta$), computational tricks such as those used in hidden Markov models should be implemented. See for instance the appendix in \cite{Huang-1990}.
  \item \textbf{For an undirected graph $G$, the distance $\dist^{\phi}_{ij}$
      approaches half the commute cost distance} when $\theta$ becomes small,
    $\theta \rightarrow 0^{+}$ (see \ref{app_commuteCostDist} for the
    proof).
    Note that, for a given graph $G$, the commute cost between two nodes is proportional to the
    commute time between these two nodes, and therefore also proportional to
    the resistance distance (see \cite{Chandra-1989,Kivimaki-2012}).
    
    \item \textbf{The distance $\dist^{\phi}_{ij}$ extends the Bellman-Ford formula computing the shortest path distance to integrate sub-optimal paths (exploration)} by simply replacing the $\min$ operator by the $\mathrm{softmin}$ operator in the recurrence formula. This property is discussed in the next subsection.
\end{itemize}

All of these properties make the potential distance quite attractive as it defines a family of distances interpolating between the shortest path and the resistance distance. Our conjecture is that interpolating between these two distances hopefully alleviates the ``lost in space" effect \cite{vonLuxburg-2010} as the distance gradually focuses on shorter paths, while still exploring sub-optimal paths, when parameter $\theta$ increases. A recent paper \cite{Hashimoto-2015} addresses this question by showing the consistency and the robustness of the Laplacian transformed hitting time (the Laplace transform of hitting times), a measure related to the potential distance. One of our future work will be to evaluate if their analysis can be transposed to our measures.
But, of course, ultimately, the ``best" distance is application- and data-dependent and it is difficult to know in advance which one will perform best. 

Note that, even if the potential distance converges to the commute cost when $\theta \rightarrow 0^{+}$, we have to stress that $\theta$ should not become equal to zero because the matrix $\mathbf{W}$ becomes rank-deficient when $\theta = 0$. This means that the Equation (\ref{Eq_fundamental_matrix01}) cannot be used for computing the commute cost when $\theta$ is \emph{exactly} equal to zero. Despite this annoying fact, we found that the approximation is quite accurate for small values of $\theta$.

Concerning the surprisal distance, because it was shown in the previous section that $\surdist_{ij} = \theta\dist^{\phi}_{ij} + \log \mathcal{Z}_{\mathrm{h}}$ for all $i \ne j$, we deduce that the ranking of the node distances for a given $\theta$ is the same for the two distances.

\subsection{Relationships with the Bellman-Ford formula}

As shown in \ref{app_shortestPathDist}, Equation (\ref{Eq_recurrence_formula_potential01}), the potential $\phi(i,k)$ for a fixed ending node $k$ can be computed thanks to the following recurrence formula
\begin{equation}
\phi(i,k) =
  \begin{cases}
   -\dfrac{1}{\theta} \log \left[ {\displaystyle \sum_{j \in \mathcal{S}ucc(i)}} p_{ij}^{\mathrm{ref}} \exp[-\theta (c_{ij} + \phi(j,k))] \right] & \text{if } i \ne k \\
   0       & \text{if } i=k
  \end{cases}
\label{Eq_potential_recurrence_formula01}
\end{equation}
which is an extension of Bellman-Ford's formula for computing the shortest path distance in a graph \cite{Bertsekas-2000,Christofides_1975,Cormen-2009,Jungnickel-2005,Rardin-1998,Sedgewick-2011}.
The Equation (\ref{Eq_potential_recurrence_formula01}) has to be iterated until convergence.
Note that this result is related to the concept of ``path integral control" developed in control theory; see, e.g., the survey \cite{Kappen-2007}.

Interestingly and intriguingly, this expression is obtained by simply replacing the $\min$ operator by a weighted version of the $\mathrm{softmin}$ operator \cite{Cook-2011} in the Bellman-Ford recurrence formula,
\begin{equation}
\mathrm{softmin}_{\mathbf{q},\theta}(\mathbf{x}) = -\frac{1}{\theta} \log\bigg( \sum_{j=1}^{n} q_{j} \exp[-\theta x_{j}] \bigg) \text{ with all } q_{j} \ge 0 \text{ and } {\textstyle \sum_{j=1}^{n} q_{j}=1}
\end{equation}
which interpolates between weighted average and minimum operators (see \ref{app_shortestPathDist} or \cite{Cook-2011,Tahbaz-2006}).
Indeed, the potential $\phi(i,j)$ tends to the average first-passage cost when $\theta \rightarrow 0^{+}$ and to the shortest path cost when $\theta \rightarrow \infty$. This formula is a generalization of the distributed consensus algorithm developed in \cite{Tahbaz-2006}, and considering binary costs only.



\section{Extending the bag of paths by considering non-uniform priors on nodes}
\label{Eq_non_uniform_priors01}

This section extends the bag of hitting paths model by considering non-uniform a priori probabilities of selecting the starting and ending nodes\footnote{The development for non-hitting paths is similar and will therefore be omitted.}. For instance, if the nodes represent cities, it could be natural to weigh each city by its population. These prior probabilities, weighting each node of $G$, will be denoted as $q_{i}^{\mathrm{s}}$ and $q_{j}^{\mathrm{e}}$ with $\sum_{i=1}^{n} q_{i}^{\mathrm{s}} = 1$, $\sum_{j=1}^{n} q_{j}^{\mathrm{e}} = 1$ and all weights non-negative.


In this situation, because the reference probability $\tilde{\text{P}}^{\mathrm{ref}}(\wp_{ij})$ becomes
\begin{equation}
\tilde{\text{P}}^{\mathrm{ref}}(\wp_{ij})
= q_{i}^{\mathrm{s}} q_{j}^{\mathrm{e}} \, \tilde{\pi}^{\mathrm{ref}}(\wp_{ij}) ,
\end{equation}
instead of $\frac{1}{n^2} \tilde{\pi}^{\mathrm{ref}}(\wp_{ij})$, the probability of sampling a hitting path $i,j$ in Equation (\ref{Eq_bag_of_paths_probability_hitting01}) is redefined as
\begin{align}
\text{P}_{\mathrm{h}}(s=i,e=j)
= &\frac{{\displaystyle \sum_{\wp\in\mathcal{P}_{ij}^{\mathrm{h}}}}\tilde{\text{P}}^{\mathrm{ref}}(\wp)\exp\left[-\theta \tilde{c}(\wp) \right]}{{\displaystyle \sum_{\wp' \in\mathcal{P}^{\mathrm{h}}}} \tilde{\text{P}}^{\mathrm{ref}}(\wp')\exp\left[-\theta \tilde{c}(\wp') \right]} \nonumber \\
= &\frac{ q_{i}^{\mathrm{s}} \left( {\displaystyle \sum_{\wp\in\mathcal{P}^{\mathrm{h}}_{ij}} }\tilde{\pi}^{\mathrm{ref}}(\wp)\exp\left[-\theta \tilde{c}(\wp) \right] \right) q_{j}^{\mathrm{e}}}{ {\displaystyle \sum_{i',j'=1}^{n}} q_{i'}^{\mathrm{s}} \left( {\displaystyle  \sum_{\wp' \in\mathcal{P}^{\mathrm{h}}_{i'j'}} } \tilde{\pi}^{\mathrm{ref}}(\wp')\exp\left[-\theta \tilde{c}(\wp') \right] \right) q_{j'}^{\mathrm{e}} }
\label{Eq_bag_of_paths_probability_prior01}
\end{align}
where $\tilde{\pi}^{\mathrm{ref}}(\wp_{ij})$ is, as before, the likelihood of the path $\wp_{ij}$ given that the starting and ending nodes are $i$, $j$. Therefore this expression can be computed thanks to Equation (\ref{Eq_backward_variable_hitting_paths01}) as the \emph{weighted} quantity
\begin{equation}
\text{P}_{\mathrm{h}}(s=i,e=j)
= \frac{ q_{i}^{\mathrm{s}} \left( \dfrac{z_{ij}}{z_{jj}} \right) q_{j}^{\mathrm{e}}}{ {\displaystyle \sum_{i',j'=1}^{n}} q_{i'}^{\mathrm{s}} \left( \dfrac{z_{i'j'}}{z_{j'j'}} \right) q_{j'}^{\mathrm{e}} }
= \frac{ q_{i}^{\mathrm{s}} \left( \dfrac{z_{ij}}{z_{jj}} \right) q_{j}^{\mathrm{e}}}{ \mathcal{Z}_{\mathrm{hw}} }
 \label{Eq_bag_of_paths_probability_hitting_prior02}
\end{equation}
and the denominator
\begin{equation}
\mathcal{Z}_{\mathrm{hw}} = \sum_{i,j=1}^{n} q_{i}^{\mathrm{s}} \, \left( \frac{z_{ij}}{z_{jj}} \right) \, q_{j}^{\mathrm{e}}
\label{Eq_partition_function_prior01}
\end{equation}
is the new, weighted by priors, partition function.
The numerator of (\ref{Eq_bag_of_paths_probability_hitting_prior02}) is the fundamental matrix of the hitting paths system for weighted nodes, containing elements
\begin{equation}
q_{i}^{\mathrm{s}} \, \left( \frac{z_{ij}}{z_{jj}} \right) \, q_{j}^{\mathrm{e}}
= q_{i}^{\mathrm{s}} \, z_{ij}^{\mathrm{h}} \, q_{j}^{\mathrm{e}} \quad \text{where, as before, } z_{ij}^{\mathrm{h}} = \frac{z_{ij}}{z_{jj}}
\label{Eq_fundamental_matrix_prior01}
\end{equation}

In matrix form, the counterpart of Equation (\ref{Eq_bag_of_hitting_paths_probabilities02}) -- but now including priors on the nodes -- is
\begin{equation}
\boldsymbol{\Pi}_{\mathrm{h}} = \frac{ \mathbf{Diag}(\mathbf{q}_{\mathrm{s}}) \mathbf{Z}\mathbf{D}_{\mathrm{h}}^{-1} \mathbf{Diag}(\mathbf{q}_{\mathrm{e}}) }{\mathbf{\mathbf{q}}^{\text{T}}_{\mathrm{s}} \mathbf{Z}\mathbf{D}_{\mathrm{h}}^{-1} \mathbf{q}_{\mathrm{e}} }, \text{ with } \mathbf{D}_{\mathrm{h}} = \mathbf{Diag}(\mathbf{Z})
\label{Eq_bag_of_hitting_paths_probabilities_prior02}
\end{equation}
where the vectors $\mathbf{q}_{\mathrm{s}}$ and $\mathbf{q}_{\mathrm{e}}$ contain the a priori probabilities $q_{i}^{\mathrm{s}}$ and $q_{i}^{\mathrm{e}}$. Of course, we recover Equation (\ref{Eq_bag_of_hitting_paths_probabilities02}) when $\mathbf{q}_{\mathrm{s}} = \mathbf{q}_{\mathrm{e}} = \mathbf{e}/n$.

Interestingly, the surprisal and potential distances defined on the weighted nodes still verify the triangle inequality and are therefore distance measures; this is shown in \ref{Sec_triangle_inequality_prior01}. 
Therefore, both the surprisal and the potential distances are defined in the same way as in previous section (see Equations (\ref{Eq_hitting_probability_distance01}) and (\ref{Eq_bag_of_paths_potential_distance01})), but based this time on the weighted quantities defined in Equations (\ref{Eq_bag_of_paths_probability_hitting_prior02}) and (\ref{Eq_fundamental_matrix_prior01}). More precisely, the directed surprisal distance is computed by taking $-\log$ of the probabilities (\ref{Eq_bag_of_paths_probability_hitting_prior02}) or (\ref{Eq_bag_of_hitting_paths_probabilities_prior02}) (matrix form) while the directed potential distance is redefined as $\phi(i,j) = -\frac{1}{\theta} \log (q_{i}^{\mathrm{s}} z_{ij}^{\mathrm{h}} q_{j}^{\mathrm{e}})$ (see \ref{Sec_triangle_inequality_prior01} for details).

\section{Experiments on semi-supervised classification tasks}
\label{sec:Exp}

This experimental section aims at investigating the potential of the bag-of-hitting-paths distances and kernels derived from them in a semi-supervised classification task, on which they are compared with other competitive techniques.

Notice, however, that the goal of this experiment
is not to design a state-of-the-art classifier. Rather, the main objective
is to study the performances of the proposed measures in comparison with other measures and therefore investigate their usefulness in solving pattern recognition tasks. More precisely, this experiment investigates to which extent the distance measures are able to accurately capture the global structure of the graph through a spectral method.

\begin{table}
\begin{center}
\scalebox{0.75}{%
\begin{tabular}{l r l r l r}
	Topic & Size & Topic & Size & Topic & Size \\
	\hline
	news-2cl-1 & & news-2cl-2 & & news-2cl-3\\
	Politics/general	& 200 & Computer/graphics & 200 & Space/general & 200\\
	Sport/baseball & 200 & Motor/motorcycles & 200  & Politics/mideast & 200\\
	\\
	news-3cl-1 & & news-3cl-2 & & news-3cl-3\\
	Sport/baseball & 200 & Computer/windows & 200 & Sport/hockey & 200\\
	Space/general & 200 & Motor/autos & 200 & Religion/atheism & 200\\
	Politics/mideast & 200 & Religion/general & 200 & Medicine/general & 200\\
	\\
	news-5cl-1 & & news-5cl-2 & & news-5cl-3\\
	Computer/windowsx & 200 & Computer/graphics & 200 & Computer/machardware & 200\\
	Cryptography/general & 200 & Computer/pchardware & 200 & Sport/hockey & 200\\
	Politics/mideast & 200 & Motor/autos & 200 & Medicine/general & 200\\
	Politics/guns & 200 & Religion/atheism & 200 & Religion/general & 200\\
	Religion/christian & 200 & Politics/mideast & 200 & Forsale/general & 200\\
\end{tabular}}
\caption{Document subsets for semi-supervised classification experiments. Nine subsets have been extracted from the original 20 Newsgroups dataset, with 2, 3 and 5 topics as proposed in \cite{Yen-2008}. Each class is composed of 200 documents.}\label{newsgroup_table}
\end{center}
\end{table}

\subsection{Graph based semi-supervised classification}
\label{sec:SemiSupervised}
Semi-supervised graph node classification has received an increasing interest in recent years (see \cite{Abney-2008,Chapelle-2006,Hofmann-2008,Zhu-2008,Zhu-2009} for surveys). It considers the task of using the graph structure and other available information for inferring the class labels of unlabeled nodes of a network in which only a part of the class labels of nodes are known a priori. Several categories of approaches have been suggested for this problem. Among them, we may mention random walks \cite{Zhou-04,Szummer-01,Callut-2008}, graph mincuts \cite{Blum-2001}, spectral methods \cite{Chapelle-2002,Smola-03,Kondor-2002,Kapoor-2005}, regularization frameworks \cite{Belkin-2004,Wang-2009,Yajima-2006,Zhou-2003,Zhou-05}, transductive and spectral SVMs \cite{Joachims03}, to name a few.

Still another family of approaches is based on kernel methods, which embed the nodes of the input graph into a Euclidean feature space where a decision boundary can be estimated using standard kernel (semi-)supervised methods, such as SVMs. Fouss et al. \cite{FoussKernelNN-2011} investigated the applicability of nine such graph kernels in collaborative recommendation and semi-supervised classification by adopting a simple sum-of-similarities\footnote{The equivalent of nearest neighbors classification when dealing with similarities (a kernel matrix) instead of distances.} rule (SoS). Zhang et al. \cite{Zhang-2008,Zhang-2008b} as well as Tang et al. \cite{Tang-2009,Tang-2009b,Tang-2010} extract the dominant eigenvectors (a latent space) of graph kernels or similarity matrices and then input them to a supervised classification method, such as a logistic regression or a SVM, to categorize the nodes. These techniques based on similarities and eigenvectors extraction allow to scale to large graphs, depending on the kernel.


Another category of classification methods relies on random walks performed on a weighted and possibly directed graph seen as a Markov chain. The random walk with restart \cite{Pan-2004,Tong-2006,Tong-2007}, directly inspired by the PageRank algorithm, is one of them.
The method of Callut et al.~\cite{Callut-2008}, based on discriminative random walks, or $\D$-walks, belongs to the same category. It defines, for each class, a group betweenness measure based on passage times during special random walks of bounded length. Those walks are constrained to start and end in nodes within the same class, defining distinct random walks for each class. The number of passages on nodes is computed for each type of such random walk, therefore defining a distinct betweenness for each class. The main advantage of some of these random walk based approaches is that class labels can be computed efficiently (in linear time) while providing competitive results.

\subsection{Datasets description}
\label{sec:Data}
Comparison of the different methods will be performed on several well-known real world graph datasets (14 in total). Note that, in some cases, only the largest connected components of the following graphs have been selected:
\begin{itemize}
\item \textbf{20 Newsgroups (9 subsets)}:  This dataset\footnote{Available, e.g., from http://people.csail.mit.edu/jrennie/20Newsgroups/.} is composed of 20000 text documents taken from 20 discussion groups of the Usenet diffusion list (available on UCI \cite{Lichman-2013}). Nine subsets related to different topics are extracted from the original dataset, as listed in Table \ref{newsgroup_table} \cite{Yen-2008}. Each subset is composed of 200 documents extracted randomly from the different newsgroups. The subsets with two classes (news-2cl-1,2,3) contain 400 documents, 200 in each class.  Identically, subsets with three classes contain 600 documents and subsets with five classes contain 1000 documents. Each subset is composed of different topics, each of which are either easy to separate (Computer/windowsx and Religion/christian) or harder to separate (Computer/graphics and Computer/pchardware). Initially, this dataset does not have a graph structure but is represented in a word vector space of high dimensionality.  To transform this dataset into a graph structure, a fairly standard preprocessing has been performed, which is directly inspired by the paper of Yen et al. \cite{Yen-2008}. 

Basically, the first step is to reduce the high dimensionality of the feature space (terms), by removing stop words, applying a stemming algorithm on each term, removing too common or uncommon terms and by removing terms with low mutual information with documents.  
Second, a term-document matrix $\mathbf{W}$ is constructed with the remaining terms and documents. The elements $w_{ij}$ are \textit{tf-idf} values \cite{Manning-2008} of term $i$ in document $j$. Each row of the term-document matrix $\mathbf{W}$ is then normalized to 1. Finally, the adjacency matrix defining the links between documents is given by 
$\mathbf{A} = \mathbf{W^{T}W}$.

\item \textbf{IMDB}: The collaborative Internet Movie Database ({\tt IMDb}, \cite{Macskassy-07}) has several applications such as making movie recommendations, clustering or movie category classification.  It contains a graph of movies linked together whenever they share the same production company. The weight of an edge in the resulting graph is the number of production companies two movies have in common. The classification problem focuses on identifying clusters of movies that share the same notoriety (whether the movie is a box-office hit or not).

\item \textbf{WebKB (4 datasets)}: These networks consist of sets of web pages gathered from four computer science departments (one for each university, \cite{Macskassy-07}), with each page manually labeled into 6 categories: course, department, faculty, project, staff, and student. Two pages are linked by co-citation (if $x$ links to $z$ and $y$ links to $z$, then $x$ and $y$ are co-citing $z$).


\end{itemize}

The adjacency matrices provided by these datasets are all undirected and some are weighted. In a standard way, the costs associated to the edges are set to $c_{ij} = 1/a_{ij}$. That is, the elements of the adjacency matrix are considered as conductances and the costs as resistances. For unweighted graphs, affinities and costs are both equal to 1 for existing edges, meaning that the paths are weighted by their total length (number of steps).

\subsection{Compared distances, kernels, and algorithms}\label{introducedKernels}

This paper derived distance measures from the bag-of-paths probabilities.  In order to use these distances in machine learning and pattern recognition methods, it is convenient to transform them into similarity matrices, simply called kernels for convenience.

\subsubsection{Deriving a kernel from a distance}
 
From classical multidimensional scaling (MDS, see, e.g., \cite{Borg-1997,Cox-2001}), a centered kernel matrix $\mathbf{K}$ can be derived from a matrix of squared distances $\boldsymbol{\Delta}^{(2)}$ as follows
\begin{equation}
\mathbf{K}^{\mathrm{mds}} = -\frac{1}{2} \mathbf{H}\boldsymbol{\Delta}^{(2)}\mathbf{H}
\label{Eq_centering}
\end{equation}
where $\mathbf{H}=(\mathbf{I}-\mathbf{ee}^T/n)$ is the centering matrix and matrix $\boldsymbol{\Delta}^{(2)}$ contains the elementwise squared distances. Then, computing the dominant eigenvectors of this matrix (see the next section on Experimental settings) corresponds exactly to classical multidimensional scaling.

Still another popular way to map the distance matrix to a kernel matrix aims to use the Gaussian mapping or kernel (see, e.g., \cite{Scholkopf-2002})
\begin{equation}
\mathbf{K}^{\mathrm{g}} = \exp \left[- \boldsymbol{\Delta}^{(2)}/2 \sigma^{2} \right]
\label{Eq_gaussianKernel01}
\end{equation}
where the exponential is taken elementwise. Both approaches will be investigated. Computing the dominant eigenvectors of this matrix corresponds to a kernel principal components analysis \cite{Scholkopf-2002,Scholkopf-1998}

However, the obtained kernels are not necessarily positive semi-definite until the distance is Euclidean, which is required for kernel methods. This problem can be fixed by removing the negative eigenvalues (see, e.g., \cite{Mardia-1979}), which will be applied in all our experiments\footnote{Note that, probably because only the dominant eigenvectors are extracted, we did not observe any significant difference in the experimental results when removing and not removing the negative eigenvalues of the kernels (results not reported).}.

For classifying the nodes, the five dominant eigenvectors of the resulting kernels will be extracted from these kernels and then injected into a SVM classifier (see the next Subsection \ref{Subsec_experimental_settings} for details).

\subsubsection{Compared methods}

The following list presents the methods based on kernels computed from the distances introduced in this paper, as well as from two other recent families of dissimilarities, for comparison. The derived kernels are computed by using both (1) multidimensional scaling (\textbf{mds}, Equation (\ref{Eq_centering})) and (2) a Gaussian kernel (\textbf{g}, Equation (\ref{Eq_gaussianKernel01})).
\begin{itemize}
\item The kernels associated to the bag-of-hitting-paths potential distance ($\mathbf{K}_\text{BoPP}^{\mathrm{mds}}$, $\mathbf{K}_\text{BoPP}^{\mathrm{g}}$)  (Equations (\ref{Eq_bag_of_paths_potential_distance01}) and (\ref{Eq_centering})-(\ref{Eq_gaussianKernel01})). The corresponding methods are denoted as \textbf{BoPP-mds} and \textbf{BoPP-g}. 
\item The kernels associated to the bag-of-hitting-paths surprisal distance ($\mathbf{K}_\text{BoPS}^{\mathrm{mds}}$, $\mathbf{K}_\text{BoPS}^{\mathrm{g}}$) (Equations (\ref{Eq_hitting_probability_distance01})) and (\ref{Eq_centering})-(\ref{Eq_gaussianKernel01})). The corresponding methods are denoted as \textbf{BoPS-mds} and \textbf{BoPS-g}.
\item The randomized shortest path (RSP) kernel ($\mathbf{K}_\text{RSP}^{\mathrm{mds}}$, $\mathbf{K}_\text{RSP}^{\mathrm{g}}$) computed from the RSP dissimilarity (see \cite{Kivimaki-2012,Yen-08K,Saerens-2008} and Equations (\ref{Eq_centering})-(\ref{Eq_gaussianKernel01})). The corresponding methods are denoted as \textbf{RSP-mds} and \textbf{RSP-g}.
\item The logarithmic forest (LF) kernel ($\mathbf{K}_\text{LF}^{\mathrm{mds}}$, $\mathbf{K}_\text{LF}^{\mathrm{g}}$) computed from the logarithmic forest distance (see \cite{Chebotarev-2011,Chebotarev-2012} and Equations (\ref{Eq_centering})-(\ref{Eq_gaussianKernel01})). The corresponding methods are denoted as \textbf{LF-mds} and \textbf{LF-g}.
\end{itemize}

In addition, five state-of-the-art similarity matrices and kernels on a graph are added to this list and compared to the previous ones. We selected the three kernels providing consistently the best results in \cite{FoussKernelNN-2011}, which were based on a sum-of-similarities instead of the spectral method investigated in this paper.
\begin{itemize}
\item The modularity matrix ($\mathbf{Q}$)~\cite{Newman-2006,Newman-2010}, which was used as a kernel for semi-supervised learning earlier by Zhang et al. \cite{Zhang-2008,Zhang-2008b} as well as Tang et al. \cite{Tang-2009,Tang-2009b,Tang-2010}. The modularity matrix performed best in their experiments, in comparison with other state-of-the-art methods. This is our \emph{first baseline} method, denoted as \textbf{Q}.
\item The Markov diffusion kernel ($\mathbf{K}_{\text{MD}}$) \cite{FoussKernelNN-2011} computed from the Markov diffusion map distance \cite{Nadler-2005,Nadler-2006} and studied in \cite{Yen-2011,FoussKernelNN-2011}. This kernel, as well as the two following ones, provided good results in \cite{FoussKernelNN-2011}. The corresponding method is denoted as \textbf{MD}.
\item The regularized Laplacian, or matrix forest, kernel ($\mathbf{K}_{\text{RL}}$) \cite{Ito-2005,Chebotarev-1997,Chebotarev-1998a,FoussKernelNN-2011}. The corresponding method is denoted as \textbf{RL}.
\item The regularized commute time kernel ($\mathbf{K}_{\text{RCT}}$) \cite{FoussKernelNN-2011,Mantrach-2011}. The corresponding method is denoted as \textbf{RCT}.
\item The bag-of-paths modularity matrix ($\mathbf{K}_{\text{BoPM}}$) studied in \cite{Devooght-2014}. The corresponding method is denoted as \textbf{BoPM}.
\end{itemize}

Finally, our introduced distances are also compared to an efficient, alternative, way of performing semi-supervised classification on a network:
\begin{itemize}
\item A sum-of-similarities (SoS) algorithm based on the regularized commute time kernel, which provided good results on large datasets in \cite{Mantrach-2011}; see this paper for details. This is our \emph{second baseline} method, denoted as \textbf{SoS}.
\end{itemize}

\begin{figure}[t!]
\begin{center}
\begin{tabular}{ccc}
\includegraphics[width=0.3\textwidth]{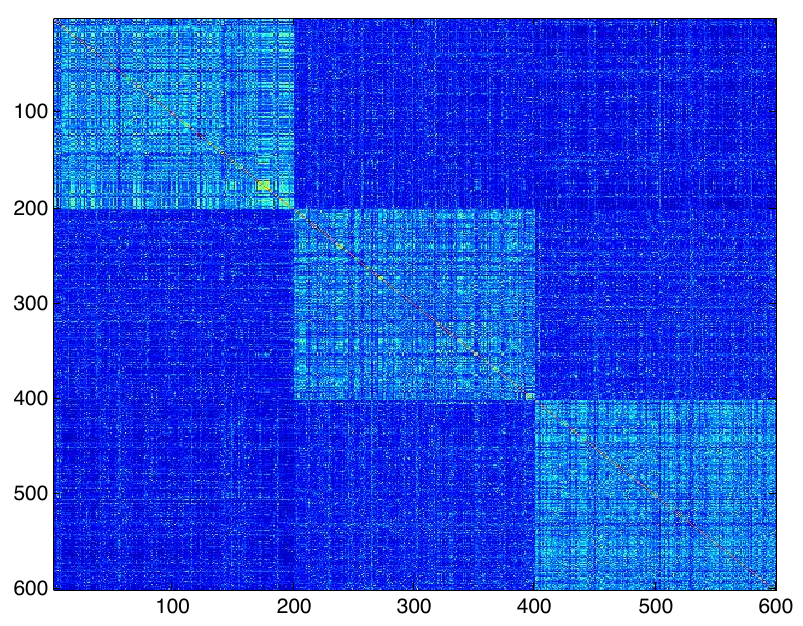} & \includegraphics[width=0.3\textwidth]{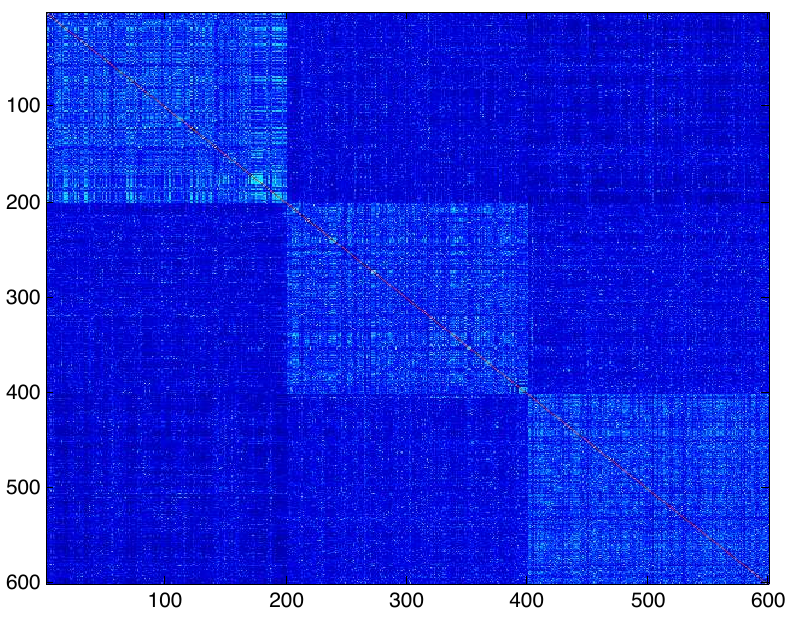} &
 \includegraphics[width=0.3\textwidth]{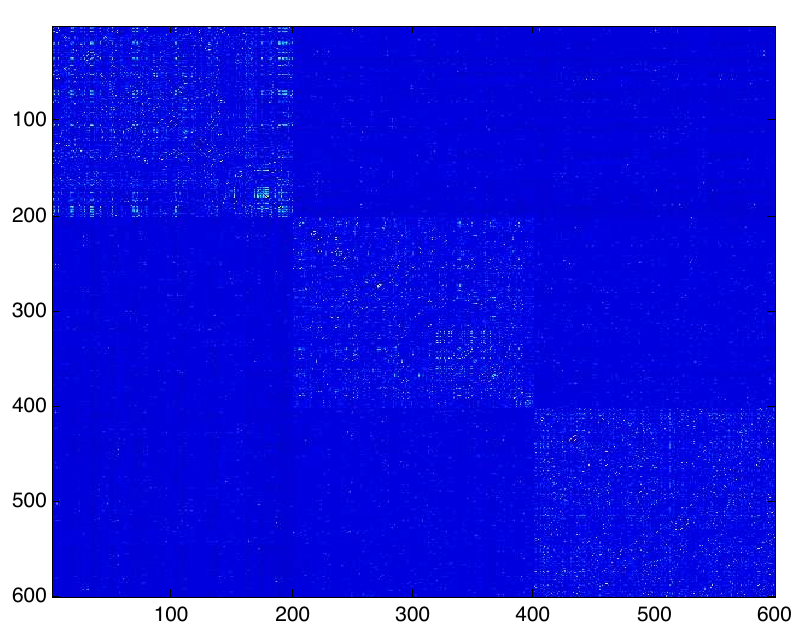} \\
\mbox{\small{(a)}} & \mbox{\small{(b)}} & \mbox{\small{(c)}} \\
\end{tabular}
\end{center}
\caption{Images of the different similarity matrices, (a) $\mathbf{K}_\text{BoPP}^{\mathrm{mds}}$, (b) $\mathbf{K}_\text{BoPS}^{\mathrm{mds}}$, and (c) $\mathbf{Q}$, computed on the news-3cl-1 dataset. Nodes have been sorted according to classes. We observe that classes are clearly visible in (a) and (b). For the standard modularity (c), the class discrimination is less clear.
}
\label{kernel_images}
\end{figure}

These kernels and similarity matrices are real symmetric when working with undirected graphs. All the above kernels and methods will be compared by following the experimental settings described hereafter. For illustration, a picture of some of the kernels is shown in Figure \ref{kernel_images}.

\subsection{Experimental settings}
\label{Subsec_experimental_settings}
In this experiment, we address the task of classification of unlabeled
nodes in partially labelled graphs. The method we use is directly inspired from \cite{Tang-2009}. It consists of two steps: (1) extracting the latent social dimensions, which may be done using any matrix decomposition technique or by using a graphical topic model. Here, we used, as in \cite{Tang-2009}, a simple spectral decomposition of the relevant matrices. More precisely, we extracted the top eigenvectors of the compared kernel matrices just described (see Subection \ref{introducedKernels}). This aims to perform a classical multidimensional scaling from distances when using the MDS transformation of Equation (\ref{Eq_centering}) and a kernel principal components analysis when using the Gaussian mapping of Equation (\ref{Eq_gaussianKernel01}). (2) training a classifier on the extracted latent space. In this space, each feature corresponds to one latent variable (i.e.\ one of the top eigenvectors). The number of social dimensions has been set to 5 for all the suggested measures and the classifier is a one-vs-rest linear SVM. Note that we also investigated different numbers of social dimensions $[10, 50, 500]$ but the performances did not change significantly -- these results are therefore not reported here. 

The classification accuracy is computed for a labeling rate of 20\%, i.e.\ proportion of
nodes for which the label is known\footnote{Other settings were also investigated, leading to similar conclusions; they are therefore omitted here.}. The labels of remaining nodes (80\%)
are removed and used as test data. For this considered labeling rate, an external stratified 5-fold cross-validation (each fold defining in turn the 20\% labeled data) was performed, on which performances are averaged.  For each fold of the external cross-validation, a 5-fold internal cross-validation is performed on the remaining labelled nodes in order to tune the hyper-parameters of the SVM and each kernel/distance ($\theta=\{0.01, 0.1, 1, 2, 5, 10\}$ for the bag-of-paths based approaches and $c=\{0.01, 0.1, 1, 10, 100\}$ for the SVM). Then, performances on each fold are assessed on the remaining, unlabeled, nodes (test data) with the hyper-parameter tuned during the internal cross-validation.

For each unlabeled node, the various classifiers predict the most suitable category according to the procedure described below.
We compute, for each method, the average classification accuracy obtained on the five folds of the cross-validation. A nonparametric Friedman-Nemenyi statistical test \cite{Demsar2006} is then performed across all datasets in order to compare the different methods.

\begin{landscape}

\begin{table}
\begin{centering}
\scalebox{0.80}{%
\begin{tabular}{l rrrrrrrrrrrrrr}
\midrule
  Method:  &  BoPM  &  BoPP-g  &  BoPP-mds  &  BoPS-g  &  BoPS-mds  &  LF-g  &  LF-mds  &  MD  &  Q  &  RCT  &  RL  &  RSP-g  &  RSP-mds  &  SoS \\
Dataset: & & & & & \\
\midrule

  webKB-texas  &  74.85  &  \textbf{77.40}  &  74.92  &  76.57  &  76.95  &  74.92  &  72.75  &  58.01  &  72.75  &  70.89  &  48.73  &  74.92  &  75.75  &  74.63 \\
  webKB-washington  &  66.19  &  71.49  &  70.68  &  68.61  &  70.05  &  \textbf{72.24}  &  70.10  &  66.53  &  59.50  &  67.40  &  40.78  &  70.68  &  70.33  &  65.61 \\
  webKB-wisconsin  &  72.84  &  \textbf{75.50}  &  73.49  &  73.13  &  74.14  &  73.78  &  71.91  &  70.48  &  72.70  &  70.76  &  45.04  &  73.35  &  72.49  &  73.71 \\
  webKB-cornell  &  \textbf{60.04}  &  55.57  &  58.31  &  56.29  &  58.46  &  58.38  &  56.43  &  51.73  &  51.23  &  46.82  &  41.91  &  58.31  &  56.87  &  58.67 \\
  imdb  &  74.44  &  50.75  &  50.68  &  50.77  &  50.68  &  50.71  &  50.71  &  52.67  &  66.64  &  56.93  &  68.44  &  50.75  &  50.68  &  \textbf{78.14} \\
  news-2cl-1  &  96.00  &  95.06  &  94.25  &  95.25  &  94.75  &  95.06  &  95.94  &  \textbf{97.56}  &  94.81  &  90.94  &  90.69  &  94.31  &  94.06  &  92.50 \\
  news-2cl-2  &  89.83  &  91.02  &  90.70  &  \textbf{91.71}  &  91.58  &  90.89  &  89.26  &  90.64  &  91.02  &  86.43  &  87.50  &  90.52  &  90.89  &  89.89 \\
  news-2cl-3  &  94.49  &  \textbf{95.99}  &  95.68  &  95.80  &  95.99  &  95.55  &  95.05  &  95.49  &  94.17  &  92.86  &  93.36  &  95.99  &  95.30  &  94.11 \\
  news-3cl-1  &  \textbf{94.42}  &  93.92  &  93.08  &  92.92  &  93.08  &  93.17  &  92.25  &  91.75  &  93.33  &  72.17  &  78.25  &  93.50  &  92.67  &  91.75 \\
  news-3cl-2  &  \textbf{93.31}  &  92.98  &  92.06  &  92.89  &  92.39  &  91.68  &  91.39  &  89.38  &  92.64  &  54.98  &  55.64  &  92.98  &  92.18  &  89.72 \\
  news-3cl-3  &  91.18  &  93.03  &  93.24  &  \textbf{93.99}  &  93.78  &  91.39  &  91.01  &  81.68  &  90.55  &  64.50  &  57.61  &  93.11  &  93.07  &  90.84 \\
  news-5cl-1  &  86.32  &  \textbf{87.98}  &  87.57  &  87.80  &  87.47  &  86.02  &  86.50  &  76.40  &  81.04  &  48.72  &  27.73  &  86.90  &  87.30  &  86.52 \\
  news-5cl-2  &  79.48  &  78.25  &  81.83  &  77.80  &  81.68  &  77.23  &  80.88  &  60.41  &  75.28  &  51.88  &  47.60  &  77.25  &  81.41  &  \textbf{82.51} \\
  news-5cl-3  &  73.60  &  81.02  &  81.09  &  80.29  &  80.77  &  79.91  &  78.91  &  61.01  &  76.00  &  41.68  &  27.83  &  80.97  &  80.22  &  \textbf{81.92} \\
  
\bottomrule
\end{tabular}
}
\caption{Classification accuracy (correct classification rate) for the bag-of-paths based distances and the competing methods obtained on each dataset, using 5 social dimensions. Only the results for graphs with 20\% labeling rate are reported. The best performing method of each data set is highlighted in boldface.}
\label{SSL_classifRate}
\end{centering}
\end{table}

\end{landscape}

\subsection{Results and discussions}

Table \ref{SSL_classifRate} reports average classification accuracies of the methods on all the datasets, for a proportion of 20\% of labeling rate. The method performing best is presented in boldface for each data set. Then, a simple Borda ranking of the methods is performed and shown in Table \ref{copeland}. Each method is given a score equal to its rank (methods are sorted in ascending order of accuracy, worst first and best last) for each dataset. The best method overall is the one showing the highest Borda score.

From these tables, it can be observed that the bag-of-paths (BoP) and the randomized-shortest-paths (RSP) based approaches obtain competitive results in comparison with the other methods. Indeed, both the BoPP and the BoPS consistently provide good results. The logarithmic forest distance also obtains good overall results. However, we can further observe that the best method is dataset-dependent; this shows that it is often useful to investigate different methods when facing a network-based semi-supervised classification problem. Moreover, the differences in performance among the best performing methods is often small. This can be understood by the fact that we selected the most promising candidate methods for the comparisons, but also by the fact that several investigated distances are derived from a similar framework.

\begin{figure}[th!]
\begin{center}
\includegraphics[width=1.0\textwidth]{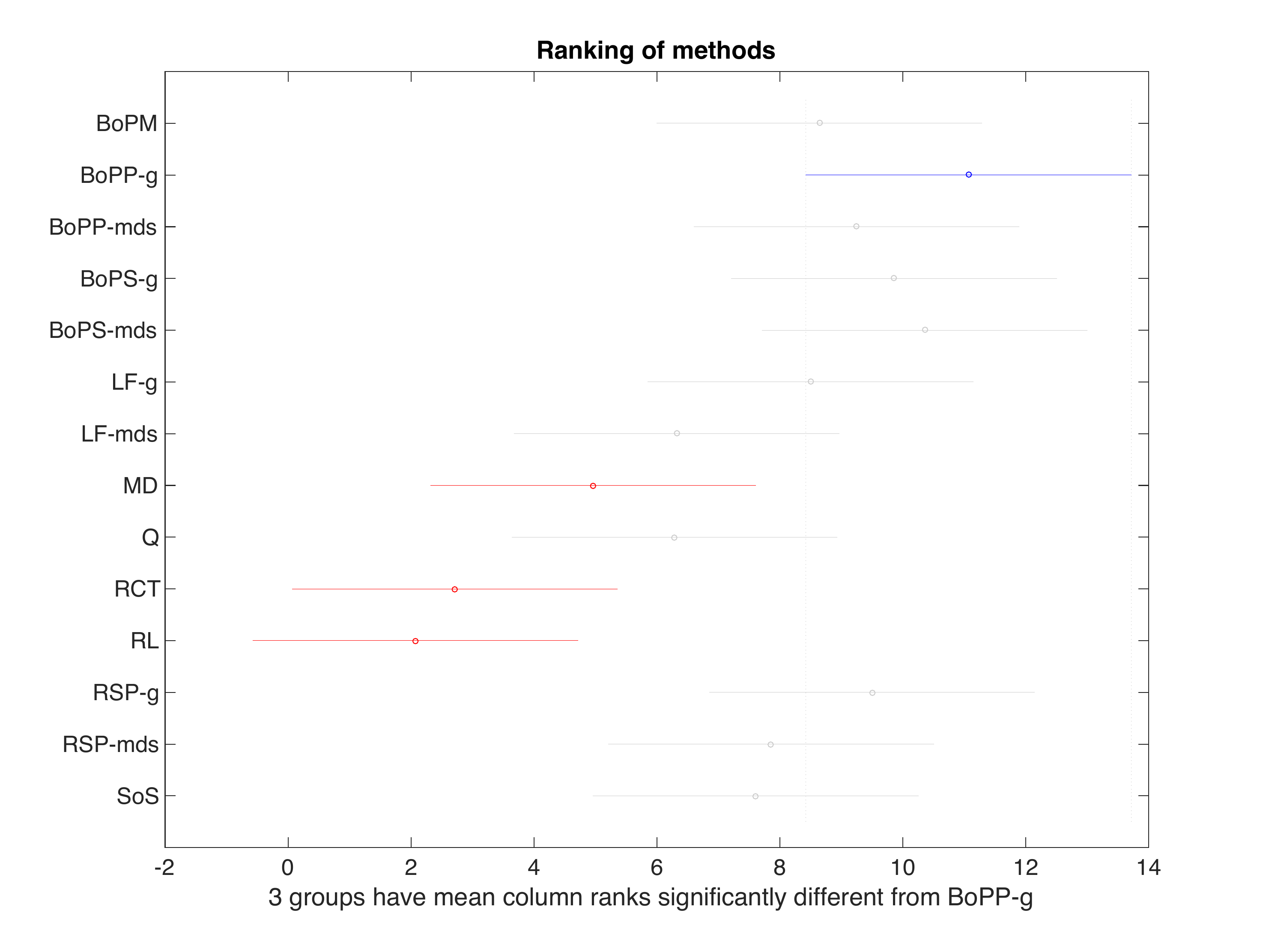}
\end{center}
\caption{Friedman-Nemenyi ranking over the 14 graphs (the larger, the better). Two methods are considered as significantly different when their confidence intervals do not overlap. The best method (BoPP-g) is highlighted.
}
\label{Fig_nemeny01}
\end{figure}

Moreover, in order to rate globally the performances of each method, we use a nonparametric Friedman-Nemenyi statistical test \cite{Demsar2006} allowing to compare them across all the datasets. The obtained ranking scores are presented in Figure \ref{Fig_nemeny01} and are similar to those provided by the Borda ranking. The figure confirms that the BoP and RSP distances provide good results, although not significantly different from the logarithmic forest and the two baseline methods (the modularity matrix Q and the sum-of-similarities SoS). This is partly because the Friedman-Nemenyi test is rather conservative, especially when comparing many different techniques.

Therefore, in order to further investigate the results, we also computed pairwise comparisons through a nonparametric one-sided Wilcoxon signed-rank test for matched data ($\alpha = 0.05$). This paired test shows that all the introduced bag-of-paths methods (BoPP-g, BoPP-mds, BoPS-g, BoPS-mds) are significantly better than our first baseline (Q), but not necessarily better than the second baseline (SoS). Indeed, only one method, BoPS-mds, provided significantly better results than SoS (but close to the critical value, $p$-value = 0.033). This confirms that the SoS can be considered as a good baseline which, in addition, is simple, efficient, and scales to large, sparse, networks \cite{Mantrach-2011}.

Although a little under the bag-of-paths based approaches, note that the randomized shortest path (RSP) and the logarithmic forest (LF) methods associated to the gaussian transformation are also competitive, consistently providing good results, and significantly better than our first baseline (Q). Note also that this simple modularity matrix based method Q, although below the best methods, especially in the 5-classes setting, provides reasonable results.

Curiously, the spectral method applied to the three kernels (the Markov diffusion kernel (MD), the regularized commute time kernel (RCT) and the regularized Laplacian kernel (RL)) provides bad performances (all three kernels perform significantly worse than the two baselines). This is especially odd, as these kernels obtained good results when used in a sum-of-similarities context \cite{FoussKernelNN-2011,Mantrach-2011} -- see the results obtained by the sum-of-similarities based on the RCT kernel (SoS) in Table \ref{SSL_classifRate} which is not statistically different from the best method. This could be related to the recent comparison in \cite{Ivashkin-2016} showing that taking the logarithm of some well-known kernels improves the performances in node clustering tasks.

Concerning the transformation from distances to inner products of Equations (\ref{Eq_centering}) and (\ref{Eq_gaussianKernel01}), the Gaussian kernel often provides slightly better results than multidimensional scaling, but not always so.

In summary, these experiments showed that the introduced BoP families of distances (BoPP, BoPS), but also the already known randomized shortest path (RSP) and the logarithmic forest (LF) distances, achieve good performances in comparison with our two baseline methods (Q and SoS) on the investigated datasets. However, we found that the introduced distances are not necessarily significantly better (although globally ranked better) than the second baseline, the sum-of-similarities method (SoP) based on the RCT kernel \cite{FoussKernelNN-2011,Mantrach-2011}. Because this SoS technique is fast and scales to large graphs \cite{Mantrach-2011}, it can be concluded that the introduced distances do not bring much added value here in our semi-supervised classification tasks. Still, this has to be confirmed in larger experiments. Indeed, in further work, we plan to conduct a systematic, comprehensive, comparison of families of distances and kernels on clustering, classification and dimensionality reduction tasks.

\begin{table}[t!]
\begin{center}
\scalebox{0.80}{%
\begin{tabular}{l c r }
\hline
Method & Rank & Score \\
\hline
BoPP-g & 1 & 162 \\
BoPS-mds & 2 & 155 \\
BoPP-mds & 3 & 143 \\
RSP-g  & 4 & 141 \\
BoPS-g & 5 & 140 \\
LogF-g & 6 & 127 \\
BoPM & 7 & 125 \\
RSP-mds & 8 & 118 \\
SoS & 9 & 109 \\
LogF-mds & 10 & 95 \\
Q & 11 & 89 \\
MD & 12 & 72 \\
RCT & 13 & 38 \\
RL & 14 & 29 \\
\hline
\end{tabular}}
\caption{Ranking of the different classification methods according to Borda's method (the higher score, the better).}
\label{copeland}
\end{center}
\end{table}

\section{Conclusion and further work}
\label{Sec_conclusion01}

This work introduced the bag-of-paths framework considering a bag containing the set of paths in the network. By defining a Gibbs-Boltzmann distribution on this set of paths penalizing long paths, we can easily compute various quantities such as distance measures between nodes.
It is also shown that one of the two introduced distance measures has some nice properties, like interpolating between the shortest path distance and the resistance distance (up to a constant factor). Experiments have shown that the BoP framework can provide competitive algorithms within a clear theoretical framework.

Indeed, as demonstrated in semi-supervised classification experiments, the kernels associated to the distance measures derived from the bag-of-paths probabilities achieve good results. Consistency of performance across the different datasets shows that the bag-of-paths framework seems to induce some promising distance and similarity measures on graphs, based on its structure.

The framework is rich and other quantities of interest can be defined within this framework, which are pursued in parallel. For instance, a betweenness measure can be defined as $\text{P}(int=j|s=i, e=k)$, the probability that a path starting in $i$ and ending in $k$ visits $j$ as an intermediate node  \cite{Lebichot-2014}. Another idea is to reformulate the modularity matrix in terms of paths instead of direct links \cite{Devooght-2014}. Still another application would be the computation of a robustness measure capturing the criticality of the nodes (under development). The idea then would be to compute the change in accessibility between nodes when deleting one node within the BoP framework. Nodes having a wide impact on reachability are then considered as highly critical. 

Another idea would be to investigate graph cut from the bag-of-hitting-paths probabilities in Equation (\ref{Eq_bag_of_hitting_paths_probabilities02}) instead of the adjacency matrix. We also plan to evaluate experimentally the potential distance (see Equation (\ref{Eq_potential_recurrence_formula01})) as a distance between sequences of characters by adapting it to a directed acyclic graph, as in \cite{Garcia-Diez-2011}.

Finally, we plan to make a systematic experimental comparison of families of distances and kernels on clustering, semi-supervised classification and dimensionality reduction tasks, while trying to analyze the theoretical properties of the proposed distances families by following \cite{Hashimoto-2015}. In particular, we will investigate the new kernels introduced recently in \cite{Ivashkin-2016} where it is shown on node clustering tasks that taking the logarithm of well-known kernels improves significantly the performances.

\section*{Acknowledgments}

This work was partially supported by the Immediate and the Brufence projects funded by InnovIris (Brussels Region), as well as former projects funded by the Walloon region. We thank these institutions for giving us the opportunity to conduct both fundamental and applied research.
We also thank Bertrand Lebichot and Yutaro Shigeto for helping us during the last experiments.
Finally, we thank the anonymous reviewers for their comments.

\appendix
\section*{Appendix}

\section{Sum of reference probabilities over hitting paths}
\label{app_hitting_paths_likelihood}

In this appendix, it is shown that the sum over all hitting paths of the reference probabilities is equal to one. We thus have to show
\begin{align}
\sum_{\wp \in\mathcal{P}^{\mathrm{h}}} \tilde{\text{P}}^{\mathrm{ref}}(\wp)
&= \sum_{t=0}^{\infty} \sum_{\wp \in\mathcal{P}^{\mathrm{h}}(t)} \tilde{\text{P}}^{\mathrm{ref}}(\wp) 
= \sum_{t=0}^{\infty} \sum_{i,j=1}^{n} \sum_{\wp_{ij} \in\mathcal{P}^{\mathrm{h}}_{ij}(t)} \tilde{\text{P}}^{\mathrm{ref}}(\wp_{ij}) \nonumber \\
&= \frac{1}{n^2} \sum_{t=0}^{\infty} \sum_{i,j=1}^{n} \sum_{\wp_{ij} \in\mathcal{P}^{\mathrm{h}}_{ij}(t)} \tilde{\pi}^{\mathrm{ref}}(\wp_{ij}) 
\overset{?}{=} 1
\label{Eq_hitting_paths_likelihoods_sum01}
\end{align}
where $\mathcal{P}^{\mathrm{h}}(t)$ is the set of all hitting paths of length exactly equal to $t$ and $\mathcal{P}_{ij}^{\mathrm{h}}(t)$ the set of such hitting paths connecting $i$ to $j$.
As stated before, because we assume that the a priori probability of choosing the starting node and ending node is uniform, $\tilde{\text{P}}^{\mathrm{ref}}(\wp_{ij}) = \frac{1}{n^{2}} \pi^{\mathrm{ref}}(\wp_{ij})$ with $\pi^{\mathrm{ref}}(\wp_{ij})$ being the likelihood of the path $\wp_{ij}$, i.e., the product of transition probabilities $p^{\mathrm{ref}}_{ll'}$ along the path of length $t$, $\pi^{\mathrm{ref}}(\wp_{ij}) = \prod_{\tau=1}^{t} p^{\mathrm{ref}}_{k_{\tau-1} k_{\tau}}$ with $k_{0}=i$, $k_{t}=j$ and no intermediate node being equal to node $j$.

As we are concerned with hitting paths stopping in node $j$, let us consider the absorbing, killing, Markov chain on $G$ with transition probabilities $p^{\mathrm{ref}}_{ll'}$ for $l \ne j$ and $p^{\mathrm{ref}}_{jl'} = 0$ for all $l'$. In other words, node $j$ is made killing and absorbing.

We now introduce a new quantity, $q_{k}^{(ij)}(t)$, on this absorbing Markov chain, defined as the probability of finding the process in state $k$ at time $t$ when considering walks from starting node $i$ to absorbing node $j$. This probability can easily be computed thanks to the following recurrence relation
\begin{equation}
\begin{cases}
q_{k}^{(ij)}(0) = \delta_{ik} &\text{for } t=0 \\
q_{k}^{(ij)}(t) = { \displaystyle \sum_{\substack{l=1 \\ l \ne j}}^{n} } q_{l}^{(ij)}(t-1) \, p^{\mathrm{ref}}_{lk} &\text{for } t \ge 1
\end{cases}
\label{Eq_recurrence_absorbtion_probability01}
\end{equation}
which says that the probability of being in node $k$ at time $t$ is the sum of the probabilities of being in any node $l$ (except node $j$ which is absorbing) at time $t-1$ times the probability of jumping from $l$ to $k$. When $k=j$, the quantity computes the probability of being absorbed in node $j$ at time $t$, given that we started from $i$ at time 0.

Let us now compute the last quantity appearing in Equation (\ref{Eq_hitting_paths_likelihoods_sum01}), the sum of hitting paths likelihoods from $i$ to $j$, assuming $i \ne j$,
\begin{equation}
\sum_{\wp_{ij} \in\mathcal{P}^{\mathrm{h}}_{ij}(t)} \tilde{\pi}^{\mathrm{ref}}(\wp_{ij})
= \sum_{\substack{k_{1}=1 \\ k_{1} \ne j}}^{n} \sum_{\substack{k_{2}=1 \\ k_{2} \ne j}}^{n} \cdots \sum_{\substack{k_{t-1}=1 \\ k_{t-1} \ne j}}^{n}  p^{\mathrm{ref}}_{i k_{1}} p^{\mathrm{ref}}_{k_{1} k_{2}} p^{\mathrm{ref}}_{k_{2} k_{3}} \cdots p^{\mathrm{ref}}_{k_{t-1} j} \quad \text{for } t > 0
\end{equation}
and it is equal to $0$ when $t=0$ because there is no path of length zero connecting two different nodes.

But the second-hand quantity in this last equation is nothing else than the sequential application of recurrence (\ref{Eq_recurrence_absorbtion_probability01}) for $t, t-1, \dots, 0$, therefore computing $q_{j}^{(ij)}(t)$, that is, the probability of being absorbed in node $j$ in exactly $t$ steps.
Therefore, $\sum_{\wp_{ij} \in\mathcal{P}^{\mathrm{h}}_{ij}(t)} \tilde{\pi}^{\mathrm{ref}}(\wp_{ij}) = q_{j}^{(ij)}(t)$ when $i \neq j$.

Moreover, as we know that the process necessarily ends in absorbing node $j$ at some point (see, e.g., \cite{Grinstead-1997}), $\sum_{t=0}^{\infty} q_{j}^{(ij)}(t) = 1$ holds when $i \ne j$.

If $i=j$, the probability of finding the process in node $j$ is 1 at $t=0$ (a zero-length path) and then collapses to 0 when $t>0$, which also provides $\sum_{t=0}^{\infty} q_{j}^{(jj)}(t) = 1$.

Equation (\ref{Eq_hitting_paths_likelihoods_sum01}) then becomes
\begin{equation}
\sum_{\wp \in\mathcal{P}^{\mathrm{h}}} \tilde{\text{P}}^{\mathrm{ref}}(\wp)
= \frac{1}{n^2} \sum_{i,j=1}^{n} \sum_{t=0}^{\infty} q_{j}^{(ij)}(t)
= \frac{1}{n^2} \sum_{i,j=1}^{n} 1
= 1
\end{equation}
which is the desired result.
In addition, this also shows that
\begin{equation}
\sum_{\wp \in\mathcal{P}^{\mathrm{h}}_{ij}} \tilde{\pi}^{\mathrm{ref}}(\wp)
= 1,
\end{equation}
that is, the sum over the path likelihoods is equal to 1 for hitting paths.

\section{Computation of the entries of $\mathbf{Z^{(-j)}}$ in terms of the fundamental matrix}\label{app_zEntries}

All the entries of $\mathbf{Z}^{(-j)}$ can be computed efficiently in terms of the fundamental matrix $\mathbf{Z} = (\mathbf{I}-\mathbf{W})^{-1}$.

This result can be understood as follows. Each non-hitting path $\wp_{ij} \in \mathcal{P}_{ij}$ can be split uniquely into two sub-paths, before hitting node $j$ for the first time, $\wp_{ij}^{\mathrm{h}} \in \mathcal{P}_{ij}^{\mathrm{h}}$, and after hitting node $j$, $\wp_{jj} \in \mathcal{P}_{jj}$.
These two sub-paths can be chosen independently because their concatenation
is a valid path, with $\wp_{ij}^{\mathrm{h}} \circ \wp_{jj} \in \mathcal{P}_{ij}$ being the concatenation of the two paths.
Now, as $\tilde{c}(\wp_{ij}) = \tilde{c}(\wp_{ij}^{\mathrm{h}}) + \tilde{c}(\wp_{jj})$ and $\tilde{\pi}^{\mathrm{ref}}(\wp_{ij}) = \tilde{\pi}^{\mathrm{ref}}(\wp_{ij}^{\mathrm{h}}) \tilde{\pi}^{\mathrm{ref}}(\wp_{jj})$ 
for any $\wp_{ij} = \wp_{ij}^{\mathrm{h}} \circ \wp_{jj}$, we obtain 
\begin{align}
z_{ij}
&= \sum_{\wp_{ij}\in\mathcal{P}_{ij}} \tilde{\pi}^{\mathrm{ref}}(\wp_{ij}) \exp[-\theta \tilde{c}(\wp_{ij})] \nonumber \\
&=  \sum_{\substack{ {\wp_{ij}^{\mathrm{h}} \in \mathcal{P}_{ij}^{\mathrm{h}} } \\ {\wp_{jj} \in \mathcal{P}_{jj} } }} \tilde{\pi}^{\mathrm{ref}}(\wp_{ij}^{\mathrm{h}}) \tilde{\pi}^{\mathrm{ref}}(\wp_{jj}) \exp[-\theta \tilde{c}(\wp_{ij}^{\mathrm{h}})]\exp[-\theta \tilde{c}(\wp_{jj})] \nonumber \\
&=  \left(\sum_{\wp_{ij}^{\mathrm{h}} \in \mathcal{P}_{ij}^{\mathrm{h}}} \tilde{\pi}^{\mathrm{ref}}(\wp_{ij}^{\mathrm{h}}) \exp[-\theta \tilde{c}(\wp_{ij}^{\mathrm{h}})] \right) \left(\sum_{\wp_{jj}\in\mathcal{P}_{jj}} \tilde{\pi}^{\mathrm{ref}}(\wp_{jj}) \exp[-\theta \tilde{c}(\wp_{jj})] \right) \nonumber \\
&= z_{ij}^{(-j)} z_{jj}
\end{align}
and therefore $z_{ij}^{(-j)} = z_{ij}/z_{jj}$. 
Using this result, Equation (\ref{Eq_numerator_bag_paths_hitting01}) can be developed as
\begin{equation}
{\displaystyle \sum_{\wp\in\mathcal{P}^{\mathrm{h}}_{ij}}} \tilde{\pi}^{\mathrm{ref}}(\wp)\exp\left[-\theta \tilde{c}(\wp) \right] 
 = z_{ij}^{(-j)} = \frac{z_{ij}}{z_{jj}}
 \label{Eq_numerator_bag_paths_hitting05}
\end{equation}


\section{Triangle inequality proof for the surprisal distance} \label{app_surprisal_dist_proof}

In order for $\surdist_{ij}$ to be a distance measure, it has to be shown that it obeys the triangle inequality, $\surdist_{ik} \le \surdist_{ij} + \surdist_{jk}$ for all $i,j,k$. Note that $\surdist_{ij} = \infty$ when node $i$ and node $j$ are not connected (they belong to different connected components) -- this is why we require $G$ to be strongly connected. In addition, note that the triangle inequality is trivially satisfied if either $i=j$, $j=k$ or $i=k$. Thus, we only need to prove the case $i \ne j \ne k \ne i$.

In order to prove the result, consider the set of paths $\mathcal{P}_{ik}$ from node $i$ to node $k$. We now compute the probability that such paths pass through an \emph{intermediate} node $int = j$ where $i \ne j \ne k \ne i$,
\begin{equation}
\text{P}(s=i,int=j,e=k)
= \frac{{\displaystyle \sum_{\wp\in\mathcal{P}_{ik}}} \delta(j \in \wp) \, \tilde{\pi}^{\mathrm{ref}}(\wp)\exp\left[-\theta \tilde{c}(\wp)\right]}{{\displaystyle \sum_{\wp' \in \mathcal{P}}} \tilde{\pi}^{\mathrm{ref}}(\wp')\exp\left[-\theta \tilde{c}(\wp')\right]}
\label{Eq_bag_of_paths_probability_intermediate01}
\end{equation}
where $\delta(j \in \wp)$ is a Kronecker delta equal to 1 if the path $\wp$ contains (at least once) node $j$, and 0 otherwise.
It is clear from Equations (\ref{Eq_bag_of_paths_probability_hitting01}) and (\ref{Eq_bag_of_paths_probability_intermediate01}) that
\begin{equation}
\text{P}(s=i,e=k) \ge \text{P}(s=i,int=j,e=k) \quad \text{for } i \ne j \ne k \ne i
\label{Eq_bag_of_paths_probability_inequality01}
\end{equation}

Let us transform Equation (\ref{Eq_bag_of_paths_probability_intermediate01}), using the fact that each path $\wp_{ik}$ between $i$ and $k$ passing through $j$ can be decomposed uniquely into a \emph{hitting} sub-path $\wp_{ij}$ from $i$ to $j$ and a non-hitting sub-path $\wp_{jk}$ from $j$ to $k$. The sub-path $\wp_{ij}$ is found by following path $\wp_{ik}$ until reaching $j$ for the first time. Therefore, for $i \ne j \ne k \ne i$,
\begin{align}
&\text{P}(s=i,int=j,e=k)
= \frac{{\displaystyle \sum_{\wp\in\mathcal{P}_{ik}}} \delta(j \in \wp) \, \tilde{\pi}^{\mathrm{ref}}(\wp)\exp\left[-\theta \tilde{c}(\wp)\right]}{\mathcal{Z}} \nonumber \\
&= \frac{{\displaystyle \sum_{\wp_{ij} \in \mathcal{P}^{\mathrm{h}}_{ij}} \sum_{\wp_{jk} \in \mathcal{P}_{jk}} } \tilde{\pi}^{\mathrm{ref}}(\wp_{ij}) \tilde{\pi}^{\mathrm{ref}}(\wp_{jk})\exp\left[-\theta (\tilde{c}(\wp_{ij}) + \tilde{c}(\wp_{jk})) \right]}{\mathcal{Z}} \nonumber \\
&= \frac{ \left[ {\displaystyle \sum_{\wp_{ij} \in \mathcal{P}^{\mathrm{h}}_{ij}} } \tilde{\pi}^{\mathrm{ref}}(\wp_{ij}) \exp\left[-\theta \tilde{c}(\wp_{ij})  \right] \right] \left[ {\displaystyle \sum_{\wp_{jk} \in \mathcal{P}_{jk}} }  \tilde{\pi}^{\mathrm{ref}}(\wp_{jk}) \exp\left[-\theta \tilde{c}(\wp_{jk}) \right] \right] }{\mathcal{Z}} \nonumber \\
&= \mathcal{Z}_{\mathrm{h}} \frac{ \left[ {\displaystyle \sum_{\wp_{ij} \in \mathcal{P}^{\mathrm{h}}_{ij}} } \tilde{\pi}^{\mathrm{ref}}(\wp_{ij}) \exp\left[-\theta \tilde{c}(\wp_{ij})  \right] \right] } {\mathcal{Z}_{\mathrm{h}}} \frac{\left[ {\displaystyle \sum_{\wp_{jk} \in \mathcal{P}_{jk}} }  \tilde{\pi}^{\mathrm{ref}}(\wp_{jk}) \exp\left[-\theta \tilde{c}(\wp_{jk}) \right] \right] }{\mathcal{Z}} \nonumber \\
&= \mathcal{Z}_{\mathrm{h}} \, \text{P}_{\mathrm{h}}(s=i,e=j) \, \text{P}(s=j,e=k), \text{ for } i \ne j \ne k \ne i
\label{Eq_bag_of_paths_probability_intermediate02}
\end{align}

Combining Inequality (\ref{Eq_bag_of_paths_probability_inequality01}) and Equation (\ref{Eq_bag_of_paths_probability_intermediate02}) yields
\begin{equation}
\text{P}(s=i,e=k) \ge \mathcal{Z}_{\mathrm{h}} \, \text{P}_{\mathrm{h}}(s=i,e=j) \, \text{P}(s=j,e=k), \text{ for } i \ne j \ne k \ne i
\label{Eq_bag_of_paths_probability_inequality03}
\end{equation}

Replacing the non-hitting bag-of-paths probabilities by their expressions (see Equation (\ref{Eq_bag_of_paths_probabilities01})) in function of the elements of the fundamental matrix, $\text{P}(s=i,e=k) = z_{ik}/\mathcal{Z}$ and $\text{P}(s=j,e=k) = z_{jk}/\mathcal{Z}$, in the previous Inequality (\ref{Eq_bag_of_paths_probability_inequality03}) provides $z_{ik}/\mathcal{Z} \ge \mathcal{Z}_{\mathrm{h}} \, \text{P}_{\mathrm{h}}(s=i,e=j) \, z_{jk}/\mathcal{Z}$. Further dividing each member by $( \mathcal{Z}_{\mathrm{h}} z_{kk} )$ gives $z_{ik}/(\mathcal{Z}_{\mathrm{h}} z_{kk}) \ge \mathcal{Z}_{\mathrm{h}} \, \text{P}_{\mathrm{h}}(s=i,e=j) \, z_{jk}/(\mathcal{Z}_{\mathrm{h}} z_{kk})$. Finally, using $\text{P}_{\mathrm{h}}(s=i,e=k) = z_{ik}/(\mathcal{Z}_{\mathrm{h}} z_{kk})$ (see Equation (\ref{Eq_bag_of_paths_probability_hitting02})), we obtain
\begin{equation}
\text{P}_{\mathrm{h}}(s=i,e=k) \ge \mathcal{Z}_{\mathrm{h}} \, \text{P}_{\mathrm{h}}(s=i,e=j) \, \text{P}_{\mathrm{h}}(s=j,e=k)
\label{Eq_bag_of_paths_probability_inequality04}
\end{equation}
for $i \ne j \ne k \ne i$.
Now, from Equation (\ref{Eq_bag_of_paths_hitting_partition01}) and the fact that the $z_{ij}$ are nonnegative, it is clear that $\mathcal{Z}_{\mathrm{h}} \ge 1$; thus
\begin{equation}
\text{P}_{\mathrm{h}}(s=i,e=k) \ge \text{P}_{\mathrm{h}}(s=i,e=j) \, \text{P}_{\mathrm{h}}(s=j,e=k), \text{ for } i \ne j \ne k \ne i
\label{Eq_bag_of_paths_probability_inequality05}
\end{equation}

Finally, by taking $- \log$ of Inequality (\ref{Eq_bag_of_paths_probability_inequality05}), we obtain
\begin{equation}
-\log\text{P}_{\mathrm{h}}(s=i,e=k) \le -\log\text{P}_{\mathrm{h}}(s=i,e=j) -\log\text{P}_{\mathrm{h}}(s=j,e=k), 
\label{Eq_bag_of_paths_probability_inequality06}
\end{equation}
for $i \ne j \ne k \ne i$.
Thus, the surprisal measure, $-\log\text{P}_{\mathrm{h}}(s=i,e=j)$, obeys the triangle inequality. Therefore the distance $\dist_{ij}^{\mathrm{h}} = -(\log\text{P}_{\mathrm{h}}(s=i,e=j) + \log\text{P}_{\mathrm{h}}(s=j,e=i))/2$ also enjoys this property.

\section{Proof of the geodetic property of the potential distance}\label{app_geodetic}
From the definition of the bag-of-paths probability (Equation (\ref{Eq_bag_of_paths_probability01})), as well as Equation (\ref{Eq_bag_of_paths_probability_intermediate01}) defining $\text{P}(s=i,int=j,e=k)$, we have for $i \ne j \ne k \ne i$
  \begin{align}
  & \text{P}(s=i,e=k)
= \frac{{\displaystyle \sum_{\wp\in\mathcal{P}_{ik}}} \tilde{\pi}^{\mathrm{ref}}(\wp)\exp\left[-\theta \tilde{c}(\wp)\right]}{\mathcal{Z}} \nonumber \\
&= \frac{{\displaystyle \sum_{\wp\in\mathcal{P}_{ik}}} \delta(j \in \wp) \, \tilde{\pi}^{\mathrm{ref}}(\wp)\exp\left[-\theta \tilde{c}(\wp)\right]}{\mathcal{Z}}
+ \frac{{\displaystyle \sum_{\wp\in\mathcal{P}_{ik}}} (1 - \delta(j \in \wp)) \, \tilde{\pi}^{\mathrm{ref}}(\wp)\exp\left[-\theta \tilde{c}(\wp)\right]}{\mathcal{Z}} \nonumber \\
&= \text{P}(s=i,int=j,e=k)
+ \frac{{\displaystyle \sum_{\wp\in\mathcal{P}_{ik}}} \delta(j \notin \wp) \, \tilde{\pi}^{\mathrm{ref}}(\wp)\exp\left[-\theta \tilde{c}(\wp)\right]}{\mathcal{Z}}
  \end{align}

Now, substituting $\text{P}(s=i,int=j,e=k)$ by $\mathcal{Z}_{\mathrm{h}} \, \text{P}_{\mathrm{h}}(s=i,e=j) \text{P}(s=j,e=k)$ (see Equation (\ref{Eq_bag_of_paths_probability_intermediate02})) in the previous equation yields
\begin{align}
\text{P}(s=i,e=k) = &\mathcal{Z}_{\mathrm{h}} \, \text{P}_{\mathrm{h}}(s=i,e=j) \text{P}(s=j,e=k) \nonumber \\
&+ \frac{{\displaystyle \sum_{\wp\in\mathcal{P}_{ik}}} \delta(j \notin \wp) \, \tilde{\pi}^{\mathrm{ref}}(\wp)\exp\left[-\theta \tilde{c}(\wp)\right]}{\mathcal{Z}}
\label{Eq_graph_geodetic_property01}
\end{align}

Further recalling that $\text{P}(s=i,e=k) = z_{ik}/\mathcal{Z}$ (Equation (\ref{Eq_bag_of_paths_probabilities01})) and $\text{P}_{\mathrm{h}}(s=i,e=j) = z_{ij}^{\mathrm{h}} / \mathcal{Z}_{\mathrm{h}}$ (Equation (\ref{Eq_bag_of_paths_probability_hitting02})), we transform Equation (\ref{Eq_graph_geodetic_property01}) into
\begin{equation}
z_{ik} = z_{ij}^{\mathrm{h}} z_{jk} + {\displaystyle \sum_{\wp\in\mathcal{P}_{ik}}} \delta(j \notin \wp) \, \tilde{\pi}^{\mathrm{ref}}(\wp)\exp\left[-\theta \tilde{c}(\wp)\right]
\end{equation}

Dividing both sides of the previous equation by $z_{kk}$ and recalling that $z_{ik}^{\mathrm{h}} = z_{ik}/z_{kk}$ (Equation (\ref{Eq_backward_variable_hitting_paths01})) provides
\begin{equation}
z_{ik}^{\mathrm{h}} = z_{ij}^{\mathrm{h}} z_{jk}^{\mathrm{h}} + \frac{1}{z_{kk}} {\displaystyle \sum_{\wp\in\mathcal{P}_{ik}}} \delta(j \notin \wp) \, \tilde{\pi}^{\mathrm{ref}}(\wp)\exp\left[-\theta \tilde{c}(\wp)\right]
\end{equation}
and we recover $z_{ik}^{\mathrm{h}} \ge z_{ij}^{\mathrm{h}} z_{jk}^{\mathrm{h}}$ (Equation (\ref{Eq_potential_function_inequality01})). The equality $z_{ik}^{\mathrm{h}} = z_{ij}^{\mathrm{h}} z_{jk}^{\mathrm{h}}$ ($i \ne j \ne k \ne i$) holds if and only if $\sum_{\wp\in\mathcal{P}_{ik}} \delta(j \notin \wp) \, \tilde{\pi}^{\mathrm{ref}}(\wp)\exp\left[-\theta \tilde{c}(\wp)\right] = 0$, which only occurs when all paths connecting $i$ and $k$ visit node $j$.
Thus, it is clear that $\dist^{\phi}_{ik} = \dist^{\phi}_{ij} + \dist^{\phi}_{jk}, i \ne j \ne k \ne i$ if and only if all paths $\wp \in \mathcal{P}^{\mathrm{h}}_{ik}$ connecting the source node $i$ and the destination node $k$ pass through node $j$. This property is called the graph-geodetic property in \cite{Chebotarev-2011}.

\section{Asymptotic result: for an undirected graph, the $\dist^{\phi}$ distance converges to the shortest path distance when $\theta \rightarrow \infty$}\label{app_shortestPathDist}

There are two ways to prove this property, each of them having its own benefits. The first proof is based on the bag-of-paths framework and is shorter. The second proof is inspired by \cite{Tahbaz-2006} and is longer, but establishes some interesting links with the Bellman-Ford formula for computing the shortest path distance in a network (see, e.g., \cite{Bertsekas-2000,Christofides_1975,Cormen-2009,Rardin-1998,Sedgewick-2011}).

\subsection{First proof}

Assuming $i \ne j$ and $\theta > 0$, let us recall (Equation (\ref{Eq_bag_of_paths_potential_distance01})), that is, $\dist^{\phi}_{ij} = ( \phi(i,j) + \phi(j,i) )/2$ with $\phi(i,j) = - \frac{1}{\theta} \log z_{ij}^{\mathrm{h}}$, and where $z_{ij}^{\mathrm{h}}$ is given by (Equation (\ref{Eq_backward_variable_hitting_paths01}), recalled here for convenience):
\begin{equation}
z_{ij}^{\mathrm{h}} = {\displaystyle \sum_{\wp\in\mathcal{P}^{\mathrm{h}}_{ij}}} \tilde{\pi}^{\mathrm{ref}}(\wp)\exp\left[-\theta \tilde{c}(\wp)\right]
\label{Eq_potential_function_hitting02}
\end{equation}
which is always positive for a strongly connected graph.

We now have to compute the asymptotic form of $z_{ij}^{\mathrm{h}}$ for $\theta \rightarrow \infty$ or, equivalently, $T \rightarrow 0$. Let the lowest-cost (shortest) paths from $i$ to $j$ be denoted as $\{ \wp_k^{*} \}$ and let $c^{*} = \tilde{c}(\wp_k^{*})$ be the cost of such a lowest-cost path. $c^{*}$ is therefore the minimum cost among all possible paths from $i$ to $j$. Say there are $m \ge 1$ such lowest-cost paths. Now, as $\sum_{\wp\in\mathcal{P}^{\mathrm{h}}_{ij}} \tilde{\pi}^{\mathrm{ref}}(\wp) = 1$, it is clear that $z_{ij}^{\mathrm{h}}$ is bounded by
\begin{equation}
z_{ij}^{\mathrm{h}} \le \sum_{\wp\in\mathcal{P}^{\mathrm{h}}_{ij}} \tilde{\pi}^{\mathrm{ref}}(\wp)\exp\left[-\theta c^{*}\right] = \exp\left[-\theta c^{*}\right] \sum_{\wp\in\mathcal{P}^{\mathrm{h}}_{ij}} \tilde{\pi}^{\mathrm{ref}}(\wp) = \exp\left[-\theta c^{*}\right]
\end{equation}
and is therefore finite. We also observe that it converges exponentially to $0$ when $\theta \rightarrow \infty$. Moreover, this last inequality implies
\begin{equation}
{\displaystyle \sum_{\wp\in\mathcal{P}^{\mathrm{h}}_{ij}}} \tilde{\pi}^{\mathrm{ref}}(\wp)\exp\left[-\theta (\tilde{c}(\wp) - c^{*})\right] \le 1
\label{Eq_bouded_value01}
\end{equation}
which shows that the quantity on the left-hand side is bounded.

We can now rewrite
\begin{align}
z_{ij}^{\mathrm{h}}
&= {\displaystyle \sum_{\wp\in\mathcal{P}^{\mathrm{h}}_{ij}}} \tilde{\pi}^{\mathrm{ref}}(\wp)\exp\left[-\theta \tilde{c}(\wp)\right]
= \exp\left[-\theta c^{*}\right] {\displaystyle \sum_{\wp\in\mathcal{P}^{\mathrm{h}}_{ij}}} \tilde{\pi}^{\mathrm{ref}}(\wp) \exp\left[-\theta (\tilde{c}(\wp) - c^{*})\right] \nonumber \\
&= \exp\left[-\theta c^{*}\right] \left( \sum_{i=1}^{m} \tilde{\pi}^{\mathrm{ref}}(\wp_i^{*}) +  {\displaystyle \sum_{\substack{ {\wp \in \mathcal{P}_{ij}^{\mathrm{h}} } \\ {\tilde{c}(\wp) > c^{*} } }}} \tilde{\pi}^{\mathrm{ref}}(\wp) \exp\left[-\theta (\tilde{c}(\wp) - c^{*}) \right] \right)
\label{Eq_potential_function_hitting04}
\end{align}

Let us now compute the potential $\phi(i,j) = - \frac{1}{\theta} \log z_{ij}^{\mathrm{h}}$ when $\theta \rightarrow \infty$. Using Equation (\ref{Eq_potential_function_hitting04}), we get
\begin{align}
\phi(i,j) &= - \frac{1}{\theta} \log z_{ij}^{\mathrm{h}}  \nonumber \\
&= - \frac{1}{\theta} \log \left[ \exp\left[-\theta c^{*}\right] \left( \sum_{i=1}^{m} \tilde{\pi}^{\mathrm{ref}}(\wp_i^{*}) +  {\displaystyle \sum_{\substack{ {\wp \in \mathcal{P}_{ij}^{\mathrm{h}} } \\ {\tilde{c}(\wp) > c^{*} } }}} \tilde{\pi}^{\mathrm{ref}}(\wp) \exp\left[-\theta (\tilde{c}(\wp) - c^{*}) \right] \right) \right] \nonumber \\
& = c^{*} - \frac{1}{\theta} \log \left( \sum_{i=1}^{m} \tilde{\pi}^{\mathrm{ref}}(\wp_i^{*}) + {\displaystyle \sum_{\substack{ {\wp \in \mathcal{P}_{ij}^{\mathrm{h}} } \\ {\tilde{c}(\wp) > c^{*} } }}} \tilde{\pi}^{\mathrm{ref}}(\wp) \exp\left[-\theta (\tilde{c}(\wp) - c^{*}) \right] \right) \nonumber \\
& \xrightarrow{\theta \rightarrow \infty} c^{*}
\label{Eq_potential_function_hitting05}
\end{align}
Here, the last limit applies because, following Equation (\ref{Eq_bouded_value01}), the expression inside the logarithm is finite and strictly positive (the first term is a positive constant and the second is positive and bounded (see Equation (\ref{Eq_bouded_value01}))).

Moreover, observing that, in the case of an undirected graph, the lowest cost from $j$ to $i$ is equal to the lowest cost from $i$ to $j$ (i.e., $c^{*}$), the distance $\dist^{\phi}_{ij} = \frac{\phi(i,j) + \phi(j,i)}{2} \xrightarrow{\theta \rightarrow \infty} c^{*}$. Therefore, the bag-of-hitting-paths potential distance provides the shortest path distance when $\theta \rightarrow \infty$.

\subsection{Second proof}

The second proof starts from Equation (\ref{Eq_recurrence_relations_hitting_backward01}), where we replace $w_{ij} = p_{ij}^{\mathrm{ref}} \exp[-\theta c_{ij}]$ in this expression with node $k$ absorbing,
\begin{equation}
z_{ik}^{\mathrm{h}} =
  \begin{cases}
   {\displaystyle \sum_{j=1}^{n} p_{ij}^{\mathrm{ref}} \exp[-\theta c_{ij}] \, z_{jk}^{\mathrm{h}} } & \text{for } i \ne k \\
   1       & \text{for } i = k \text{ (boundary condition)}
  \end{cases}
\end{equation}

Let us now compute the value of the potential $\phi(i,k)$ (Equation (\ref{Eq_bag_of_paths_potential_distance01})) for $i \ne k$ (when $i=k$, $\phi(k,k) = -\frac{1}{\theta} \log \left(z_{kk}/z_{kk} \right) = 0$),
\begin{align}
\phi(i,k) &= -\frac{1}{\theta} \log z_{ik}^{\mathrm{h}} \nonumber 
= -\frac{1}{\theta} \log \left[ \sum_{j=1}^{n} p_{ij}^{\mathrm{ref}} \exp[-\theta c_{ij}] \, z_{jk}^{\mathrm{h}} \right] \nonumber \\
&= -\frac{1}{\theta} \log \left[ \sum_{j=1}^{n} p_{ij}^{\mathrm{ref}} \exp[-\theta c_{ij}] \, \exp[-\theta (- \frac{1}{\theta} \log z_{jk}^{\mathrm{h}})] \right] \nonumber \\
&= -\frac{1}{\theta} \log \left[ \sum_{j=1}^{n} p_{ij}^{\mathrm{ref}} \exp[-\theta c_{ij}] \, \exp[-\theta \phi(j,k)] \right] \nonumber \\
&= -\frac{1}{\theta} \log \left[ \sum_{j \in \mathcal{S}ucc(i)} p_{ij}^{\mathrm{ref}} \exp[-\theta (c_{ij} + \phi(j,k))] \right]
\label{Eq_recurrence_formula_potential01}
\end{align}
which provides a recurrence formula for computing $\phi(i,k)$, together with the boundary condition $\phi(k,k) = 0$.

Let us now study the behavior of this equation for $\theta \rightarrow \infty$.
We first observe that both the numerator and the denominator tend to $+\infty$ when $\theta \rightarrow \infty$.

Now, in order to simplify the notations, we will study the $\mathrm{softmin}$ function \cite{Cook-2011,Tahbaz-2006}, $\mathrm{softmin}_{\mathbf{q},\theta}(\mathbf{x}) = -\log(\sum_{j=1}^{n} q_{j} \exp[-\theta x_{j}])/\theta$ with $\sum_{j=1}^{n} q_{j}=1$ and all $q_{j} \ge 0$ instead, where we define $x_{j} = (c_{ij} + \phi(j,k))$ and $q_{j} = p_{ij}^{\mathrm{ref}}$ (the development is inspired by \cite{Tahbaz-2006}). Let us further define $x^{*} = \min_{j} (x_{j})$ so that $(x_{j} - x^{*}) \ge 0$; we then have
\begin{align}
\lim_{\theta \rightarrow \infty} \mathrm{softmin}_{\mathbf{q},\theta}(\mathbf{x})
&= \lim_{\theta \rightarrow \infty} -\dfrac{\log \left[ {\displaystyle \sum_{j=1}^{n}} q_{j} \exp[-\theta x_{j}] \right]}{\theta} \nonumber \\
&= \lim_{\theta \rightarrow \infty} -\dfrac{\log\left[\exp[-\theta x^{*}] \, {\displaystyle \sum_{j=1}^{n}} q_{j} \exp[-\theta (x_{j} - x^{*})] \right]}{\theta} \nonumber \\
&= \lim_{\theta \rightarrow \infty} \left[ x^{*} -  \dfrac{\log\left[{\displaystyle \sum_{j=1}^{n}} q_{j} \exp[-\theta (x_{j} - x^{*})] \right]}{\theta} \right] \nonumber \\
&= x^{*} -  \lim_{\theta \rightarrow \infty} \dfrac{\log\left[{\displaystyle \sum_{j=1}^{n}} q_{j} \exp[-\theta (x_{j} - x^{*})] \right]}{\theta} \nonumber \\
&= x^{*} 
\label{Eq_limit_potential_infty01}
\end{align}
and the last limit is 0 because no term in the exponential is positive and at least one of the $x_{j}$ is \emph{exactly} equal to $x^{*}$ (the minimum) so that the sum $\sum_{j=1}^{n} q_{j} \exp[-\theta (x_{j} - x^{*})]$ is non-zero, and thus strictly positive.

Thus, when $\theta \rightarrow \infty$, Equation (\ref{Eq_recurrence_formula_potential01}) becomes $\phi(i,k) = \min_{j} (c_{ij} + \phi(j,k))$ for $i \ne k$ and $\phi(k,k) = 0$ which is the well-known Bellman-Ford formula for computing the shortest path distance in an undirected graph (see, e.g., \cite{Bertsekas-2000,Christofides_1975,Cormen-2009,Jungnickel-2005,Rardin-1998,Sedgewick-2011}).
Moreover, for an undirected graph, the shortest path from $i$ to $j$ is equal to the shortest path from $j$ to $i$, which implies that $\dist^{\phi}$ reduces to the shortest path too when $\theta \rightarrow \infty$.

\section{Asymptotic result: for an undirected graph, the $\dist^{\phi}$ distance converges to half the commute cost distance when $\theta \rightarrow 0^{+}$}\label{app_commuteCostDist}

Let us show that the $\dist^{\phi}$ distance is half the commute cost distance when $\theta \rightarrow 0^{+}$. As before, there are two ways to prove this property. The first proof is based on the bag-of-paths framework and is somewhat shorter. The second proof, also inspired by \cite{Tahbaz-2006}, establishes some interesting links with the Bellman-Ford recurrence formula computing the average first-passage cost in a network \cite{Kemeny-1960,Norris-1997,Ross-2000,Taylor-1998}.

\subsection{First proof}

From Equations (
\ref{Eq_bag_of_paths_potential_distance01}) and (\ref{Eq_potential_function_hitting02}),
\begin{align}
  &\dist^{\phi}_{ij} = -\frac{(\log z_{ij}^{\mathrm{h}} + \log z_{ji}^{\mathrm{h}})} {2\theta} \nonumber \\
  &= -\frac{ \log (\sum_{\wp\in\mathcal{P}^{\mathrm{h}}_{ij}} \tilde{\pi}^{\mathrm{ref}}(\wp) \exp [-\theta \tilde{c}(\wp)]) + \log (\sum_{\wp\in\mathcal{P}^{\mathrm{h}}_{ji}} \tilde{\pi}^{\mathrm{ref}}(\wp) \exp [-\theta \tilde{c}(\wp)]) } {2\theta} \label{Eq_commute_cost01}
\end{align}
and, because $\sum_{\wp\in\mathcal{P}^{\mathrm{h}}_{ij}} \tilde{\pi}^{\mathrm{ref}}(\wp) = 1$ (see \ref{app_hitting_paths_likelihood}), both the numerator and the denominator tend to zero when $\theta \rightarrow 0^{+}$.
For taking the limit $\theta \rightarrow 0^{+}$ of the whole expression (\ref{Eq_commute_cost01}), we apply l'Hospital's rule (taking the derivative of the numerator and the denominator with respect to $\theta$ and then the limit $\lim_{\theta \rightarrow 0^{+}}$ of the resulting expression). Because the Gibbs-Boltzmann probability distribution over the hitting paths tends to $\tilde{\pi}^{\mathrm{ref}}$ when $\theta \rightarrow 0^{+}$ (see Equation (\ref{Eq_Lagrange_function01})), this provides
\begin{equation}
  \lim_{\theta \rightarrow 0^{+}} \dist^{\phi}_{ij} = \frac{ \sum_{\wp\in\mathcal{P}^{\mathrm{h}}_{ij}} \tilde{\pi}^{\mathrm{ref}}(\wp) \, \tilde{c}(\wp) + \sum_{\wp\in\mathcal{P}^{\mathrm{h}}_{ji}} \tilde{\pi}^{\mathrm{ref}}(\wp) \, \tilde{c}(\wp) } {2}
\label{Eq_commute_cost02}
\end{equation}

The quantity $\sum_{\wp\in\mathcal{P}^{\mathrm{h}}_{ij}} \tilde{\pi}^{\mathrm{ref}}(\wp) \, \tilde{c}(\wp)$ can be interpreted as the average first-passage cost from $i$ to $j$, i.e. the average cost undergone by a random walker using transition probabilities $p^{\mathrm{ref}}_{ij}$ for reaching destination node $j$ for the first time when starting from $i$. Consequently, the average of the two quantities defined in (\ref{Eq_commute_cost02}) is half the commute cost distance.

\subsection{Second proof}

Restarting from Equation (\ref{Eq_recurrence_formula_potential01}), we now have to take the limit $\theta \rightarrow 0^{+}$. Assuming $\sum_{j=1}^{n} q_{j} = 1$, let us compute the limit $\theta \rightarrow 0^{+}$ of $\mathrm{softmin}_{\mathbf{q},\theta}(\mathbf{x})$, instead of $\theta \rightarrow \infty$ in Equation (\ref{Eq_limit_potential_infty01}), and apply as before l'Hospital's rule
\begin{align}
\lim_{\theta \rightarrow 0^{+}} \mathrm{softmin}_{\mathbf{q},\theta}(\mathbf{x})
&= \lim_{\theta \rightarrow 0^{+}} -\dfrac{\log \left({\displaystyle \sum_{j=1}^{n}} q_{j} \exp[-\theta x_{j}] \right)}{\theta} \nonumber \\
&= \lim_{\theta \rightarrow 0^{+}} \dfrac{ {\displaystyle \sum_{j=1}^{n}} \, q_{j} x_{j} \exp[-\theta x_{j}] }{{\displaystyle \sum_{j'=1}^{n}} q_{j'} \exp[-\theta x_{j'}]} = \dfrac{ {\displaystyle \sum_{j=1}^{n}} \, q_{j} x_{j} }{{\displaystyle \sum_{j'=1}^{n}} q_{j'} }
\label{Eq_limit_potential_zero01}
\end{align}

Therefore, as in our case $x_{j} = (c_{ij} + \phi(j,k))$ and $q_{j} = p_{ij}^{\mathrm{ref}}$ with $\sum_{j=1}^{n} p_{ij}^{\mathrm{ref}} = 1$, we obtain $\phi(i,k) = \sum_{j=1}^{n} p_{ij}^{\mathrm{ref}} (c_{ij} + \phi(j,k))$ for $i \ne k$, together with the boundary condition $\phi(k,k) = 0$. But this is exactly the recurrence formula computing the average first-passage cost in a regular Markov chain \cite{Kemeny-1960,Norris-1997,Ross-2000,Taylor-1998}. Thus, when $\theta \rightarrow 0^{+}$, $\dist^{\phi} = (\phi(i,j) + \phi(j,i))/2$ reduces to half the commute cost distance between $i$ and $j$.

\section{Triangle inequality for hitting paths and weights on nodes}
\label{Sec_triangle_inequality_prior01}

To prove the result we simply adapt the corresponding proof of \ref{app_surprisal_dist_proof}. Note that Equation (\ref{Eq_bag_of_paths_probability_inequality01}) still holds. Moreover, Equation (\ref{Eq_bag_of_paths_probability_intermediate02}) becomes
\begin{align}
&\text{P}(s=i,int=j,e=k)
= \frac{ q_{i}^{\mathrm{s}} q_{k}^{\mathrm{e}} {\displaystyle \sum_{\wp\in\mathcal{P}_{ik}}} \delta(j \in \wp) \, \tilde{\pi}^{\mathrm{ref}}(\wp)\exp\left[-\theta \tilde{c}(\wp)\right]}{\mathcal{Z}_{\mathrm{w}}} \nonumber \\
&= \frac{ q_{i}^{\mathrm{s}} q_{k}^{\mathrm{e}}{\displaystyle \sum_{\wp_{ij} \in \mathcal{P}^{\mathrm{h}}_{ij}} \sum_{\wp_{jk} \in \mathcal{P}_{jk}} } \tilde{\pi}^{\mathrm{ref}}(\wp_{ij}) \tilde{\pi}^{\mathrm{ref}}(\wp_{jk})\exp\left[-\theta (\tilde{c}(\wp_{ij}) + \tilde{c}(\wp_{jk})) \right]}{\mathcal{Z}_{\mathrm{w}}} \nonumber \\
&= \frac{ q_{i}^{\mathrm{s}} \left[ {\displaystyle \sum_{\wp_{ij} \in \mathcal{P}^{\mathrm{h}}_{ij}} } \tilde{\pi}^{\mathrm{ref}}(\wp_{ij}) \exp\left[-\theta \tilde{c}(\wp_{ij})  \right] \right] q_{j}^{\mathrm{e}} \times q_{j}^{\mathrm{s}} \left[ {\displaystyle \sum_{\wp_{jk} \in \mathcal{P}_{jk}} }  \tilde{\pi}^{\mathrm{ref}}(\wp_{jk}) \exp\left[-\theta \tilde{c}(\wp_{jk}) \right] \right] q_{k}^{\mathrm{e}} }{q_{j}^{\mathrm{s}} q_{j}^{\mathrm{e}} \mathcal{Z}_{\mathrm{w}}} \nonumber \\
&= \frac{\mathcal{Z}_{\mathrm{hw}}}{q_{j}^{\mathrm{s}} q_{j}^{\mathrm{e}}} \frac{ \left[ q_{i}^{\mathrm{s}} {\displaystyle \sum_{\wp_{ij} \in \mathcal{P}^{\mathrm{h}}_{ij}} } \tilde{\pi}^{\mathrm{ref}}(\wp_{ij}) \exp\left[-\theta \tilde{c}(\wp_{ij})  \right] \, q_{j}^{\mathrm{e}} \right] } {\mathcal{Z}_{\mathrm{hw}}} \frac{\left[ q_{j}^{\mathrm{s}} {\displaystyle \sum_{\wp_{jk} \in \mathcal{P}_{jk}} }  \tilde{\pi}^{\mathrm{ref}}(\wp_{jk}) \exp\left[-\theta \tilde{c}(\wp_{jk}) \right] \, q_{k}^{\mathrm{e}} \right] }{\mathcal{Z}_{\mathrm{w}}} \nonumber \\
&= \frac{\mathcal{Z}_{\mathrm{hw}}}{q_{j}^{\mathrm{s}} q_{j}^{\mathrm{e}}} \, \text{P}_{\mathrm{h}}(s=i,e=j) \, \text{P}(s=j,e=k), \text{ for } i \ne j \ne k \ne i
\label{Eq_bag_of_paths_probability_intermediate_prior02}
\end{align}
where $\mathcal{Z}_{\mathrm{w}} = \sum_{i,j=1}^{n} q_{i}^{\mathrm{s}} z_{ij} q_{j}^{\mathrm{e}}$ is the partition function for non-hitting paths (the counterpart of Equation (\ref{Eq_partition_function_prior01}) for non-hitting paths).

As for Equation (\ref{Eq_bag_of_paths_probability_inequality03}), combining this last result with (\ref{Eq_bag_of_paths_probability_inequality01}) yields
\begin{equation}
\text{P}(s=i,e=k) \ge \frac{\mathcal{Z}_{\mathrm{hw}}}{q_{j}^{\mathrm{s}} q_{j}^{\mathrm{e}}} \, \text{P}_{\mathrm{h}}(s=i,e=j) \, \text{P}(s=j,e=k), \text{ for } i \ne j \ne k \ne i
\label{Eq_bag_of_paths_probability_inequality_prior03}
\end{equation}


\noindent Then, by further considering that, from Equation (\ref{Eq_partition_function_prior01}), the following inequality holds
\begin{equation}
\frac{\mathcal{Z}_{\mathrm{hw}}}{q_{j}^{\mathrm{s}} q_{j}^{\mathrm{e}}} = \frac{1}{q_{j}^{\mathrm{s}} q_{j}^{\mathrm{e}}} \sum_{i,k=1}^{n} q_{i}^{\mathrm{s}} \, \left( \frac{z_{ik}}{z_{kk}} \right) \, q_{k}^{\mathrm{e}} \ge 1
\end{equation}
because the term $i=k=j$ in the double sum is equal to 1.

We deduce that $\text{P}(s=i,e=k) \ge \text{P}_{\mathrm{h}}(s=i,e=j) \, \text{P}(s=j,e=k)$.
Then, dividing both sides by $(\mathcal{Z}_{\mathrm{hw}} z_{kk})$ and using Equation (\ref{Eq_bag_of_paths_probability_hitting_prior02}), as well as $\text{P}(s=i,e=k) = (q_{i}^{\mathrm{s}} z_{ik} q_{k}^{\mathrm{e}})/\mathcal{Z}_{\mathrm{w}}$ for weighted nodes and non-hitting paths, provides the final result
\begin{equation}
-\log\text{P}_{\mathrm{h}}(s=i,e=k) \le -\log\text{P}_{\mathrm{h}}(s=i,e=j) -\log\text{P}_{\mathrm{h}}(s=j,e=k)
\label{Eq_bag_of_paths_probability_inequality_prior06}
\end{equation}
which shows the triangle inequality for the directed surprisal distance and, hence, the surprisal distance, in the case of weighted nodes.

The same triangle inequality result holds for the directed potential distance with weighted nodes, defined by $\phi(i,j) \triangleq -\frac{1}{\theta} \log (q_{i}^{\mathrm{s}} z_{ij}^{\mathrm{h}} q_{j}^{\mathrm{e}})$, and $z_{ij}^{\mathrm{h}}$ given in Equation (\ref{Eq_fundamental_matrix_prior01}).
Indeed, by replacing $\text{P}(\cdot)$ and $\text{P}_{\mathrm{h}}(\cdot)$ by their expressions in function of the $z_{ij}^{\mathrm{h}}$ in Equation (\ref{Eq_bag_of_paths_probability_inequality_prior03}) provides
\begin{equation}
q_{i}^{\mathrm{s}} z_{ik}^{\mathrm{h}} q_{k}^{\mathrm{e}} \ge \frac{1}{q_{j}^{\mathrm{s}} q_{j}^{\mathrm{e}}} \, (q_{i}^{\mathrm{s}} z_{ij}^{\mathrm{h}} q_{j}^{\mathrm{e}}) \, (q_{j}^{\mathrm{s}} z_{jk}^{\mathrm{h}} q_{k}^{\mathrm{e}})
\end{equation}
Then, because $1/q_{j}^{\mathrm{s}} q_{j}^{\mathrm{e}} \ge 1$ for every $j$, we obtain
\begin{equation}
- \frac{1}{\theta} \log ( q_{i}^{\mathrm{s}} z_{ik}^{\mathrm{h}} q_{k}^{\mathrm{e}} ) \le - \frac{1}{\theta} \log (q_{i}^{\mathrm{s}} z_{ij}^{\mathrm{h}} q_{j}^{\mathrm{e}}) - \frac{1}{\theta} \log(q_{j}^{\mathrm{s}} z_{jk}^{\mathrm{h}} q_{k}^{\mathrm{e}})
\end{equation}
which proves triangle inequality for the directed potential distance, and therefore also for the potential distance with priors on nodes.

{\footnotesize
\bibliographystyle{abbrv}
\bibliography{BiblioBagOfPath}

\begin{thebibliography}{100}

\bibitem{Abney-2008}
S.~Abney.
\newblock {\em Semisupervised learning for computational linguistics}.
\newblock Chapman and Hall/CRC, 2008.

\bibitem{Akamatsu-1996}
T.~Akamatsu.
\newblock Cyclic flows, markov process and stochastic traffic assignment.
\newblock {\em Transportation Research B}, 30(5):369--386, 1996.

\bibitem{vonLuxburg-2011}
M.~Alamgir and U.~von Luxburg.
\newblock Phase transition in the family of p-resistances.
\newblock In J.~Shawe-Taylor, R.~Zemel, P.~Bartlett, F.~Pereira, and
  K.~Weinberger, editors, {\em Advances in Neural Information Processing
  Systems 24 (NIPS 2011)}, pages 379--387, 2011.

\bibitem{Barabasi-2015}
A.~L. Barabasi.
\newblock {\em Network science}.
\newblock To appear at Cambridge University Press; preprint available from
  {barabasi.com/networksciencebook}, 2016.

\bibitem{Belkin-2004}
M.~Belkin, I.~Matveeva, and P.~Niyogi.
\newblock Tikhonov regularization and semi-supervised learning on large graphs.
\newblock In {\em Proceedings of the IEEE International Conference on
  Acoustics, Speech, and Signal Processing (ICASSP2004)}, pages 1000--1003,
  2004.

\bibitem{Bell-1995}
M.~Bell.
\newblock Alternatives to dial's logit assignment algorithm.
\newblock {\em Transportation Research Part B: Methodological}, 29(4):287--295,
  1995.

\bibitem{Bertsekas-2000}
D.~P. Bertsekas.
\newblock {\em Dynamic programming and optimal control, 2nd ed}.
\newblock Athena Scientific, 2000.

\bibitem{Blum-2001}
A.~Blum and S.~Chawla.
\newblock Learning from labeled and unlabeled data using graph mincuts.
\newblock In {\em International Conference on Machine Learning (ICML)}, pages
  19--26, 2001.

\bibitem{Borg-1997}
I.~Borg and P.~Groenen.
\newblock {\em Modern multidimensional scaling: Theory and applications}.
\newblock Springer, 1997.

\bibitem{Brand-05}
M.~Brand.
\newblock A random walks perspective on maximizing satisfaction and profit.
\newblock {\em Proceedings of the 2005 SIAM International Conference on Data
  Mining}, 2005.

\bibitem{Callut-2008}
J.~Callut, K.~Francoisse, M.~Saerens, and P.~Dupont.
\newblock Semi-supervised classification from discriminative random walks.
\newblock In {\em Proceedings of the European conference on Machine Learning
  (ECML 2008)}, volume LNAI5211, pages 162--177, 2008.

\bibitem{Chandra-1989}
A.~K. Chandra, P.~Raghavan, W.~L. Ruzzo, R.~Smolensky, and P.~Tiwari.
\newblock The electrical resistance of a graph captures its commute and cover
  times.
\newblock {\em Annual ACM Symposium on Theory of Computing}, pages 574--586,
  1989.

\bibitem{Chapelle-2006}
O.~Chapelle, B.~Scholkopf, and A.~Zien.
\newblock {\em Semi-supervised learning}.
\newblock MIT Press, 2006.

\bibitem{Chapelle-2002}
O.~Chapelle, J.~Weston, and B.~Sch{\"o}lkopf.
\newblock Cluster kernels for semi-supervised learning.
\newblock In {\em conference on Neural Information Processing Systems}, pages
  585--592, 2002.

\bibitem{Chebotarev-2011}
P.~Chebotarev.
\newblock A class of graph-geodetic distances generalizing the shortest-path
  and the resistance distances.
\newblock {\em Discrete Applied Mathematics}, 159(5):295--302, 2011.

\bibitem{Chebotarev-2012}
P.~Chebotarev.
\newblock The walk distances in graphs.
\newblock {\em Discrete Applied Mathematics}, 160(10--11):1484--1500, 2012.

\bibitem{Chebotarev-1997}
P.~Chebotarev and E.~Shamis.
\newblock The matrix-forest theorem and measuring relations in small social
  groups.
\newblock {\em Automation and Remote Control}, 58(9):1505--1514, 1997.

\bibitem{Chebotarev-1998a}
P.~Chebotarev and E.~Shamis.
\newblock On proximity measures for graph vertices.
\newblock {\em Automation and Remote Control}, 59(10):1443--1459, 1998.

\bibitem{Christofides_1975}
N.~Christofides.
\newblock {\em Graph theory: An algorithmic approach}.
\newblock Academic Press, 1975.

\bibitem{chung06}
F.~Chung and L.~Lu.
\newblock {\em Complex Graphs and Networks}.
\newblock American Mathematical Society, 2006.

\bibitem{Cinlar-1975}
E.~Cinlar.
\newblock {\em Introduction to Stochastic Processes}.
\newblock Prentice-Hall, 1975.

\bibitem{Cook-2011}
J.~Cook.
\newblock Basic properties of the soft maximum.
\newblock Unpublished manuscript available from
  {www.johndcook.com/blog/2010/01/13/soft-maximum}, 2011.

\bibitem{Cormen-2009}
T.~Cormen, C.~Leiserson, R.~Rivest, and C.~Stein.
\newblock {\em Introduction to algorithms, 3th Edition}.
\newblock The MIT Press, 2009.

\bibitem{Cox-2001}
T.~Cox and M.~Cox.
\newblock {\em Multidimensional scaling, 2nd ed.}
\newblock Chapman and Hall, 2001.

\bibitem{Demsar2006}
J.~Dem\v{s}ar.
\newblock Statistical comparisons of classifiers over multiple data sets.
\newblock {\em Journal of Machine Learning Research}, 7:1--30, Dec. 2006.

\bibitem{Devooght-2014}
R.~Devooght, A.~Mantrach, I.~Kivim\"aki, H.~Bersini, A.~Jaimes, and M.~Saerens.
\newblock Random walks based modularity: {A}pplication to semi-supervised
  learning.
\newblock In {\em Proceedings of the 23rd International World Wide Web
  Conference (WWW '14)}, pages 213--224, 2014.

\bibitem{Dial71}
R.~Dial.
\newblock A probabilistic multipath assignment model that obviates path
  enumeration.
\newblock {\em Transportation Research}, 5:83--111, 1971.

\bibitem{Snell-1984}
P.~G. Doyle and J.~L. Snell.
\newblock {\em Random Walks and Electric Networks}.
\newblock The Mathematical Association of America, 1984.

\bibitem{Dunham-2003}
M.~Dunham.
\newblock {\em Data Mining: Introductory and Advanced Topics}.
\newblock Prentice Hall, 2003.

\bibitem{Estrada-2012}
E.~Estrada.
\newblock {\em The structure of complex networks}.
\newblock Oxford University Press, 2012.

\bibitem{FoussKernelNN-2011}
F.~Fouss, K.~Francoisse, L.~Yen, A.~Pirotte, and M.~Saerens.
\newblock An experimental investigation of kernels on graphs for collaborative
  recommendation and semisupervised classification.
\newblock {\em Neural Networks}, 31:53--72, 2012.

\bibitem{FoussKDE-2005}
F.~Fouss, A.~Pirotte, J.-M. Renders, and M.~Saerens.
\newblock Random-walk computation of similarities between nodes of a graph,
  with application to collaborative recommendation.
\newblock {\em IEEE Transactions on Knowledge and Data Engineering},
  19(3):355--369, 2007.

\bibitem{Garcia-Diez-2011}
S.~Garc\'{\i}a-D\'{\i}ez, F.~Fouss, M.~Shimbo, and M.~Saerens.
\newblock A sum-over-paths extension of edit distances accounting for all
  sequence alignments.
\newblock {\em Pattern Recognition}, 44(6):1172--1182, 2011.

\bibitem{Garcia-Diez-2011b}
S.~Garc\'{\i}a-D\'{\i}ez, E.~Vandenbussche, and M.~Saerens.
\newblock A continuous-state version of discrete randomized shortest-paths.
\newblock {\em Proceedings of the 50th IEEE International Conference on
  Decision and Control (IEEE CDC 2011)}, pages 6570--6577, 2011.

\bibitem{Grinstead-1997}
C.~Grinstead and J.~L. Snell.
\newblock {\em Introduction to probability, 2nd ed}.
\newblock The Mathematical Association of America, 1997.

\bibitem{Guex-2016}
G.~Guex.
\newblock Interpolating between random walks and optimal transportation routes:
  Flow with multiple sources and targets.
\newblock {\em Physica A: Statistical Mechanics and its Applications},
  450:264--277, 2016.

\bibitem{Guex-2015}
G.~Guex and F.~Bavaud.
\newblock Flow-based dissimilarities: shortest path, commute time, max-flow and
  free energy.
\newblock In B.~Lausen, S.~{Krolak-Schwerdt}, and M.~Bohmer, editors, {\em Data
  science, learning by latent structures, and knowledge discovery}, volume 1564
  of {\em Studies in Classification, Data Analysis, and Knowledge
  Organization}, pages 101--111. Springer, 2015.

\bibitem{Ham2004}
J.~Ham, D.~Lee, S.~Mika, and B.~Scholkopf.
\newblock A kernel view of the dimensionality reduction of manifolds.
\newblock {\em Proceedings of the 21st International Conference on Machine
  Learning (ICML2004)}, pages 369--376, 2004.

\bibitem{Hashimoto-2015}
T.~Hashimoto, Y.~Sun, and T.~Jaakkola.
\newblock From random walks to distances on unweighted graphs.
\newblock In {\em Advances in Neural Information Processing Systems 28:
  Proceedings of the NIPS '15 Conference}. MIT Press, 2015.

\bibitem{Herbster-2009}
M.~Herbster and G.~Lever.
\newblock Predicting the labelling of a graph via minimum p-seminorm
  interpolation.
\newblock {\em Proceedings of the 22nd Annual Conference on Learning Theory
  (COLT2009)}, 2009.

\bibitem{Hofmann-2008}
T.~Hofmann, B.~Sch\"olkopf, and A.~J. Smola.
\newblock Kernel methods in machine learning.
\newblock {\em The Annals of Statistics}, 36(3):1171--1220, 2088.

\bibitem{Huang-1990}
X.~Huang, Y.~Ariki, and M.~Jack.
\newblock {\em Hidden {M}arkov models for speech recognition}.
\newblock Edinburgh University Press, 1990.

\bibitem{Isaacson-1976}
D.~Isaacson and R.~Madsen.
\newblock {\em Markov chains theory and applications}.
\newblock John Wiley \& Sons, 1976.

\bibitem{Ito-2005}
T.~Ito, M.~Shimbo, T.~Kudo, and Y.~Matsumoto.
\newblock Application of kernels to link analysis.
\newblock {\em Proceedings of the eleventh ACM SIGKDD International Conference
  on Knowledge Discovery and Data Mining}, pages 586--592, 2005.

\bibitem{Ivashkin-2016}
V.~Ivashkin and P.~Chebotarev.
\newblock Logarithmic proximity measures outperform plain ones in graph nodes
  clustering.
\newblock {\em ArXiv preprint paper, arXiv:1605.01046}, pages 1--10, 2016.

\bibitem{Jaynes-1957}
E.~T. Jaynes.
\newblock Information theory and statistical mechanics.
\newblock {\em Physical Review}, 106:620--630, 1957.

\bibitem{Joachims03}
T.~Joachims.
\newblock Transductive learning via spectral graph partitioning.
\newblock In {\em Proceedings of the 20$^{th}$ International Conference on
  Machine Learning (ICDM 2003)}, page 290~297, Washington DC, 2003.

\bibitem{Jungnickel-2005}
D.~Jungnickel.
\newblock {\em Graphs, networks, and algorithms, 3th ed.}
\newblock Springer, 2008.

\bibitem{Kapoor-2005}
A.~Kapoor, Y.~A. Qi, H.~Ahn, and R.~W. Picard.
\newblock Hyperparameter and kernel learning for graph based semi-supervised
  classification.
\newblock In {\em conference on Neural Information Processing Systems (NIPS)},
  pages 627--634, 2005.

\bibitem{Kappen-2007}
H.~Kappen.
\newblock An introduction to stochastic control theory, path integrals and
  reinforcement learning.
\newblock In J.~{Marro}, P.~L. {Garrido}, and J.~J. {Torres}, editors, {\em AIP
  conference proceedings: ninth Granada lectures, Cooperative Behavior in
  Neural Systems}, volume 887 of {\em American Institute of Physics Conference
  Series}, pages 149--181, 2007.

\bibitem{Kapur-1992}
J.~N. Kapur and H.~K. Kesavan.
\newblock {\em Entropy optimization principles with applications}.
\newblock Academic Press, 1992.

\bibitem{Kemeny-1960}
J.~G. Kemeny and J.~L. Snell.
\newblock {\em Finite Markov Chains}.
\newblock Springer-Verlag, 1976.

\bibitem{Kivimaki-2014}
I.~Kivim{\"a}ki, B.~Lebichot, J.~Saramaki, and M.~Saerens.
\newblock Two betweenness centrality measures based on randomized shortest
  paths.
\newblock {\em Scientific Reports}, 6:srep19668, 2016.

\bibitem{Kivimaki-2012}
I.~Kivim{\"a}ki, M.~Shimbo, and M.~Saerens.
\newblock Developments in the theory of randomized shortest paths with a
  comparison of graph node distances.
\newblock {\em Physica A: Statistical Mechanics and its Applications},
  393:600--616, 2014.

\bibitem{Klein-1993}
D.~J. Klein and M.~Randic.
\newblock Resistance distance.
\newblock {\em Journal of Mathematical Chemistry}, 12(1):81--95, 1993.

\bibitem{Kolaczyk-2009}
E.~Kolaczyk.
\newblock {\em Statistical analysis of network data: methods and models}.
\newblock Springer, 2009.

\bibitem{Kolaczyk-2009c}
E.~Kolaczyk, D.~Chua, and M.~Barthelemy.
\newblock Group betweenness and co-betweenness: inter-related notions of
  coalition centrality.
\newblock {\em Social Networks}, 31(3):190--203, 2009.

\bibitem{Kondor-2002}
R.~I. Kondor and J.~Lafferty.
\newblock Diffusion kernels on graphs and other discrete structures.
\newblock {\em Proceedings of the 19th International Conference on Machine
  Learning (ICML 2002)}, pages 315--322, 2002.

\bibitem{Langville-2006}
A.~N. Langville and C.~D. Meyer.
\newblock {\em Google's PageRank and Beyond: The Science of Search Engine
  Rankings}.
\newblock Princeton University Press, 2006.

\bibitem{Lebichot-2014}
B.~Lebichot, I.~Kivimaki, K.~Francoisse, and M.~Saerens.
\newblock Semi-supervised classification through the bag-of-paths group
  betweenness.
\newblock {\em IEEE Transactions on Neural Networks and Learning Systems},
  25(6):1173--1186, 2014.

\bibitem{Lewis09}
T.~G. Lewis.
\newblock {\em Network Science : Theory and Applications}.
\newblock Wiley, 2009.

\bibitem{Lichman-2013}
M.~Lichman.
\newblock {UCI} machine learning repository, 2013.

\bibitem{Lu-2011}
L.~L\"{u} and T.~Zhou.
\newblock Link prediction in complex networks: a survey.
\newblock {\em Physica A}, 390:1150--1170, 2011.

\bibitem{Macskassy-07}
S.~A. Macskassy and F.~Provost.
\newblock Classification in networked data: A toolkit and a univariate case
  study.
\newblock {\em Journal of Machine Learning Research}, 8:935--983, 2007.

\bibitem{Manning-2008}
C.~Manning, P.~Raghavan, and H.~Schutze.
\newblock {\em Introduction to information retrieval}.
\newblock Cambridge University Press, 2008.

\bibitem{Mantrach-2011}
A.~Mantrach, N.~van Zeebroeck, P.~Francq, M.~Shimbo, H.~Bersini, and
  M.~Saerens.
\newblock Semi-supervised classification and betweenness computation on large,
  sparse, directed graphs.
\newblock {\em Pattern Recognition}, 44(6):1212 -- 1224, 2011.

\bibitem{Mantrach-2009}
A.~Mantrach, L.~Yen, J.~Callut, K.~Francoise, M.~Shimbo, and M.~Saerens.
\newblock The sum-over-paths covariance kernel: a novel covariance between
  nodes of a directed graph.
\newblock {\em IEEE Transactions on Pattern Analysis and Machine Intelligence},
  32(6):1112--1126, 2010.

\bibitem{Mardia-1979}
K.~V. Mardia, J.~T. Kent, and J.~M. Bibby.
\newblock {\em Multivariate analysis}.
\newblock Academic Press, 1979.

\bibitem{Meyer-2000}
C.~D. Meyer.
\newblock {\em Matrix analysis and applied linear algebra}.
\newblock SIAM, 2000.

\bibitem{Nadler-2005}
B.~Nadler, S.~Lafon, R.~Coifman, and I.~Kevrekidis.
\newblock Diffusion maps, spectral clustering and eigenfunctions of
  fokker-planck operators.
\newblock {\em Advances in Neural Information Processing Systems (NIPS) 18},
  pages 955--962, 2005.

\bibitem{Nadler-2006}
B.~Nadler, S.~Lafon, R.~Coifman, and I.~Kevrekidis.
\newblock Diffusion maps, spectral clustering and reaction coordinate of
  dynamical systems.
\newblock {\em Applied and Computational Harmonic Analysis}, 21:113--127, 2006.

\bibitem{Newman-2006}
M.~Newman.
\newblock Modularity and community structure in networks.
\newblock {\em Proceedings of the National Academy of Sciences (USA)},
  103:8577--8582, 2006.

\bibitem{Newman-2010}
M.~Newman.
\newblock {\em Networks: an introduction}.
\newblock Oxford University Press, 2010.

\bibitem{Norris-1997}
J.~R. Norris.
\newblock {\em Markov chains}.
\newblock Cambridge University Press, 1997.

\bibitem{Pan-2004}
J.-Y. Pan, H.-J. Yang, C.~Faloutsos, and P.~Duygulu.
\newblock Automatic multimedia cross-modal correlation discovery.
\newblock {\em Proceedings of the 10th ACM SIGKDD international conference on
  Knowledge Discovery and Data Mining (KDD 2004)}, pages 653--658, 2004.

\bibitem{Pons-2005}
P.~Pons and M.~Latapy.
\newblock Computing communities in large networks using random walks.
\newblock In P.~Yolum, T.~Gungor, F.~Gurgen, and C.~Ozturan, editors, {\em
  Proceedings of the 20th International Symposium on Computer and Information
  Sciences (ISCIS '05)}, volume 3733 of {\em Lecture Notes in Computer
  Science}, pages 284--293. Springer, 2005.

\bibitem{Pons-2006}
P.~Pons and M.~Latapy.
\newblock Computing communities in large networks using random walks.
\newblock {\em Journal of Graph Algorithms and Applications}, 10(2):191--218,
  2006.

\bibitem{Gori-2006WebKDD}
A.~Pucci, M.~Gori, and M.~Maggini.
\newblock A random-walk based scoring algorithm applied to recommender engines.
\newblock {\em Proceedings of the International Workshop on Knowledge Discovery
  on the Web (WebKDD 2006)}, pages 127--146, 2006.

\bibitem{Qiu2005}
H.~Qiu and E.~R. Hancock.
\newblock Image segmentation using commute times.
\newblock {\em Proceedings of the 16th British Machine Vision Conference (BMVC
  2005)}, pages 929--938, 2005.

\bibitem{Rardin-1998}
R.~Rardin.
\newblock {\em Optimization in operations research}.
\newblock Prentice Hall, 1998.

\bibitem{Ross-2000}
S.~Ross.
\newblock {\em Introduction to probability models, 10th Ed.}
\newblock Academic Press, 2010.

\bibitem{Saerens-2008}
M.~Saerens, Y.~Achbany, F.~Fouss, and L.~Yen.
\newblock Randomized shortest-path problems: Two related models.
\newblock {\em Neural Computation}, 21(8):2363--2404, 2009.

\bibitem{Saerens04PCA}
M.~Saerens, F.~Fouss, L.~Yen, and P.~Dupont.
\newblock The principal components analysis of a graph, and its relationships
  to spectral clustering.
\newblock {\em Proceedings of the 15th European Conference on Machine Learning
  (ECML 2004). Lecture Notes in Artificial Intelligence, vol. 3201,
  Springer-Verlag, Berlin}, pages 371--383, 2004.

\bibitem{Sarkar2007}
P.~Sarkar and A.~Moore.
\newblock A tractable approach to finding closest truncated-commute-time
  neighbors in large graphs.
\newblock {\em Proceedings of the 23rd Conference on Uncertainty in Artificial
  Intelligence (UAI)}, 2007.

\bibitem{Scholkopf-2002}
B.~Scholkopf and A.~Smola.
\newblock {\em Learning with kernels}.
\newblock The MIT Press, 2002.

\bibitem{Scholkopf-1998}
B.~Scholkopf, A.~Smola, and K.-R. Muller.
\newblock Nonlinear component analysis as a kernel eigenvalue problem.
\newblock {\em Neural Computation}, 5(10):1299--1319, 1998.

\bibitem{Sedgewick-2011}
R.~Sedgewick.
\newblock {\em Algorithms, 4th ed}.
\newblock Addison-Wesley, 2011.

\bibitem{Smola-03}
A.~J. Smola and R.~Kondor.
\newblock Kernels and regularization on graphs.
\newblock In M.~Warmuth and B.~Sch\"olkopf, editors, {\em Proceedings of the
  Conference on Learning Theory (COLT)}, pages 144--158, 2003.

\bibitem{Steele-2001}
J.~M. Steele.
\newblock {\em Stochastic calculus and financial application}.
\newblock Springer-Verlag, 2001.

\bibitem{Szummer-01}
M.~Szummer and T.~Jaakkola.
\newblock Partially labeled classification with markov random walks.
\newblock In T.~Dietterich, S.~Becker, and Z.~Ghahramani, editors, {\em
  Advances in Neural Information Processiong Systems}, volume~14, Vancouver,
  Canada, 2001. {MIT} Press.

\bibitem{Tahbaz-2006}
A.~Tahbaz and A.~Jadbabaie.
\newblock A one-parameter family of distributed consensus algorithms with
  boundary: from shortest paths to mean hitting times.
\newblock In {\em Proceedings of IEEE Conference on Decision and Control},
  pages 4664--4669, 2006.

\bibitem{Tang-2009}
L.~Tang and H.~Liu.
\newblock Relational learning via latent social dimensions.
\newblock In {\em Proceedings of the ACM conference on Knowledge Discovery and
  Data Mining (KDD 2009)}, pages 817--826, 2009.

\bibitem{Tang-2009b}
L.~Tang and H.~Liu.
\newblock Scalable learning of collective behavior based on sparse social
  dimensions.
\newblock In {\em Proceedings of the ACM conference on Information and
  Knowledge Management (CIKM 2009)}, pages 1107--1116, 2009.

\bibitem{Tang-2010}
L.~Tang and H.~Liu.
\newblock Toward predicting collective behavior via social dimension
  extraction.
\newblock {\em IEEE Intelligent Systems}, 25(4):19--25, 2010.

\bibitem{Taylor-1998}
H.~M. Taylor and S.~Karlin.
\newblock {\em An introduction to stochastic modeling, 3th Ed.}
\newblock Academic Press, 1998.

\bibitem{Thelwall04}
M.~Thelwall.
\newblock {\em Link analysis: An information science approach}.
\newblock Elsevier, 2004.

\bibitem{Tong-2006}
H.~Tong, C.~Faloutsos, and J.-Y. Pan.
\newblock Fast random walk with restart and its applications.
\newblock {\em Proceedings of sixth IEEE International Conference on Data
  Mining}, pages 613--622, 2006.

\bibitem{Tong-2007}
H.~Tong, C.~Faloutsos, and J.-Y. Pan.
\newblock Random walk with restart: fast solutions and applications.
\newblock {\em Knowledge and Information Systems}, 14(3):327--346, 2008.

\bibitem{vonLuxburg-2010}
U.~von Luxburg, A.~Radl, and M.~Hein.
\newblock Getting lost in space: large sample analysis of the commute distance.
\newblock {\em Proceedings of the 23th Neural Information Processing Systems
  conference (NIPS 2010)}, pages 2622--2630, 2010.

\bibitem{Wang-2009}
J.~Wang, F.~Wang, C.~Zhang, H.~Shen, and L.~Quan.
\newblock Linear neighborhood propagation and its applications.
\newblock {\em IEEE Transactions on Pattern Analysis and Machine Intelligence},
  31(9):1600--1615, 2009.

\bibitem{Wasserman-1994}
S.~Wasserman and K.~Faust.
\newblock {\em Social network analysis: methods and applications}.
\newblock Cambridge University Press, 1994.

\bibitem{Yajima-2006}
Y.~Yajima and T.-F. Kuo.
\newblock Efficient formulations for 1-svm and their application to
  recommendation tasks.
\newblock {\em Journal of Computers}, 1(3):27--34, 2006.

\bibitem{Yen-2008}
L.~Yen, F.~Fouss, C.~Decaestecker, P.~Francq, and M.~Saerens.
\newblock Graph nodes clustering with the sigmoid commute-time kernel: A
  comprehensive study.
\newblock {\em Data \& Knowledge Engineering}, 68(3):338--361, 2009.

\bibitem{Yen-08K}
L.~Yen, A.~Mantrach, M.~Shimbo, and M.~Saerens.
\newblock A family of dissimilarity measures between nodes generalizing both
  the shortest-path and the commute-time distances.
\newblock In {\em Proceedings of the 14th SIGKDD International Conference on
  Knowledge Discovery and Data Mining (KDD 2008)}, pages 785--793, 2008.

\bibitem{Yen-2011}
L.~Yen, M.~Saerens, and F.~Fouss.
\newblock A link analysis extension of correspondence analysis for mining
  relational databases.
\newblock {\em IEEE Transactions on Knowledge and Data Engineering},
  23(4):481--495, 2011.

\bibitem{Luh-2005}
L.~Yen, D.~Vanvyve, F.~Wouters, F.~Fouss, M.~Verleysen, and M.~Saerens.
\newblock Clustering using a random-walk based distance measure.
\newblock {\em Proceedings of the 13th Symposium on Artificial Neural Networks
  (ESANN 2005)}, pages 317--324, 2005.

\bibitem{Zhang-2008b}
D.~Zhang and R.~Mao.
\newblock Classifying networked entities with modularity kernels.
\newblock In {\em Proceedings of the 17th ACM Conference on Information and
  Knowledge Management (CIKM 2008)}, pages 113--122. ACM, 2008.

\bibitem{Zhang-2008}
D.~Zhang and R.~Mao.
\newblock A new kernel for classification of networked entities.
\newblock In {\em Proceedings of 6th International Workshop on Mining and
  Learning with Graphs}, Helsinki, Finland, 2008.

\bibitem{Zhou-2003}
D.~Zhou, O.~Bousquet, T.~Lal, J.~Weston, and B.~Scholkopf.
\newblock Learning with local and global consistency.
\newblock In {\em Conference on Neural Information Processing Systems (NIPS
  2003)}, pages 237--244, 2003.

\bibitem{Zhou-05}
D.~Zhou, J.~Huang, and B.~Sch{\"o}lkopf.
\newblock Learning from labeled and unlabeled data on a directed graph.
\newblock {\em Proceedings of the 22nd International Conference on Machine
  Learning}, pages 1041--1048, 2005.

\bibitem{Zhou-04}
D.~Zhou and B.~Scholkopf.
\newblock Learning from labeled and unlabeled data using random walks.
\newblock {\em Proceedings of the 26th DAGM Symposium, (Eds.) Rasmussen}, pages
  237--244, 2004.

\bibitem{Zhu-2008}
X.~Zhu.
\newblock Semi-supervised learning literature survey.
\newblock In {\em
  http://pages.cs.wisc.edu/~jerryzhu/research/ssl/semireview.html}, 2008.

\bibitem{Zhu-2009}
X.~Zhu, G.~Andrew, B., B.~Ronald, J., and D.~Thomas, G.
\newblock {\em Introduction to Semi-supervised Learning (Synthesis Lectures on
  Artificial Intelligence and Machine Learning)}.
\newblock Morgan \& Claypool Publishers, 2009.

\end{thebibliography}
}

\end{document}